\documentclass[twocolumn,english,prl,superscriptaddress,nofootinbib]{revtex4-1}
\usepackage[T1]{fontenc}
\usepackage[latin9]{inputenc}
\usepackage{color}
\usepackage{babel}
\usepackage{units}
\usepackage{amsmath}
\usepackage{amssymb}
\usepackage{graphicx}
\usepackage[unicode=true,pdfusetitle,
bookmarks=true,bookmarksnumbered=false,bookmarksopen=false,
breaklinks=false,pdfborder={0 0 0},pdfborderstyle={},backref=false,colorlinks=true]
{hyperref}
\hypersetup{
	urlcolor=orange,linkcolor=blue,citecolor=red}

\makeatletter
\usepackage{tikz}
\usepackage{color}
\usepackage{newtxtext}
\usepackage{newtxmath}
\usepackage{dsfont}

\makeatletter
\newsavebox{\@brx}
\newcommand{\llangle}[1][]{\savebox{\@brx}{\(\m@th{#1\langle}\)}%
	\mathopen{\copy\@brx\kern-0.5\wd\@brx\usebox{\@brx}}}
\newcommand{\rrangle}[1][]{\savebox{\@brx}{\(\m@th{#1\rangle}\)}%
	\mathclose{\copy\@brx\kern-0.5\wd\@brx\usebox{\@brx}}}
\makeatother

\makeatother
\begin{document}
	\title{Learning through atypical ''phase transitions'' in overparameterized  neural networks}
	
	\author{Carlo Baldassi}
	\affiliation{Artificial Intelligence Lab, Bocconi University, 20136 Milano, Italy}
	\author{Clarissa Lauditi}
	\affiliation{Department of Applied Science and Technology, Politecnico di Torino, 10129 Torino, Italy}
	\author{Enrico M. Malatesta}
	\affiliation{Artificial Intelligence Lab, Bocconi University, 20136 Milano, Italy}
	\author{Rosalba Pacelli}
	\affiliation{Department of Applied Science and Technology, Politecnico di Torino, 10129 Torino, Italy}
	\author{Gabriele Perugini}
	\affiliation{Artificial Intelligence Lab, Bocconi University, 20136 Milano, Italy}
	\author{Riccardo Zecchina}
	\affiliation{Artificial Intelligence Lab, Bocconi University, 20136 Milano, Italy}

	\email{enrico.malatesta@unibocconi.it}	
	
	\date{\today}
	\begin{abstract}
		Current deep neural networks are highly overparameterized (up to billions of connection weights) and nonlinear. Yet they can fit data almost perfectly through variants of gradient descent algorithms and achieve unexpected levels of prediction accuracy without overfitting. These are formidable results that defy predictions of statistical learning and pose conceptual challenges for non-convex optimization.
		In this paper, we use methods from statistical physics of disordered systems to analytically study the computational fallout of overparameterization in non-convex binary neural network models, trained on data generated from a structurally simpler but ``hidden'' network.
		As the number of connection weights increases, we follow the changes of the geometrical structure of different minima of the error loss function and relate them to learning and generalization performance.
		A first transition happens at the so-called interpolation point, when solutions begin to exist (perfect fitting becomes possible). This transition reflects the properties of typical solutions, which however are in sharp minima and hard to sample.
		After a gap, a second transition occurs, with the discontinuous appearance of a different kind of ``atypical'' structures: wide regions of the weight space that are particularly solution-dense and have good generalization properties. The two kinds of solutions coexist, with the typical ones being exponentially more numerous, but empirically we find that efficient algorithms sample the atypical, rare ones. This suggests that the atypical phase transition is the relevant one for learning.
		The results of numerical tests with realistic networks on observables suggested by the theory are consistent with this scenario.
	\end{abstract}
	\maketitle

	Machine learning has recently advanced in a totally unexpected way thanks to deep learning (DL), reaching unprecedented performance in many fields of  data-driven research and applications. 
	Impressive  spin-offs are emerging not only in technological applications but also in a wide variety of basic scientific fields, from molecular biology and language processing, to the solution of partial differential equations for the study of materials and fluids, to name a few recent examples.
	At the same time theoretical research is trying to build a unifying framework that explains deep learning performance, enables its development based on first principles, and paves the way towards interdisciplinary methodological and modeling connections, including computational neuroscience.

	Among the most disruptive aspects of deep learning models are their highly overparameterized and non-convex nature. Both of these aspects are a common trait of all the DL models and have led to unexpected results for classical statistical learning theory and non-convex optimization.
	Current deep neural networks (DNN) are composed of millions (or even billions) of connection weights and the learning process seeks to minimize the number of classification errors made by the DNN over a training set. This optimization problem is highly non-convex, and learning algorithms need to efficiently find good minima in a space of extremely high dimensionality without being trapped in local minima or saddle points for long times. Good minima are those that have good generalization capabilities, namely that do not suffer from overfitting given the inherent noisiness in the data and the huge number of parameters that can be adjusted. Surprisingly, this goal can often be achieved by relatively simple algorithms based on variants of the gradient descent method.
	
	We are thus facing of two conceptually stimulating facts:
	(i) highly expressive neural network models can fit the training data via simple variants of algorithms originally designed for convex optimization;
	(ii) even if trained with little control over their statistical complexity, these models achieve high levels of prediction accuracy, contrary to what classical statistical intuitions (such as the bias-variance tradeoff) would suggest. 
	
	In this paper, we focus on the computational fallout of overparameterization in non-convex models. As the number of parameters increases, we study the changes in the geometric structure of the different minima of the error loss function and we relate this to learning performance.
	
	Intuitively, one might be tempted to think that non-convex neural network models become effectively convex (with most of the weight volume of the minima associated to wide, accessible ones) when the number of weights becomes sufficiently large relative to the number of data to be classified. We will show analytically that this is not the case already in a simple one-layer binary weights overparametrized model. To the contrary, we find that an exponential number of sharp, isolated solutions with poor generalization properties exist even for very high levels of overparameterization. Indeed, these kind of solutions are by far in the majority, and algorithms that sample solutions with a flat measure find these typical ones (almost surely, in the limit of large system sizes); however, they also take an exponential amount of time in doing so. Thus, in practice, these typical solutions can only be found empirically in rather small networks. Efficient algorithms, that scale polynomially with the size of the problem, sample instead from wide regions of the space of the weights that are particularly dense with solutions and have good generalization properties.
	
	Both kinds of solutions have been studied in simpler, non-overparameterized models, using tools from statistical physics of disordered systems: the typical solutions are the equilibrium ones~\cite{krauth1989storage,huang2014origin}, and the atypical, highly entropic ones can be described by a large deviation technique~\cite{baldassi2015subdominant,unreasoanable,relu_locent,baldassi2020clustering,baldassi2020shaping,baldassi2020wide} or by using a robustness bias~\cite{baldassi2021unveiling}. Those techniques are non-rigorous, but a few rigorous confirmation of some of the findings have been obtained~\cite{perkins2021frozen,abbe2021proof,abbe2021binary}.
	
	Here, we extend those techniques to the study of the effect of overparameterization. We show that, contrary to what happens in overparameterized convex models, there are two transition points, separated by a gap. The first one is the information-theoretic interpolation threshold of the model: this is the point when zero-error solutions appear and perfect fitting of the data becomes possible. This point is obtained from an equilibrium computation and thus it is related to the typical, basically inaccessible solutions. The second transition point coincides with the sharp appearance of the highly locally entropic atypical solutions, that are attractive to learning algorithms.
	These dense regions stem from the development of new solutions which connect the preexisting ones.
	
	We shall call this second transition the \textit{Local Entropy} (LE) transition.
	This type of phase transition is not usually encountered in statistical physics, as it is driven by the appearance of rare structures in the solution space. Still it can be of basic relevance for learning processes (even very simple ones) that are not bound to try to sample from the dominating set of minima (i.e. are not designed to have the Gibbs distribution as stationary probability measure). This is indeed the case for all algorithms used for learning, which are subject to external perturbations, use ad hoc loss functions, and adopt peculiar optimization and initialization strategies, see also ref.~\cite{FengYuhai}.
	
	Interestingly, the phase transitions to rare states have similarities to the localization phase transitions that are well known in quantum mechanics~\cite{nandkishore2015many}. This fact is also consistent with the effectiveness of quantum annealing for learning problems similar to those discussed in this paper~\cite{baldassi2018efficiency}.
	
	The paper is organized as follows.
	In sec.~\ref{sec::1} we review some related literature, and introduce some basic non-convex analytically tractable versions of the random features models. In sec.~\ref{sec::2} we study analytically the geometric structure of the loss landscape, derive Bayesian generalization bounds and the phase diagram for the interpolation and LE transitions. In sec.~\ref{sec::3}, we report the results of numerical experiments on progressively less idealized and more realistic settings, validating the analytical findings, and confirming in particular that when the training algorithms start to be able to fit the data, they have already passed the interpolation point, and that they sample wide minima.

	\section{non-convex overparameterized neural classifiers}\label{sec::1}
	
	{\em Related work.} The  effects of overparameterization and the interpolation threshold have been recently studied in  convex neural classifiers in which the input data are projected in a arbitrarily high dimensional space. These models are variants of the  Random Features Model (RFM) which was first introduced as a tool to
	accelerate the training of Kernel machines~\cite{rahimi2007random,NealBook,lee2017neural}.
	More recently, the observation~\cite{jacot2018neural}  that infinitely wide neural networks operate in the so-called ``lazy regime'', where the weights do not change much from their initial values during the gradient descent training dynamics, suggested that the behavior of neural networks can be approximated to some extent by random feature models, where the randomness in the features comes from the random initialization of the network weights (see for example~\cite{geiger2020perspective} for a recent review). In the absence of specific regularization controls and for a given training set, as the size of the model increases the training and testing errors tend, initially, to decrease jointly. When the training error is about to reach perfect interpolation of the data, the test error begins to increase, giving rise to the famous U-shaped curve that describes the so called bias-variance trade-off in classical statistics. Not without surprise, if we keep adding parameters to the model, the test error behaves in a non-monotonic way: when the model  exceeds the interpolation threshold, the training error remains zero and the test error starts to fall again, and tends to an absolute minimum in the regime of extreme overparameterization where the number of parameters is much larger than the number of samples. This phenomenon, called ``double-descent''~\cite{Belkin2019, spigler2019jamming}, has been studied and reproduced in a number of different frameworks, ranging from rigorous computations~\cite{Mei2019} to statistical physics computations~\cite{Goldt2020, dascoli2020double, gerace2020generalisation,rocks2020memorizing}
	in simple models of neural nets, to realistic architectures, see for example refs.~\cite{nakkiran2021deep,geiger2020scaling}. 
	Subsequent numerical analysis of the Hessian of largely overparameterized models~\cite{sagun2017empirical} showed that minimizers present many flat directions, and that it is not hard to find a path of zero training error connecting two solutions~\cite{LiVisualizing2018,Draxler}. In underparameterized neural networks, on the other hand, the authors of~\cite{baity2018comparing} showed that the landscape is very rough and dynamics is glassy. 
	This led to think that the landscape of overparameterized networks where the dynamics is not glassy anymore, presents no ``poor'' minima at all~\cite{Spigler_2019}. According to our analysis, this is not the case. As we anticipated in the introduction, overparametrization has the effect of letting those connected regions appear at the LE transition, not letting ``poor'' minima completely disappear. Overparametrizing the network even further it is possible to increase the size of the connected region; ``poor'' or ``sharp'' solutions however remain the most numerous ones and dominate the Gibbs measure.

	{\em Overparameterized non-convex tractable model.} Here we consider a non-convex RFM for binary classification with two layers. We consider random weights in the first layer (the ``random features'') and a second layer with $N$ binary weights $\boldsymbol{w} \in \left\{ -1, 1 \right\}^N$ that are learned. Indeed, using binary weights suffices to make the overall learning problem highly non-convex. 
	
	

	In the model, each pattern $\boldsymbol{\xi}$ is generated on a hidden manifold, of dimension $D$, and projected as a pattern $\boldsymbol{\tilde\xi}$ on a visible feature space of dimension $N$. This models the common situation in which the raw input data is highly redundant, and its effective dimensionality is much lower.
	The projection is defined by:
	\begin{equation}
		\label{eq:projected_xi}
		\tilde{\xi}_i = \sigma \left(  \frac{1}{\sqrt{D}} \sum_{k=1}^{D} F_{ki} \xi_k \right)
	\end{equation}
	where $F$ is a $D\times N$ feature matrix and $\sigma\!\left( \cdot \right)$ is a non-linear activation function. In the following we will consider for definiteness $\sigma(x) = \text{sign}(x)$ and a feature matrix of the Gaussian Orthogonal Ensemble (GOE) type, i.e. every element of $F$ is a standard normal Gaussian; however our analytical results are valid for any $\sigma(\cdot)$ and every matrix having independent random entries with matching first and second moments, and that satisfy the hypothesis of the Gaussian Equivalence theorem~\cite{Mei2019,Goldt2020,gerace2020generalisation,goldt2021,hu2020}; see the details in the Supplementary Information (SI).
	
	The corresponding output of the network is:
	\begin{equation}
		\label{eq:perceptron}
		y_{\text{out}} \equiv \text{sign} \left( \frac{1}{\sqrt{N}} \sum_{i=1}^{N} w_i \, \tilde{\xi}_i \right)
	\end{equation}
	We consider a training set composed of $P = \alpha N$ random patterns extracted from a standard normal distribution; the label $y^\mu$ corresponding to a given pattern $\boldsymbol{\xi}^\mu$ is assigned by a ``teacher'' network having random binary weights $\boldsymbol{w}^T \in \left\{-1, 1\right\}^D$ as
	$y^\mu = \text{sign} \left( \frac{1}{\sqrt{D}} \sum_{k=1}^{D} w_k^T \xi_k^\mu \right)$.
	This intends to model the situation in which the true labels depend in a simple way from a latent representation, to which however the student network does not have access.
	The learning task consists in finding the weights $\boldsymbol{w}$ that fit all the data in the training set and that generalize well on the whole generative model.

	\section{Geometry of minima vs overparameterization: threshold phenomena} \label{sec::2}
	In the following we consider the primitive loss function that counts the number of misclassified patterns in the training set whose stability is greater than a given  margin of $\kappa \geq 0$. For each pattern, the stability $\Delta^\mu$ is defined as the product of the pre-activation of the output unit $\lambda^\mu(\boldsymbol{w}) $ and the binary  label of pattern  $y^\mu = \pm 1$: 
	\begin{equation}
		\Delta^\mu(\boldsymbol{w}) \equiv y^\mu 	\lambda^\mu(\boldsymbol{w})
	\end{equation}
	where
	\begin{equation}	
		\lambda^\mu(\boldsymbol{w}) \equiv \frac{1}{\sqrt{N}} \sum_{i=1}^{N} w_i \tilde{\xi}_i
	\end{equation}
	The loss function per pattern is defined as
	\begin{equation}
		\ell_{NE}\left(-\Delta^\mu(\boldsymbol{w}); \kappa\right) = \Theta\left(-\Delta^\mu(\boldsymbol{w}) + \kappa \right)
	\end{equation}
	where $\Theta(\cdot)$ is the Heaviside step function: $\Theta\left(x\right) = 1$ if $x>0$ and zero otherwise. For $\kappa = 0$ this loss reduces to the one that counts the number of training errors; with a slight abuse of language we call it ``number-of-errors loss'' even if the margin is non-zero. 
	For the analytical study, we will be interested in the large-size limit, where our calculations can be performed by asymptotic methods: $N, \, D, \, P \to \infty$ while keeping finite the ratios
	\begin{equation}
		\label{eq::thermodynamic_limit}
		\alpha \equiv \frac{P}{N} \; \; , \; \alpha_T \equiv \frac{P}{D} \; \; , \; {\alpha_D} \equiv \frac{D}{N} 
		\,, 
	\end{equation}
	with $\alpha = \alpha_T \alpha_D$. In order to compute the typical properties of the solution space, the key quantity of interest is the averaged free entropy of the model, i.e.
	\begin{equation}
		\label{eq::free_entropy}
		\phi = \lim\limits_{N, P, D \to \infty}\frac{1}{N} \langle \ln Z \rangle_{\xi, F}
	\end{equation}
	where we denoted with $\langle \bullet \rangle_{\xi, F}$ the average over both the patterns (including the desired outputs and thus the teacher) and the features. Here $Z$ denotes the partition function of the model which reads
	\begin{equation}
		Z(\beta) = \sum_{\boldsymbol{w}} \, e^{-\beta \sum_{\mu=1}^{P} \ell_{NE}\left( - \Delta^\mu(\boldsymbol{w}); \kappa \right)}
	\end{equation}
	For generic $\beta$, $Z(\beta)$ is the generating function in the  variable $e^{-\beta}$ of the number of errors. In the analytical computations however we have only considered the large $\beta$ limit, where the partition function reduces to counting the number of global minima, i.e. zero-error configurations (solutions) when they exist:
	\begin{equation}
		Z =\sum_{\boldsymbol{w}} \, \prod_{\mu=1}^P \Theta\left( \Delta^\mu(\boldsymbol{w}) - \kappa \right) \equiv \sum_{\boldsymbol{w}} \, \mathbb{X}_{\xi,F}(\boldsymbol{w}; \kappa)
	\end{equation}
	where $\mathbb{X}_{\xi,F}$ is the indicator function on $\boldsymbol{w}$ that all patterns are being correctly classified with the required robustness.
	
	The averages of the logarithm in eq.~\eqref{eq::free_entropy}, give access to the most probable number of solutions for a randomly chosen training set, and can be computed by asymptotic methods developed in the theory of disordered systems, either the so called replica method or the cavity method~\cite{mezard1987spin}.
	
	A first basic result of the analysis is that, for fixed $\alpha_T$ and $\kappa$, there is an $\alpha_\mathrm{max}(\kappa,\alpha_T)$ for which $\phi \ge 0$, signalling that, with high probability, for $\alpha > \alpha_\mathrm{max}(\kappa,\alpha_T)$ solutions with stability $\kappa$ or larger cease to exist. In this context, supposing that the learning problem and thus $\alpha_T$ was fixed and that we are controlling the degree of overparameterization via $\alpha$, the ``interpolation threshold'' $\alpha_c(\alpha_T)$ is the value of $\alpha$ for which \emph{all} solutions disappear, i.e. $\alpha_c(\alpha_T) = \alpha_\mathrm{max}(0,\alpha_T)$.
	
	Conversely, we also define the maximum margin $\kappa_\text{max}(\alpha,\alpha_T)$ for fixed values of $\alpha,\alpha_T$ as the value of $\kappa$ for which $\phi = 0$. The solutions with maximum margin play a central role in our analysis, since they lie in the middle of dense regions whose breakup (as $\alpha$ increases) signals the LE phase transition. It is useful to point out that even if the entropy of the solutions vanishes at $\kappa_\mathrm{max}$, their typical overlap (i.e. normalized dot product, also called cosine similarity) is still strictly smaller than $1$.
	
	\begin{figure}
		\begin{centering}
			\includegraphics[width=\columnwidth]{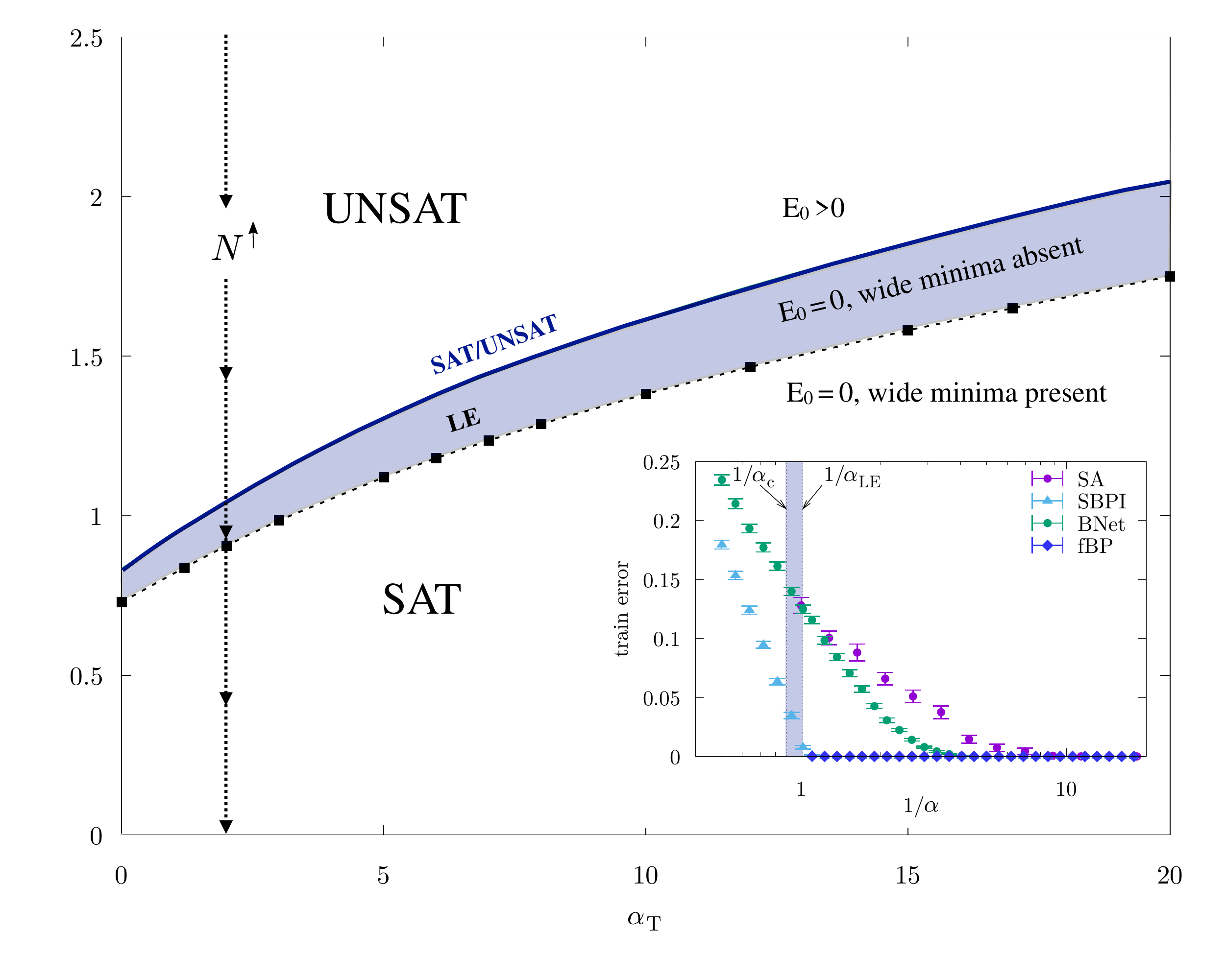}
		\end{centering}	
		\caption{SAT/UNSAT interpolation threshold and Local Entropy transition versus $\alpha_T$ for the binary non-convex model of random features. For $\alpha_T \to 0$ we recover the critical capacity~\cite{krauth1989storage} and the local entropy transition~\cite{baldassi2021unveiling} of a non-overparameterized binary perceptron trained on random patterns. In the inset we show the training error of SA, fBP and SBPI versus the degree of overparameterization $1/\alpha$ for $D=201$ and $\alpha_T = 3$. Points are averages over $20$ independent samples (except for fBP where we used $10$ samples) and $5$ independent runs per samples ($3$ for fBP). None of those algorithms is able to find solutions for $\alpha > \alpha_{\text{LE}}$. }
		\label{Fig::phase_diagram} 
	\end{figure}

	{\em Phase diagram}. 
	Before diving into analytical details, we anticipate how the geometry of the space of solutions changes as we increase the degree of overparameterization. The phase diagram of the model is reported in Fig.~\ref{Fig::phase_diagram}. The plane $(\alpha_T, \alpha)$ is divided into three distinct regions: 
	
	(1) an UNSAT region when the value of the density of constraints exceeds the interpolation threshold: $\alpha > \alpha_c(\alpha_T)$. In this region there exists no configuration of weights that is able to the whole training set. This threshold is independent of the learning algorithm, it depends only on the properties of the training data and of the architecture. On the other hand for $\alpha < \alpha_c(\alpha_T)$ we have a SAT region, so in principle the complexity of the model is sufficient to learn the data.
	
	(2) for $\alpha_{\text{LE}}(\alpha_T) < \alpha < \alpha_{c}(\alpha_T)$, despite the existence of configurations of weights that fit all the training set, they are either isolated or belong to minima that have a small characteristic size. These solutions turn out to be not easily accessible by learning algorithms.
	
	(3) for $\alpha \le \alpha_{\text{LE}}(\alpha_T)$  highly entropic wide minima start to appear. These flat minima, though exponentially rare compared to the isolated solutions, are accessible by simple, efficient algorithms. The threshold $\alpha_{\text{LE}}(\alpha_T)$ is thus the location of the Local Entropy transition, which we interpret as an upper bound for the effectiveness of learning algorithms.
	
	As an experimental check of this picture, we show in the inset of Fig.~\ref{Fig::phase_diagram} the train error of four algorithms that are representative of a spectrum of sampling strategies.
	On one extreme of the spectrum, we used Simulated Annealing (SA)~\cite{KirkpatrickSA}, which samples from the equilibrium Gibbs distribution. On the opposite end, we used focusing Belief Propagation (fBP)~\cite{unreasoanable}, which is a modified version of the message-passing Belief Propagation (BP) algorithm~\cite{yedidia2003understanding} and is designed to target high local entropy regions (if present). The goal of the original BP algorithm is to perform statistical inference, and at convergence its messages allow to derive the marginal probabilities for each variable, computed for a uniform distribution over the solutions of the training task. The modification introduced by fBP consists in forcing the messages to progressively focus on the most dense regions, until they become peaked on a single configuration, thereby resulting in an efficient solver. The focusing process is controlled by fixing an overall ``strength'' $y>1$ and by scheduling a parameter $\gamma$ from $0$ to $\infty$. Two more heuristic algorithms are specifically designed to work efficiently on binary architectures. One is the Stochastic BP-inspired (SBPI) algorithm~\cite{baldassi2007efficient}, which can be regarded as a simple and fast approximate version of fBP. The other is BinaryNet (BNet)~\cite{hubara2016binarized}, which is a modified version of Stochastic Gradient Descent (SGD). As we can observe in the figure, none of these algorithms can find solutions below $1/\alpha_{\mathrm{LE}}$.
	

	
	{\em Typical solutions}. 
	\begin{figure*}	
		\begin{centering}
			\includegraphics[width=0.49\textwidth]{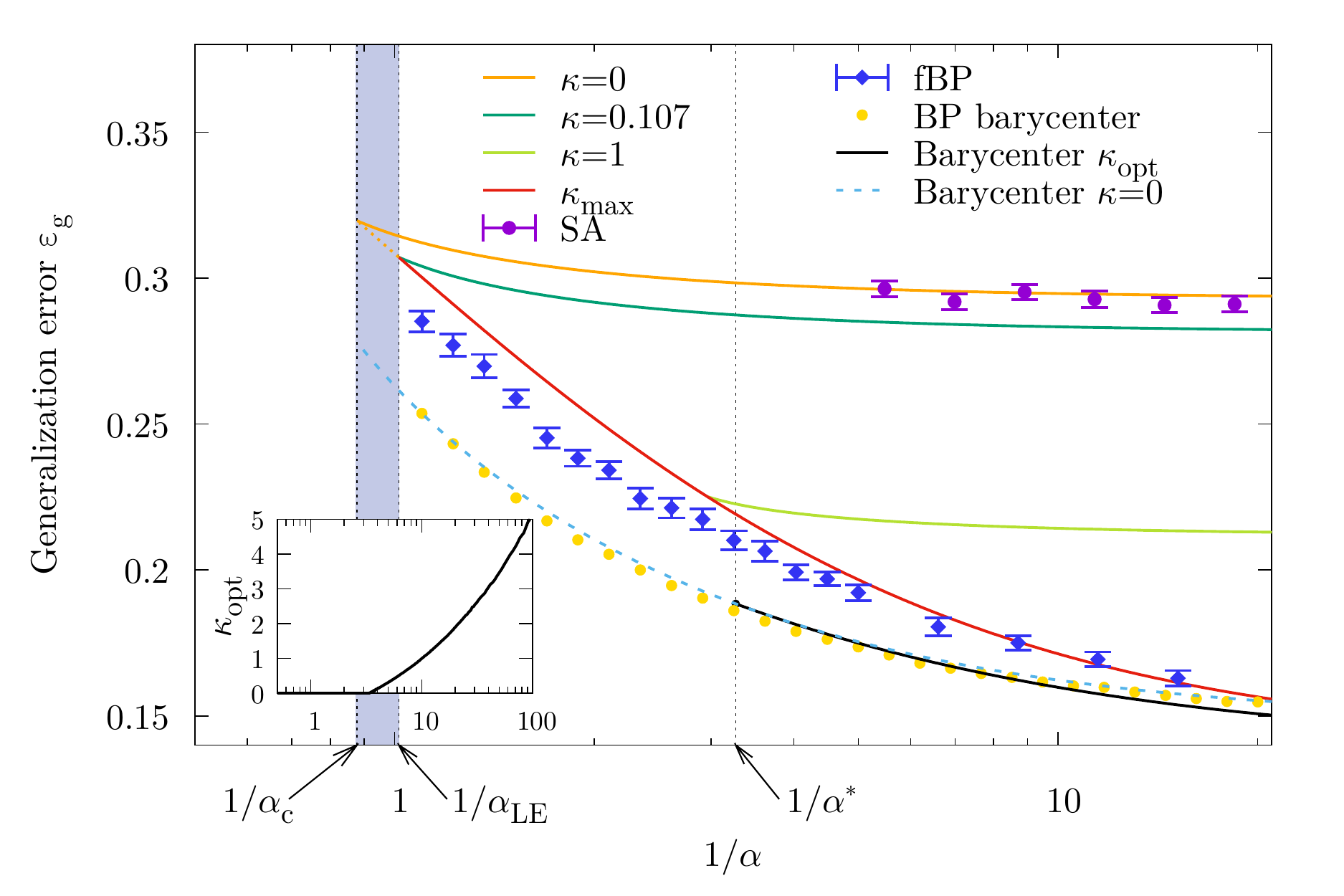}
			\includegraphics[width=0.49\textwidth]{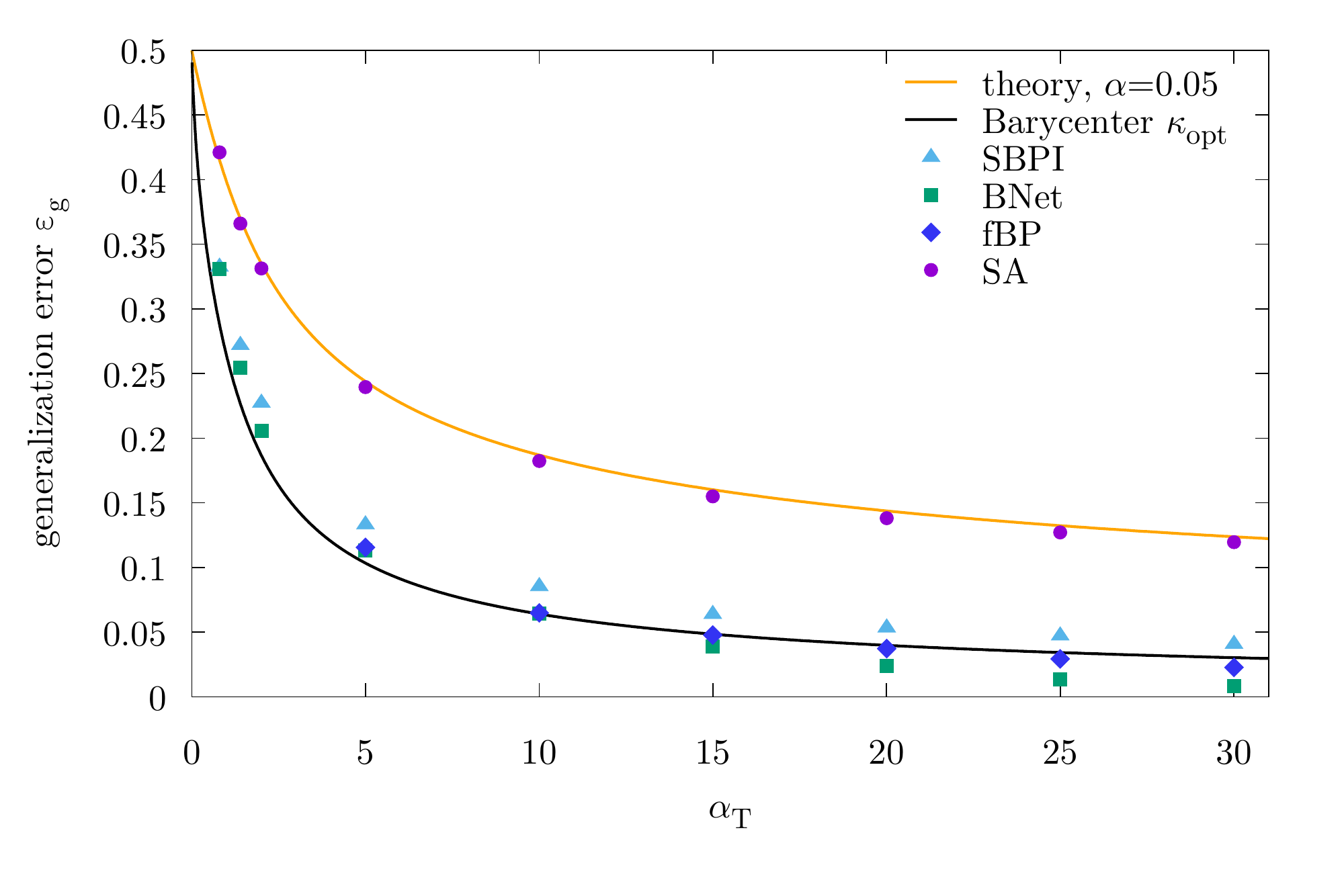}
		\end{centering}
		\caption{(Left panel) Generalization error as a function of the degree of overparameterization $1/\alpha$, for $\alpha_T = 3$. Vertical dashed lines denote the SAT-UNSAT transition $\alpha_c^{-1}$, the local entropy transition $\alpha_{\text{LE}}^{-1}$, and the $\kappa_{\mathrm{opt}}$ transition $\left(\alpha^*\right)^{-1}$. We show the generalization error of typical solutions with fixed margin $\kappa = 0$, $0.107$, $1$ and with maximum margin $\kappa_\mathrm{max}(\alpha,\alpha_T)$. The dashed turquoise curve is the error of the barycenter of typical solutions having zero margin, whereas the black line represents the generalization error of the ``best'' barycenter which was found by optimizing the margin. In the inset we show that this optimal margin $\kappa_{\mathrm{opt}}$ undergoes a transition when crossing $\alpha^{*}$. We also show numerical results ($D=201$) of two representative algorithms: SA (violet points) and fBP (blue points). When SA is able to find solutions, the corresponding generalization error is compatible to the one obtained by typical zero-margin configurations. In the large overparameterization regime fBP behaves similarly to the generalization error of the barycenter of zero-margin solutions. Yellow points (error bars not shown for clarity) represent the barycenter of zero margin solutions as computed by using the BP estimation of the posterior distribution. (Right panel) Generalization error versus $\alpha_T$ in the large overparameterization regime ($\alpha=0.05$, $D=201$). While SA gives the same generalization error of typical zero-margin solutions, SBPI, BNet and fBP, that do not target or sample from the Gibbs measure, perform much better. All points are averages over $5$ independent samples, $2$ independent runs per sample.
		}
		\label{Fig::gen_error} 
	\end{figure*}
	Using the replica method in its replica symmetric (RS) version (see SI), the averaged free entropy in eq.~\eqref{eq::free_entropy} turns out to depend on the ``order parameters'' $q$, $p$, $p_d$, $r$ and their conjugate Lagrange multipliers  $\hat q$, $\hat p$, $\hat p_d$, $\hat r$. Geometrically $q$ represent the typical overlap between a pair of solutions; $p$ is the typical overlap between a pair of solutions projected in the teacher space (which has dimension $D$), the projection being performed simply by using the feature matrix $F_{ki}$; $p_d$ is the typical squared norm of a projected solution and finally $r$ denotes the typical overlap between a projected solution and the teacher. 
	
	Eventually, $\phi$ can be found with the saddle point method, by optimizing over eight order parameters
	\begin{equation}
		\label{eq::RS_entropy}
		\begin{split}
			\phi &= \underset{q,\hat q, p, \hat p, p_d, \hat p_d, r, \hat r}{\max} \phi_{RS}\left( q,\hat q, p, \hat p, p_d, \hat p_d, r, \hat r \right)
		\end{split}
	\end{equation}
	where $\phi_{RS}$ is the RS expression for  $\phi$ (see SI). 
	Knowing the order parameters for which the function $\phi_{RS}$ is maximal allows to compute not only the entropy but also other quantities of interest, such as the generalization error $\epsilon_g$, defined as the probability of wrongly classifying a new (unseen) pattern
	\begin{equation}
		\epsilon_g = \mathbb{E}_\xi \, \ell_{NE}(-\Delta(\boldsymbol{w}); \kappa) \,.
	\end{equation}
	We find
	\begin{equation}
		\epsilon_g = \frac{1}{\pi} \arccos\left( \frac{M}{\sqrt{Q_d}} \right)
	\end{equation}
	where $M \equiv \mu_1 r$, $Q_d \equiv \mu_\star^2 + \mu_1^2 p_d$, and $\mu_1$, $\mu_\star$ are constants that depend only on the nonlinear function  $\sigma$ (see SI for their expressions). From the solutions of the saddle point equations we can also compute the probability that the average of the outputs of students sampled from the posterior on a random new pattern has different sign than that given by the teacher (see the SI for the definition); this turns out to be equivalent to computing the generalization error of the barycenter of typical solutions. All the details of the computation are reported in the SI; here we report the final result 
	\begin{equation}
		\epsilon_g^B = \frac{1}{\pi} \arccos\left( \frac{M}{\sqrt{Q}} \right)\,,
	\end{equation}
	where $Q \equiv \mu_\star^2 q + \mu_1^2 p$. In Fig.~\ref{Fig::gen_error} we show the plot of the generalization error of typical solutions with zero, non-zero and maximum possible margin versus the degree of overparameterization $1/\alpha$, together with the generalization error of the barycenter of typical solutions having zero margin. All those curves are monotonically decreasing.
	
	Moreover we show that the margin $\kappa_{\text{opt}}$ that should be imposed in order to minimize the generalization error of the barycenter undergoes a transition from zero (for $\alpha>\alpha^{*}$) to non-zero values (for $\alpha<\alpha^*$) whenever we increase the degree of overparameterization. The value of the optimal margin $\kappa_{\text{opt}}$ is plotted in the inset of Fig.~\ref{Fig::gen_error}. 
	
	
	{\em Numerical Checks}. In order to corroborate the analytical findings, we have performed some numerical experiments (see Fig.~\ref{Fig::gen_error}) using the four algorithms mentioned above: SA, fBP, SBPI and BNet.
	Similarly to  what happens in spin glass models, we found that for sufficiently low $\alpha$ (i.e. for relatively small system sizes) SA is able to escape from local minima and  find solutions that have generalization error which matches the one obtained by replica theory. We also found a perfect agreement between the theoretical results and the numerical experiments when we computed the distribution of the stabilities of typical configurations (see SI). We remark that the ability of SA to find solutions for low values of $\alpha$ is due to finite-size effects: indeed we show in the SI that scaling up the sizes while keeping $\alpha,\alpha_T$ fixed, at a certain point SA is no longer able to find solutions.
	We find that fBP, SBPI and BNet, despite being mildly affected by finite-size effects as well, converge to entropic states that have a much better generalization error, as also predicted by the theory.
	
	
	\begin{figure}
		\begin{centering}
			\includegraphics[width=0.98\columnwidth]{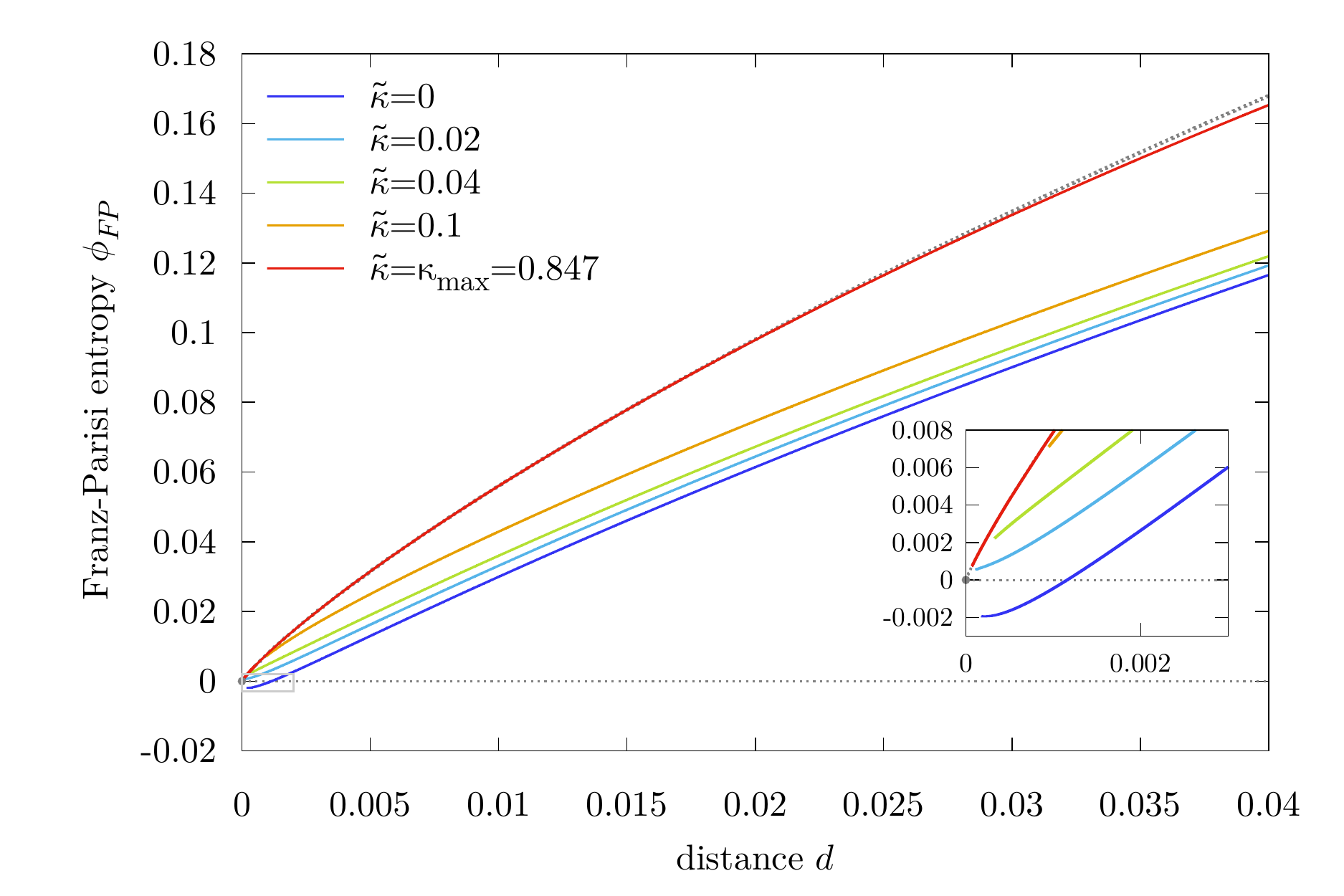}
		\end{centering}
		\caption{Local entropy of solutions with zero margin $\kappa = 0$ as a function of the distance $d$, evaluated for typical references having different values of the margin $\tilde{\kappa}$. Here $\alpha = 0.5$ and $\alpha_T = 8$. The (barely visible) dotted gray line at the top represents the total number of configurations at that given distance, which is a geometrical upper bound for the local entropy. The maximum margin that can be imposed is $\kappa_{\text{max}} \simeq 0.847$ and corresponds to the curve with the largest local entropy. The inset refers to the entropy curves for small distances (they are not complete due to numerical issues). For all curves $\phi_{\text{FP}}(d=0;\tilde{\kappa},\kappa) = 0$, so for $\tilde{\kappa}=0$ the curve is non-monotonic in a neighborhood of $d=0$, while for $\tilde{\kappa}>0$ the profiles shown are all positive.}
		\label{Fig::FP_RFM} 
	\end{figure}
	
	{\em The entropy landscape around a typical solution}. Having established that algorithms find solutions with different generalization properties, it remains to understand in which regions of the landscape those solutions end up and how they arise in terms of the degree of overparameterization. 
	
	A way to answer those questions is by studying the \emph{local entropy} landscape by the computation of the so-called \emph{Franz-Parisi} potential~\cite{franz1995recipes}. This technique has been introduced as a tool to study the role of metastable states in spin glasses~\cite{franz1995recipes} and recently~\cite{huang2014origin,relu_locent} it was used to show that, in one and two-layer binary neural networks, typical solutions with zero margin are organized as clusters with vanishing internal entropy, a scenario that has been called \emph{frozen}-1RSB. 
	
	Given a configuration $\tilde{\boldsymbol{w}}$ with margin $\tilde \kappa$ that we call the ``reference'', the local entropy is the log of the number of configurations $\mathcal{N}(\tilde{\boldsymbol{w}}, d; \kappa)$ that are solutions with margin $\kappa$ and that are constrained to be at a given normalized Hamming distance $d$ from $\tilde{\boldsymbol{w}}$:
	\begin{equation}
		\mathcal{N}(\tilde{\boldsymbol{w}}, d; \kappa) = \sum_{\boldsymbol{w}} \mathbb{X}_{\xi,F}(\boldsymbol{w}; \kappa)\,\delta\!\left( N(1-2d) - \sum_i w_i \tilde w_i \right) \,.
	\end{equation}
	The properties of the landscape around typical references can then be investigated by studying their average local entropy, which is called~\emph{Franz-Parisi} free entropy~\cite{franz1995recipes,huang2014origin}
	\begin{equation}
		\phi_{\text{FP}}(d; \tilde \kappa, \kappa) = \left\langle\frac{1}{Z} \sum_{\tilde{\boldsymbol{w}}} \mathbb{X}_{\xi, F}(\tilde{\boldsymbol{w}}; \tilde{\kappa}) \ln \mathcal{N}(\tilde{\boldsymbol{w}}, d; \kappa) \right\rangle_{\xi, F} \,.
	\end{equation}
	This quantity can be again computed by the replica method with a double analytic continuation (details in the SI).
	
	Here, following ref.~\cite{baldassi2021unveiling}, we are chiefly interested in the behavior of this quantity when $\kappa =  0$. For a given value of $\alpha < \alpha_c$, the local entropy curves of the references exhibit different characteristics as $\tilde \kappa$ varies. This is shown in Fig.~\ref{Fig::FP_RFM}. The overall picture closely resembles that of simpler models~\cite{baldassi2021unveiling}, and we point out some noteworthy results (here and in the following we omit $\alpha_T$, which we consider to be fixed, for simplicity):
	
	(1) Zero-margin references are isolated: $\phi_{\text{FP}}(d; \tilde \kappa=0, \kappa=0)$ is always negative in a neighborhood of $d=0$. This explains the poor performance, both in terms of efficiency in finding a solution and in terms of the generalization properties of the solution it finds, of the SA algorithm, which directly targets the Gibbs measure. Even for small non-zero values of $\tilde \kappa$ the local entropy is negative for some distances $d$, or it is non-monotonic, denoting the existence of small isolated clusters of solutions.
	
	(2) Fixing a small enough value of $\alpha$ and keeping increasing the margin of the reference configuration one eventually reaches a threshold value $\tilde\kappa = \kappa_u(\alpha)$ that separates a region for $\tilde\kappa < \kappa_u(\alpha)$ where the local entropy is non-monotonic (as described in the previous point) from a phase where the local entropy is monotonic (for $\tilde\kappa > \kappa_u(\alpha)$). 
	This means that those references are located inside a dense region of solutions that extends to very large scales. The monotonic local entropy phase extends up to $\tilde\kappa = \kappa_\text{max}(\alpha)$. As shown in Fig.~\ref{Fig::FP_RFM}, the highest curve in terms of local entropy is found by the typical configurations having maximum margin $\tilde\kappa = \kappa_\text{max}(\alpha)$. These large-scale regions are apparently targeted by efficient solvers, which also have lower generalization errors than SA.

	{\em Local Entropy transition}. A fundamental question that remains to be answered is how those (atypical) dense regions change when increasing $\alpha$. In previously studied convex models those regions tend to shrink continuously and they reduce to a point at the SAT/UNSAT transition. This is not the case here: similarly to what happens in previously studied teacher-student non-convex models, those regions shrink when increasing $\alpha$, until a critical value $\alpha_{\text{LE}} < \alpha_c$ is reached, beyond which they fracture in multiple pieces. This is the LE transition. For $\alpha>\alpha_{\text{LE}}$ no algorithm is seemingly able to find a solution efficiently, whereas below it efficient algorithms with good scaling properties only find solutions in non-isolated regions. Thus, $\alpha_{\text{LE}}$ can be regarded as a fairly good upper bound to the algorithmic capacity for the most efficient algorithms. 
	
\begin{figure}
		\begin{centering}
			\includegraphics[width=\columnwidth]{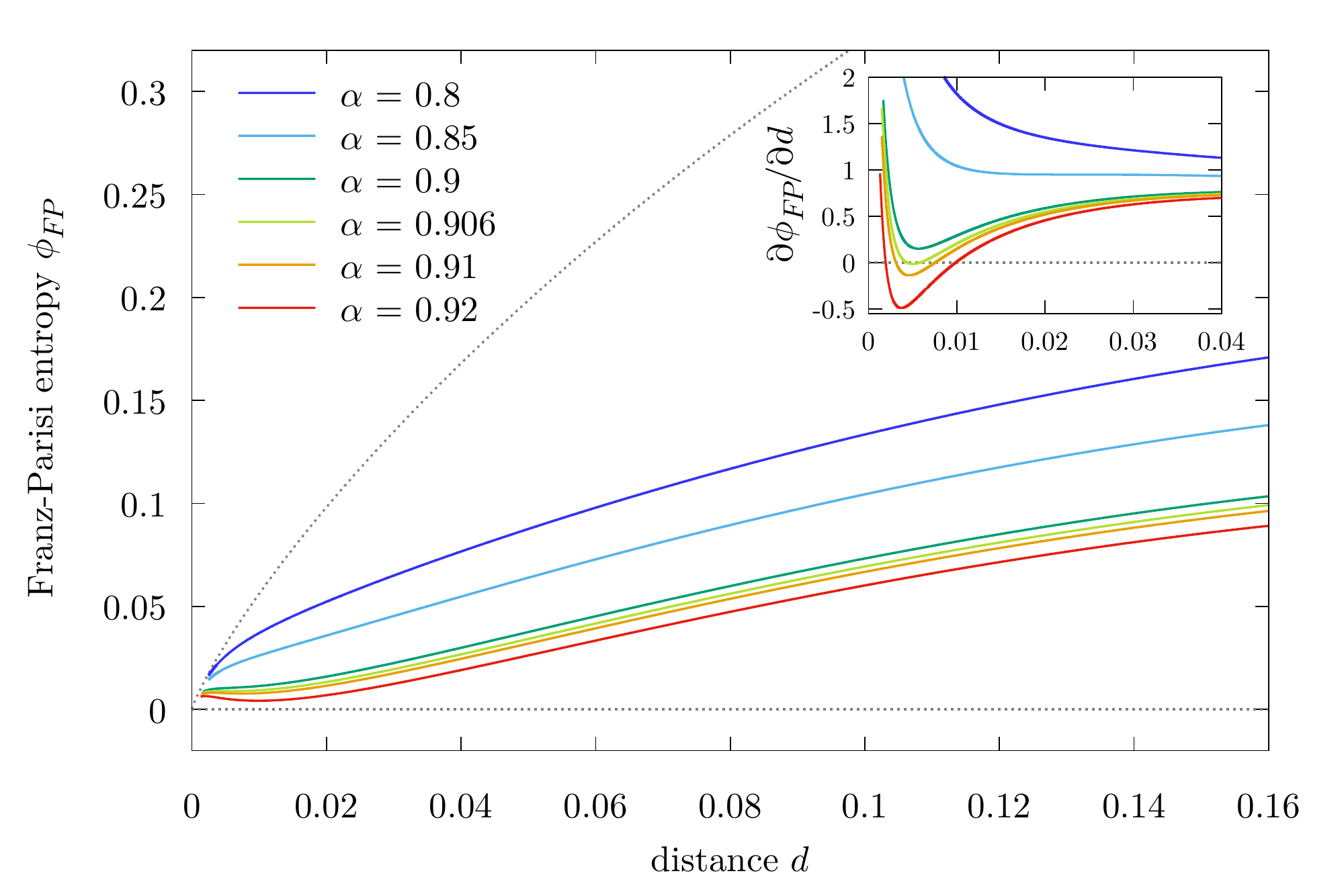}
		\end{centering}
		\caption{Estimating the Local Entropy transition $\alpha_{\text{LE}}$ by looking to the local entropy profiles of maximum margin references and its derivative with respect to distance (inset). Here $\alpha_T = 2$ and the critical capacity is $\alpha_c \simeq 1.045$. For low values of $\alpha$, e.g. $0.8$, $0.85$ and $0.9$ typical references with maximum margin are located inside a dense region extending to a very long scale, since the local entropy is monotonic. Near $\alpha = \alpha_{\text{LE}} \simeq 0.906$ the derivative of the local entropy develops a new zero for small distances. This signals a transition in the geometrical structure of the dense regions. For $\alpha > \alpha_{\text{LE}}$, the local entropy is not monotonic anymore, meaning that typical maximum margin references (as well as all other typical solutions with smaller margin) are located in disconnected balls in configuration space.}
		\label{Fig::FP_RFM_alphaLE} 
	\end{figure}

	Different approaches have been devised in order to estimate analytically $\alpha_{\text{LE}}$. The first one is based on the use of a large deviations analysis~\cite{baldassi2015subdominant,unreasoanable} which  however leads to a quite heavy formalism for the models under study.  We have thus adopted  a recently introduced simpler method~\cite{baldassi2021unveiling} which gives similar results to the large deviations approach. 
	It is based on the observation that, by definition of $\alpha_{\text{LE}}$, references located in the large-scale dense region should not exist anymore when $\alpha > \alpha_{\text{LE}}$. We can therefore estimate $\alpha_{\text{LE}}$ by the condition
	\begin{equation}
		\kappa_{u}(\alpha_{\text{LE}}) = \kappa_{\text{max}}(\alpha_{\text{LE}})
	\end{equation}
	meaning that $\alpha_{\text{LE}}$ is the value of $\alpha$ after which not even maximum margin solutions have a monotonic local entropy profile: all $\tilde\kappa$-margin solutions are located in disconnected balls in configuration space (see Fig.~\ref{Fig::FP_RFM_alphaLE}). This is a stricter condition than the one obtained from the large deviation analysis, which uses the criterion that \emph{all} solutions have non-monotonic profiles; thus, it likely slightly under-estimates the true $\alpha_{\text{LE}}$, but this difference is smaller than the resolution that can be detected by our numerical experiments.
	
	\section{Numerical experiments}\label{sec::3}
	
	In order to assess the relevance of the analysis presented above to more realistic cases, we have performed a series of numerical studies that consider progressively less idealized scenarios. First, we investigated the simplest non-convex continuous overparameterized model, namely a tree-like committee machine trained on randomly generated and randomly projected data, again with labels provided by a random teacher. 
	Second, we moved to deeper networks: we  studied a fully-connected multi-layer network with a fixed number of variable-width layers, trained with gradient descent using the popular ADAM optimizer, and a deep convolutional networks trained with both SGD and the ADAM optimizer. These deep models have been trained respectively on the first 10 principal components of a reduced version of the MNIST dataset and on images of CIFAR10.
	
	Across these tests, we found some common characteristics, compatible with the analytical findings. In order to find a solution, the networks require a minimum number of parameters that is larger than the size of the input. When that degree of overparameterization is achieved, we observe that indeed we have already passed the ``interpolation'' point, since many solutions exist (even after having accounted for the permutation and rescaling  symmetries in the networks) and they are located far apart from each other and belong to a flat region. As we increase the amount of overparameterization, the solutions that we found grow further apart in distance\footnote{The distance appears to plateau at a value strictly lower than the geometrical bound.}, their local landscapes become even flatter, and their generalization properties improve. 
	We also observe that within such flat regions, different algorithms sample solutions of different types, more or less barycentric, and with a different flatness (as estimated by their local energy profiles, defined below).

	{\em Overparameterized tree committee machine:} We studied an overparameterized tree-committee architecture with $K$ hidden units, trained on random patterns. The teacher and the patterns are generated in the same way as for the perceptron of eq.~\ref{eq:perceptron}, and in particular the device receives binary inputs $\boldsymbol{\tilde{\xi}}$ of length $N$ obtained by projecting randomly-generated $D$-dimensional inputs $\boldsymbol{\xi}$ through a random matrix $F$ and a non-linearity $\sigma$, as in eq.~(\ref{eq:projected_xi}). Again, we choose $\sigma=\mathrm{sign}$. We now consider only values of $N$ divisible by $K$, and divide the inputs into groups of $N/K$, each of which is fed to one of the $K$ hidden units; the final output is then decided by majority voting, as:
	
	\begin{equation}
		\label{eq:committee}
		y_{\text{out}} \equiv \mathrm{sign}\left(\sum_{h=1}^K \text{sign} \left( \frac{1}{\sqrt{N/K}} \sum_{i=\left(h-1\right)\frac{N}{K}+1}^{h \frac{N}{K}} w_i\tilde{\xi}_i  \right)\right)
	\end{equation}
	
	Beside the architecture, one major difference with the perceptron case is that here the weights $\boldsymbol{w}$ are assumed to be continuous. Due to the $\mathrm{sign}$ activation function, each unit is invariant to scaling, and thus we normalize the weights of the units by fixing their norms to $1$.
	
	We consider two learning algorithms for this architecture. The first one is a version of focusing-BP (fBP) that operates with continuous weights~\cite{baldassi2020shaping}. The implementation exploits the central limit theorem and thus it only works well for relatively large values of $N/K$; furthermore, even in the large $N$ limit, it is only approximately correct on the tree-committee machine architecture. Despite this, in practice it produces excellent results.
	
	The second algorithm is Stochastic Gradient Descent with cross-entropy loss. Following ref.~\cite{baldassi2020shaping}, we substituted the (non-differentiable) units' activation function in eq.~\ref{eq:committee}, $\mathrm{sign}\left(\Delta\right)$, with $\tanh\left(\beta \Delta\right)$. The new parameter $\beta$ can be regarded as taking the role of the norm of the unit's weights, since we keep the weights normalized at each step. We explicitly schedule this parameter, letting it start from a small value and making it diverge during the training, thereby recovering the original $\mathrm{sign}$ activation at the end~\footnote{Note that the divergence of the norms would occur naturally anyway in standard SGD with the cross-entropy loss.}. Analogously, we also schedule a parameter $\gamma$ that has the role of the norm of the (fixed) weights in the second layer, and that we can simply plug in the cross-entropy (see the Materials and Methods).
	
	\begin{figure}
		\begin{centering}
			\includegraphics[width=\columnwidth]{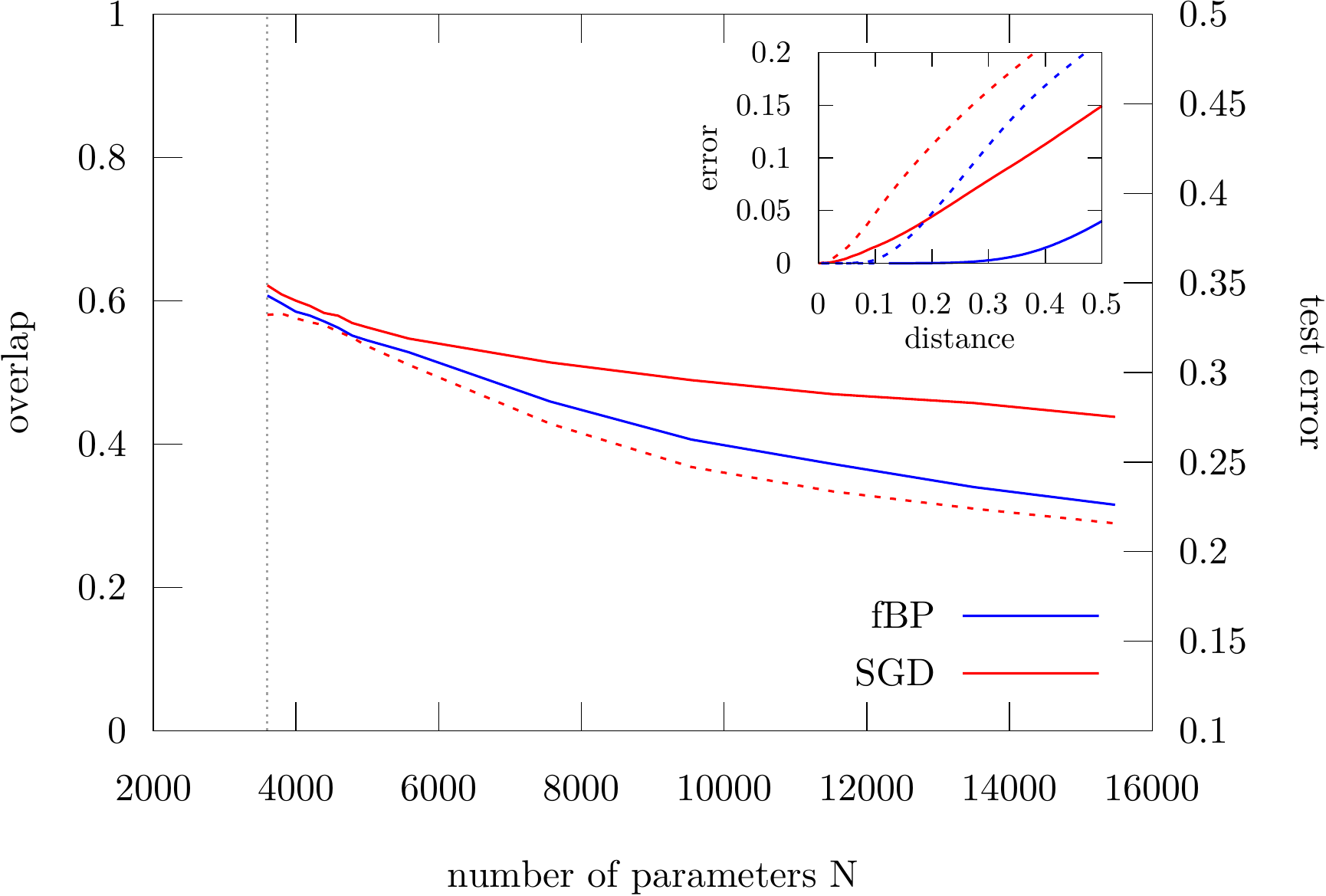}
		\end{centering}
		\caption{Results for the overparameterized continuous tree-like committee machine. Tests performed with $D=2001$, $\alpha_T=5$, $K=9$. All points are averages over 5 samples, with 5 independent runs per sample for SGD. Dashed red line: mean overlap between two SGD solutions on the same sample, as a function of the number of parameters $N$. Solid lines: test error for BP and SGD. The vertical dashed grey line at $N=3600$ ($\alpha^{-1} \approx 0.36$) denotes the algorithmic threshold below which the algorithms solve fewer than 50\% of the samples. The overlaps are still far from $1$ at this point. Inset: Local energy profiles, i.e. average train error as a function of the Euclidean distance from a solution. The dashed lines are measured at $N=3996$ ($\alpha^{-1}\approx 0.4$), the solid lines at $N\approx 13500$ ($\alpha^{-1}\approx 1.35$).}
		\label{Fig:committee} 
	\end{figure}
	
	Here, we report the result of tests performed on a committee machine with $K=9$ hidden units, trained on $P=10005$ patterns produced in $D=2001$ dimensions, thus at a fairly large $\alpha_T=P/D=5$, while varying the degree of overparameterization $N=P \alpha^{-1}$ (see the Materials and Methods for the details of the settings used for the training). Our results are reported in Fig.~\ref{Fig:committee}. We found that both fBP and SGD fail to find a solution below $\alpha^{-1} \approx 0.36$, which we thus take to be a plausible estimate for the algorithmic threshold $\alpha_{\mathrm{LE}}^{-1}$ where the phase of the robust solutions presumably changes. This is corroborated by the study of the overlaps: for any given training set, the SGD algorithm finds different solutions when started from different random initial conditions (this is not true for fBP due to its deterministic nature). We measured the average overlap (cosine similarity) between the solutions, $\langle \frac{w^a \cdot w^b}{N} \rangle$, and found that when reducing $\alpha^{-1}$ the overlap grows, but it does not tend to $1$ as $\alpha$ tends to $\alpha_{\mathrm{LE}}$, which one would expect if the solutions shrank to a single interpolation point like in convex models. The generalization error behaves as expected, decreasing monotonically with $N$; SGD is slightly worse than fBP in this regard. We also measured the flatness of the minima found by the algorithm by plotting the "average local energy"\footnote{Notice that the local energy of a configuration is highly correlated with its local entropy, see e.g.~\cite{Pittorino_2021}}, i.e. the average training error profile of the landscape surrounding each solution, as a function of the distance. This can be estimated straightforwardly and robustly by randomly perturbing the weights, with a varying degree of multiplicative noise (the weights are still renormalized after the perturbation). In Fig.~\ref{Fig:committee}, we show two sets of curves, one at $\alpha^{-1} \approx 0.4$, close to $\alpha_{\mathrm{LE}}^{-1}$, and one at $\alpha^{-1} \approx 1.35$, at the opposite end. As expected, close to the threshold the minima are generally sharper, but in all cases both SGD and fBP have flat profiles for small distances, reflecting the fact that both algorithms are inherently biased towards wide flat minima \cite{baldassi2020shaping}. The bias is stronger for the fBP algorithm, which was explicitly designed for this purpose, and its profiles are indeed flatter.
	
	Overall, all the phenomenological features that we could measure on this model are compatible with the theoretical analysis of the previous section on the binary perceptron, despite the more complex architecture and the continuous weights, even in the context of gradient-based learning.
	
	
	{\em Comparing solutions in Deep architectures: removing symmetries}
	
	When discussing the space of configurations of standard multi-layer architectures, we need to be more careful compared to the simple models discussed so far, due to the presence of additional symmetries~\cite{pittorino2022deep}.
	
	First, the ReLU activation function that is commonly used in deep learning models has the property that $\mathrm{ReLU}\left(a x\right) = a\ \mathrm{ReLU}\left(x\right)$, which implies that if we scale all the input weights of a hidden unit by a factor $a^{-1}$ and all its output weights by a factor $a$ the network's output will be unaffected. By setting the factor $a$ to the norm of the input weights, one can normalize a hidden unit by simply "pushing up" its norm to the next layer. Furthermore, when a network is used for classification tasks, the output label is determined by an $\mathrm{argmax}$ operation, which is invariant to scaling. Thus, normalizing the last layer too is possible without affecting the classification properties of the network. In the full configuration space, each neural network has infinitely many parameter representations, and the error rate landscape has some trivially null directions. This issue can be avoided by normalization, which can be performed simply by starting from the first layer and moving up, as described above.
	
	There is also a second, discrete symmetry, since networks are invariant to permutations of the units inside any hidden layer. If we failed to take into account this, we could measure a non-zero distance between networks which are just permuted versions of each other and thus functionally equivalent. One natural way to break this symmetry is to normalize and align the networks before comparing their weights. In our tests, we adopted again a sequential approach for aligning two given networks. Starting from the first hidden layer, we find the permutation of the second networks' units that minimizes the distance between the weights of the two networks for that layer\footnote{This can be accomplished by a matching algorithm, which is $O(H^3)$ if $H$ is the number of hidden units of the layer. In practical terms, it is typically much quicker than the training.}, apply it, and proceed to the following layer.
	
	In the following paragraphs, we present experiments on deep architectures that have both these symmetries. We use standard techniques to train them, and thus do not explicitly keep the norms and permutations under control. We do however normalize and align them when we compare two solutions, either to compare their error rates or to measure their distance.
	
	{\em Multi-layer neural network}
	\begin{figure}
		\begin{centering}
			\includegraphics[width=1\columnwidth]{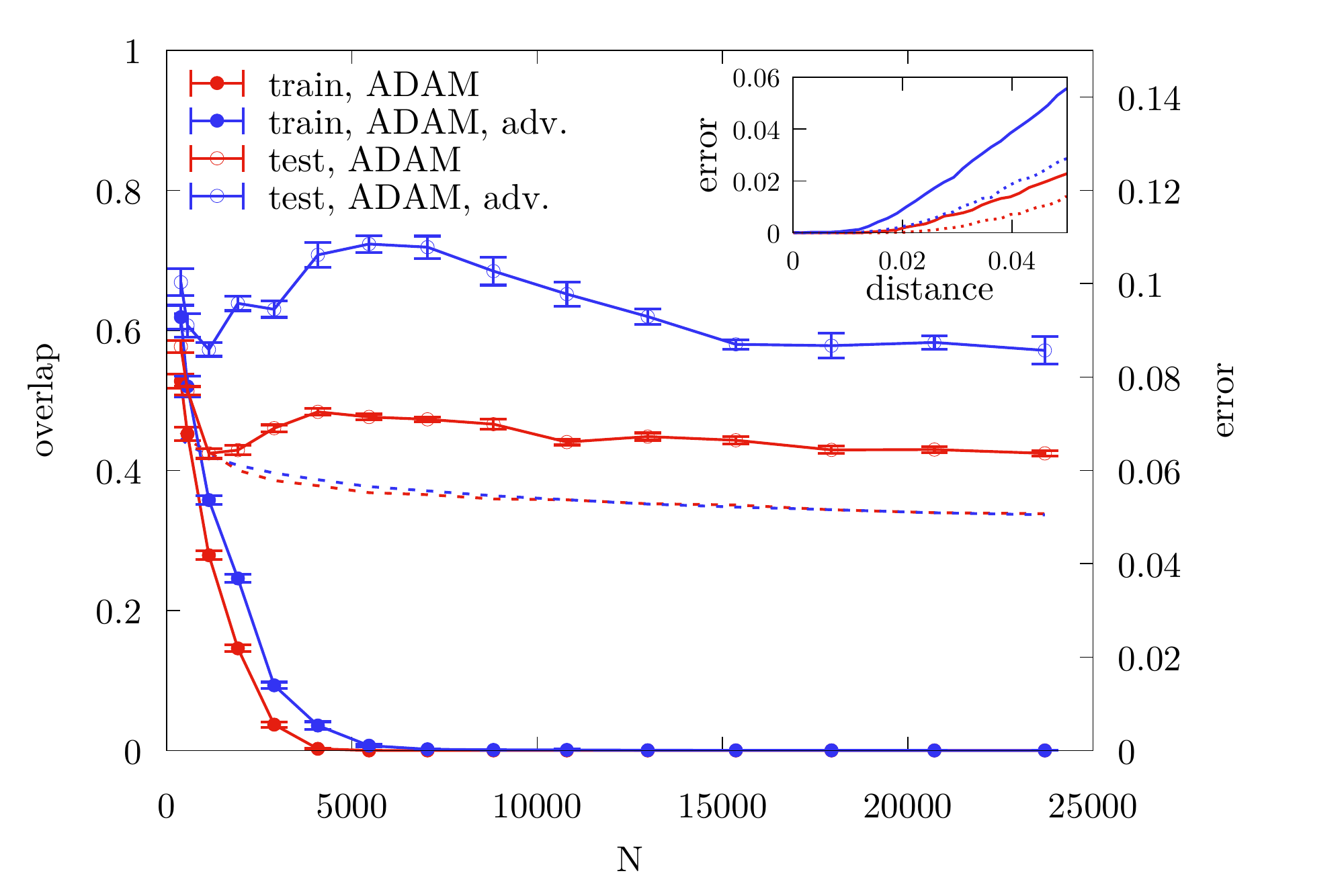}
		\end{centering}
		\caption{Train (full points) and test (empty points) error as a function of the network parameters $N$ for a fully-connected network with $5$ hidden layers. All points are averages over $10$ independent runs. The network is trained using full batch gradient descent with ADAM optimization, with random orthogonal initialization (red curves) and adversarial initialization (blue curves). When the algorithms start finding zero errors solutions, the mean overlap between solutions is far below $1$ (dashed lines). Inset: local energy profiles for both algorithms at two different values of the overparameterization ($N \simeq 9 \cdot 10^{3}$ (full lines) and $N \simeq 2 \cdot 10^{4}$ (dashed lines)).}
		\label{Fig:fully_connected} 
	\end{figure}
	We studied a simple fully-connected multi-layer perceptron inspired by ref.~\cite{geiger2020scaling}. The network has a fixed number $H$ of hidden layers ($H=5$ in the numerical experiments) whose width is varied in order to increase the number of model parameters. 
	The model is required to perform a binary classification task on the parity of digits of $10000$ MNIST images, using as inputs only the first $10$ principal components of each image.
	We trained the model using full batch gradient descent with ADAM optimization, 
	both with random orthogonal initialization~\cite{saxe2013exact} and with adversarial  initialization~\cite{liu2020bad}. 
	The results are reported in Fig.~\ref{Fig:fully_connected} where it can be seen that in general different optimization schemes lead to different generalization error plateaus and different algorithmic thresholds. When the model starts to fit the training set, the solution is not unique and as a consequence the mean overlap between independent instances is lower than $1$. The inset shows that the solutions are indeed robust to noise perturbations even in the proximity of the algorithmic threshold, and they become flatter as the overpameterization increases.
	
\begin{figure}
		\begin{centering}
			\includegraphics[width=\columnwidth]{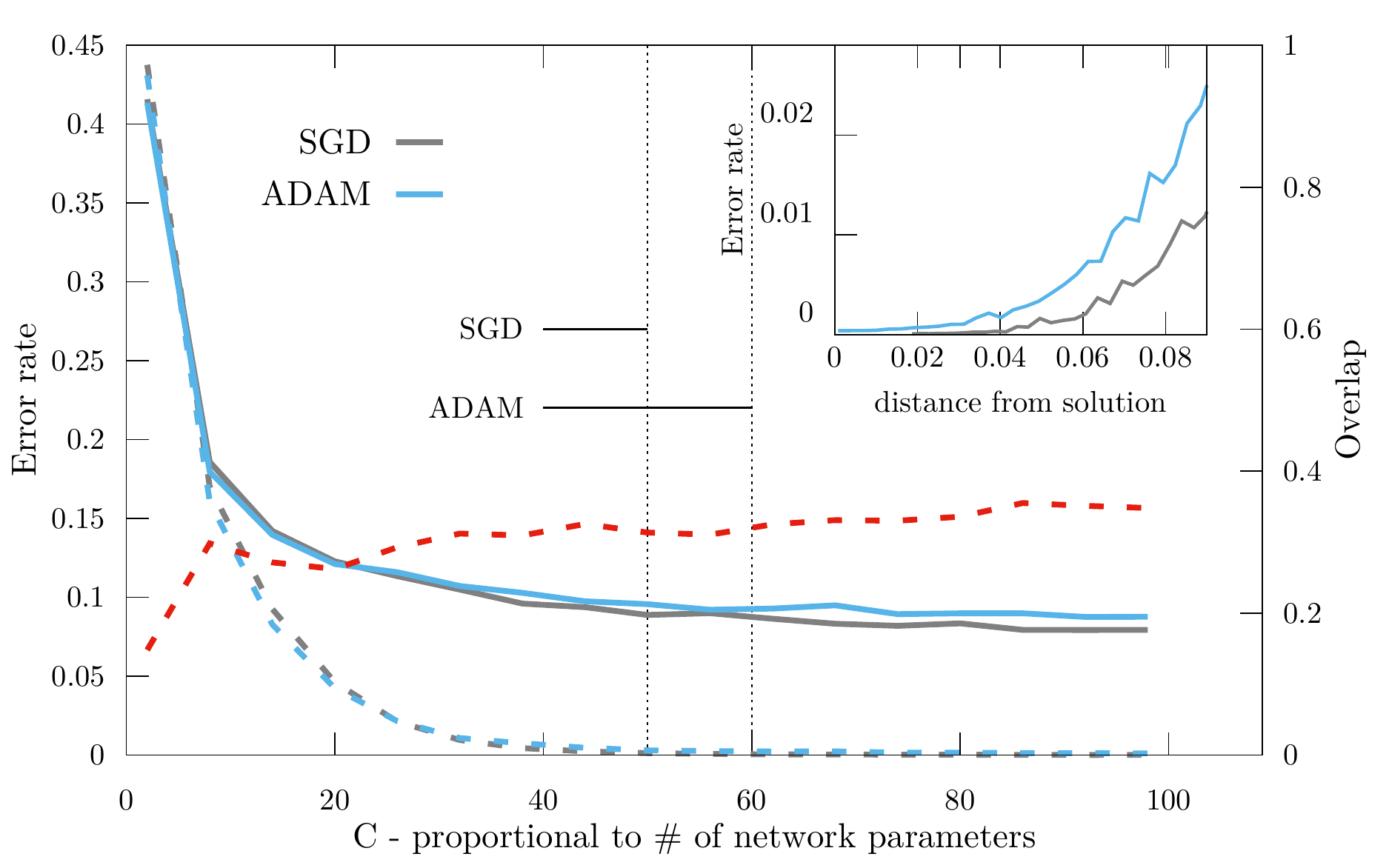}
		\end{centering}
		\caption{Train (dashed) and test (solid) errors of solutions obtained with SGD and ADAM (gray and blue lines respectively) as a function of $C$. Each point is the average on 5 independent samples. 
			Red dotted line: average overlap between 3 different pairs of solutions (ADAM). Inset: local energy as a function of the distance from solutions (gray: SGD with $C = 54$; blue: ADAM with $C = 64$).}
		\label{Fig:cnn} 
	\end{figure}

	{\em Convolutional networks} As a second representative case of deep architectures we analyze a 5-layer NN, with 4 convolutional layers followed by a fully connected one, as in ref.~\cite{nakkiran2021deep}.
	After each 2d convolution, a batch normalization is performed before applying a ReLU nonlinearity.
	The overparameterization in this model is adjusted via a parameter $C$: in layer $\ell$ there are $2^\ell \cdot C$, 3x3 convolutional filters. 
	This CNN is trained on CIFAR10 for 200 epochs, using two different learning algorithms: ADAM with momentum and SGD with a learning rate $\eta = 10^{-2}$. For real datasets like the one we considered, while it is relatively easy to achieve a very low training error, getting to precisely 0 error requires a disproportionate amount of additional computation. For this reason, we consider as solutions all configurations that misclassify at most 1 pattern (<$0.0017\%$ training error), and estimate the algorithmic LE threshold according to this criterion. In Fig.~\ref{Fig:cnn}, train (dashed line) and test (solid line) errors are shown for both optimizers. SGD and ADAM begin to fit the training set data at $C = 50$ and $C = 60$ respectively, while the generalization error is monotonically decreasing with the number of network parameters. It is worth noticing that these architectures work in a relatively lazy training regime: on one hand, the first layer is less affected by the training, while the following layers change progressively more; on the other hand, the distance of the trained configurations from their initial conditions does not drop to zero as the number of parameters diverges (see details in the  SI).
	Consistently with the analytical results, and similarly to other networks we have studied, we find that the learning algorithms, namely SGD and ADAM, find different solutions depending on initial conditions. In particular, even when we get close to the algorithmic threshold, we observe that the overlaps, in contrast to what happens at the interpolation threshold in convex networks, do not tend to their maximum value of one (red curve in Fig.~\ref{Fig:cnn}).
	The solutions found by SGD and ADAM show similar geometric properties, i.e., they belong to flat regions of the training error landscape. This can be seen in the inset of Fig.~\ref{Fig:cnn}, where we display the results of the numerical analysis of the average (local energy) landscape around a given solution. Like for the other models, this was measured by random sampling, perturbing the solutions with multiplicative noise (see the SI for details). The first derivative at short distances is essentially zero.

	\section{Concluding remarks}
	
	Our results characterize the interplay between overparameterization and nonconvexity in neural networks learning data generated by structurally different networks randomly chosen from a natural distribution.
	In particular, we identify a new phenomenon, namely the existence of a phase transition driven by the appearance of solution sets that are statistically atypical but which seem to be the ones targeted by learning algorithms.
	For the same systems we are able to derive the generalization error of different types of solutions and predict how to optimize the Bayesian error.
	The analytical techniques also suggest a number of numerical verifications that can be done on deep networks and for different learning algorithms. The consistency of the results is very good, suggesting that the scenario identified in the analytically tractable models, i.e., the essential role played by highly entropic atypical solutions, may in fact be general.
	
	There are several natural future directions.  On the one hand, in-depth numerical studies of large deep networks should be conducted, and algorithms should be further optimized building on the information derived from the structure of solutions. Most algorithms already do this as a result of the tuning process that has been put in practice during the last decade. Still further progress appear to be possibile, and
	some steps in this direction have already been taken.
	On the other hand, it would be important to corroborate our results with rigorous bounds, for more general data distributions, in order to reach a more complete mathematical theory  for learning in non-convex overparameterized systems.
	Finally, a theoretical confirmation that dynamics of a broad class of algorithms is indeed attracted to these structures would be of great interest (an analysis that shows that SGD is biased towards flat minima can be found in~\cite{FengYuhai}, but some of the assumptions are justified phenomenologically).
	
	From a physics and modeling perspective, it seems to us that having identified that atypical states play a key role in learning processes opens the way toward a fertile connection between out-of-equilibrium physics and modeling of learning systems.
	Indeed, such states are inherently atypical with respect to algorithmic dynamics that tend to sample energy with a Gibbs measure and where the basic energy function, or loss function, is defined directly on the data as the number of errors.

	\section*{Acknowledgements} 
	We gratefully thank Fabrizio Pittorino for sharing with us his code implementing matching of CNN.

	\noindent \bibliographystyle{unsrturl}
	\bibliography{references}
	
	\onecolumngrid 
	\appendix

	\section{Some preliminary definitions}
	
	We denote by $w_k^T$ the weights of a teacher that lives in a $D$-dimensional space ($k = 1, \dots, D$). The teacher assigns to i.i.d. standard normal random input variables $\xi_k^\mu$ (with $\mu = 1, \dots, P$) a label, via
	\begin{equation}
		\label{eq::label}
		y^\mu = \text{sign} \left( u^\mu \right) = \text{sign} \left( \frac{1}{\sqrt{D}} \sum_{k=1}^{D} w_k^T \xi_k^\mu \right) \,.
	\end{equation}
	The student sees a projection of the patterns in an $N$ dimensional space plus a non linearity $\sigma$. The dimensionality of the space $N$ can either be higher or lower than the true dimension $D$. The projection is therefore identified by an $D \times N$ feature matrix $F_{ki}$, so that
	\begin{equation}
		\tilde \xi_i^\mu = \sigma\left( \frac{1}{\sqrt{D}} \sum_{k=1}^{D} F_{ki} \xi_k^\mu \right)
	\end{equation}
	The student then classifies the projected patterns with its weights as
	\begin{equation}
		\label{eq::prediction}
		\hat y^\mu = \text{sign} \left( \lambda^\mu \right) = \text{sign} \left( \frac{1}{\sqrt{N}}  \sum_{i=1}^{N} w_i \tilde \xi_i^\mu \right) = \text{sign} \left( \frac{1}{\sqrt{N}}  \sum_{i=1}^{N} w_i \, \sigma\left( \frac{1}{\sqrt{D}} \sum_{k=1}^{D} F_{ki} \xi_k^\mu \right) \right)
	\end{equation} 
	Notice that we have denoted with $u^\mu$ and $\lambda^\mu$  the preactivation of pattern $\mu$ for the teacher and the student respectively. The only assumptions we make on the feature matrix are
	\begin{subequations}
		\begin{align}
			\frac{1}{D}\sum_{k=1}^{D} F_{ki}^2 &= 1\,, \qquad\;\,\quad \forall i \label{eq::assumption1} \\
			\frac{1}{\sqrt{D}}\sum_{k=1}^{D} F_{ki} F_{kj} &= \mathcal{O}(1)\,, \qquad \forall i \ne j \label{eq::assumption2} \\
			s_{k_1, \dots, k_s}^{a_1, \dots, a_n} \equiv \frac{1}{\sqrt{N}}\sum_{i=1}^{N} w_i^{a_1} \dots w_i^{a_n} F_{k_{1}i} \dots F_{k_{s}i} &= \mathcal{O}(1)\,, \qquad \forall n, s\ge 1 \label{eq::assumption3}
		\end{align}
	\end{subequations}
	In particular the second requirement tells us that two different sub-perceptrons are almost uncorrelated. 
	In general, choosing the entries of the matrix $F_{ki}$ to be random i.i.d. standard Gaussian random variables or random i.i.d. binary ones will do the job.
	The partition function of the model is, therefore,
	\begin{equation}
		Z = \int \prod_i dw_i \, P_{w}(\boldsymbol w) \, e^{-\beta \sum_{\mu=1}^{P} \ell\left( - y^\mu \lambda^\mu \right)}
	\end{equation}
	where $\ell(\cdot)$ is a loss function per pattern and $P_{w}(\boldsymbol w)$ represents the prior over the weights and identifies their space of definition. We will adopt a similar probability density $P_{w^T}(\boldsymbol w^T)$ for the teacher weights. In the following we will study analytically the non convex ``binary'' problem, where both the teacher and the student are $\pm 1$; the ``spherical'', problem where the weights live on the sphere, is convex and has been already studied in the literature~\cite{gerace2020generalisation}. 
	In this paper we will focus on the loss that simply counts the number of patterns in the training set whose stability $y^\mu \lambda^\mu$ is larger than a given positive margin $\kappa$
	\begin{equation}
		\ell_{NE}(-x; \kappa) = \Theta\left(-x + \kappa \right)
	\end{equation}
	where $\Theta(\cdot)$ is the Heaviside step function: $\Theta\left(x\right) = 1$ if $x>0$ and zero otherwise. For $\kappa = 0$ this loss reduces to the one that counts the number of errors; with a slight abuse of notation we call it number of errors loss even if the margin is non-zero. 
	In the following we will be interested in the binary weights case; we will compute the free entropy of solution in the thermodynamic limit
	\begin{equation}
		\label{eq::thermodynamic_limit}
		N, \, D, \, P \to \infty \qquad \text{fixing} \; \alpha \equiv \frac{P}{N} \;\; \text{and} \;\; \alpha_T = \frac{P}{D} \equiv \frac{\alpha}{\alpha_D}
		\,.
	\end{equation}
	We will also limit ourselves to the case of random i.i.d. standard Gaussian features $F_{ki}$.
	
	\section{Replica Method}
	\noindent Introducing replicas we get
	\begin{equation}
		Z^n = \int \prod_{i a}  dw_i^a \, P_{w}(\boldsymbol w^a) \, e^{-\beta \sum_{\mu=1}^{P} \sum_{a=1}^n \Theta\left( - y^\mu \lambda^{\mu a} + \kappa \right)}
	\end{equation}
	We now enforce the definitions of the preactivations of the teacher and the student by using delta functions
	\begin{multline}
		\label{eq::replicated_part_func}
		\mathbb{E}_{\left\{\boldsymbol{\xi}^\mu\right\} } [Z^n] = \int \prod_{i a}  dw_i^a \, P_{W}(\boldsymbol w^a)  \int \prod_{\mu} d u^\mu \prod_{\mu a} d \lambda^\mu_a\, e^{-\beta \sum_{\mu=1}^{P} \sum_{a=1}^n \Theta\left( - \text{sign}(u^\mu) \lambda^\mu_a + \kappa \right)} \\ 
		\times \mathbb{E}_{\left\{\boldsymbol{\xi}^\mu\right\}}\left[ \prod_{\mu}\delta\left( u^\mu - \frac{1}{\sqrt{D}} \sum_{k=1}^{D} w_k^T \xi_k^\mu \right) \prod_{\mu a} \delta \left( \lambda^\mu_a - \frac{1}{\sqrt{N}}  \sum_{i=1}^{N} w_i^a \, \sigma\left( \frac{1}{\sqrt{D}} \sum_{k=1}^{D} F_{ki} \xi_k^\mu \right)  \right) \right] \,.
	\end{multline}
	Notice that if the distributions of patterns and features are symmetric, we can perform the gauge transformation $\xi_k^\mu \to w_k^T \xi_k^\mu$, $F_{ki} \to w_k^T F_{ki}$, so that we can consider $w_k^T = 1$, $\forall k$, without loss of generality.
	
	\subsection{Average over the disorder: Gaussian equivalence theorem} \label{sec::CLT}
	When $\sigma$ is linear, the average can be easily computed by using the integral representation of the delta function. When $\sigma$ is non-linear the computation of the average is more involved. We can, however, compute the moments of the variables $u^\mu$ and $\lambda^\mu_a$ as defined in equation~\eqref{eq::replicated_part_func}. One can show that in the thermodynamic limit~\eqref{eq::thermodynamic_limit}, the moments are those of a multivariate Gaussian random variable~\cite{Mei2019,Goldt2020}. This result is equivalent to the central limit theorem that is easy to derive in the classical models without (random) feature projections~\cite{gardner1988The,gardner1988optimal,Gardner_1989,Gyordyi1990} and has been renamed as ``Gaussian equivalence theorem''. In the following we will compute explicitly the first two moments of the random variables $u^\mu$ and $\lambda^\mu_a$, and we will refer to~\cite{Goldt2020} for the computation of the fourth moment. 
	
	We start defining the following useful quantities
	\begin{subequations}
		\label{eq:effective_nonlinearity}
		\begin{align}
			\mu_0 &= \int Dz \, \sigma(z) \\
			\mu_1 &= \int Dz \, z \, \sigma(z) = \int Dz \, \sigma'(z) \\
			\mu_2 &= \int Dz \, \sigma^2(z) \\
			\mu_\star^2 &= \mu_2 - \mu_1^2 - \mu_0^2
		\end{align}
	\end{subequations}
	where $Dz \equiv \frac{e^{-z^2/2}}{\sqrt{2\pi}} dz$.
	The mean of $u^\mu$ is trivial
	\begin{equation}
		\mathbb{E}_{\boldsymbol{\xi}} \left[ u^\mu \right] = 0
	\end{equation} 
	whereas that of $\lambda_a^\mu$ is
	\begin{equation}
		\begin{split}
			\mathbb{E}_{\boldsymbol{\xi}^\mu} \left[ \lambda_a^\mu \right] &= \int\prod_{i=1}^N\frac{dv_{i}^{\mu}d\hat{v}_{i}^{\mu}}{2\pi}e^{i\sum_{i}\hat{v}_{i}^{\mu}v_{i}^{\mu}}\left[\frac{1}{\sqrt{N}}\sum_{i=1}^{N}w_i^{a}\sigma\left(v_{i}^{\mu}\right)\right]\prod_{k}\mathbb{E}_{\xi_{k}^{\mu}}e^{-i\frac{\xi_{k}^{\mu}}{\sqrt{D}}\left(\sum_{i}\hat{v}_{i}^{\mu}F_{ki}\right)} \\
			&=\int\prod_{i=1}^N\frac{dv_{i}^{\mu}d\hat{v}_{i}^{\mu}}{2\pi}e^{i\sum_{i}\hat{v}_{i}^{\mu}v_{i}^{\mu}}\left[\frac{1}{\sqrt{N}}\sum_{i=1}^{N}w_i^{a}\sigma\left(v_{i}^{\mu}\right)\right]e^{-\frac{1}{2}\sum_{ij}\left(\frac{1}{D}\sum_{k}F_{ki}F_{kj}\right)\hat{v}_{i}^{\mu}\hat{v}_{j}^{\mu}}
		\end{split}
	\end{equation}
	We now use~\eqref{eq::assumption1} and~\eqref{eq::assumption2} obtaining
	\begin{equation}
		\mathbb{E}_{\boldsymbol{\xi}^\mu} \left[ \lambda_a^\mu \right] = \frac{1}{\sqrt{N}}\sum_{i=1}^{N}w_i^{a}\int \frac{dv_{\mu}d\hat{v}_{\mu}}{2\pi}e^{i \hat{v}_{\mu}v_{\mu}} \sigma\left(v_{\mu}\right) e^{-\frac{\hat{v}_{\mu}^2}{2}} = \mu_0 \frac{1}{\sqrt{N}}\sum_{i}w_i^{a} \,.
	\end{equation}
	The second moment of $u^\mu$ is
	\begin{equation}
		\mathbb{E}_{\boldsymbol{\xi}^\mu}\left[u_{\mu}^{2}\right]=\frac{1}{D}\sum_{k=1}^{D}\left(w^T_{k}\right)^2 = 1
	\end{equation}
	whereas that of $\lambda_a^\mu$ is 
	\begin{equation}
		\begin{split}
			\mathbb{E}_{\boldsymbol{\xi}^\mu}\left[\lambda_{\mu}^{a}\lambda_{\mu}^{b}\right] 
			& =\int\prod_{i=1}^N\frac{dv_{i}^{\mu}d\hat{v}_{i}^{\mu}}{2\pi}e^{i\sum_{i}\hat{v}_{i}^{\mu}v_{i}^{\mu}}\left[ \frac{1}{N} \sum_{ij}w_i^{a}w_j^{b}\sigma\left(v_{i}^{\mu}\right)\sigma\left(v_{j}^{\mu}\right)\right]e^{-\frac{1}{2}\sum_{ij}\left(\frac{1}{D}\sum_{k}F_{ki}F_{kj}\right)\hat{v}_{i}^{\mu}\hat{v}_{j}^{\mu}} \,.\\
		\end{split}
	\end{equation}
	Now we split $i = j$ and $i \ne j$ contributions. Because of~\eqref{eq::assumption2}, $\frac{1}{D}\sum_{k}F_{ki}F_{kj}$ with $i \ne j$ is of order $1/\sqrt{D}$; we can therefore expand the exponential. We have
	\begin{equation}
		\begin{split}
			e^{i \hat{v}_{i}^{\mu}v_{i}^{\mu} + i \hat{v}_{j}^{\mu}v_{j}^{\mu} -\left(\frac{1}{D}\sum_{k}F_{ki}F_{kj}\right)\hat{v}_{i}^{\mu}\hat{v}_{j}^{\mu}} &\simeq e^{i \hat{v}_{i}^{\mu}v_{i}^{\mu} + i \hat{v}_{j}^{\mu}v_{j}^{\mu}}  \left[ 1 -\frac{1}{2}\left(\frac{1}{D}\sum_{k}F_{ki}F_{kj}\right)\hat{v}_{i}^{\mu}\hat{v}_{j}^{\mu} \right]\\
			&= \left[ 1 + \left(\frac{1}{D}\sum_{k}F_{ki}F_{kj}\right) \frac{d}{d v_i^\mu} \frac{d}{d v_j^\mu} \right] e^{i \hat{v}_{i}^{\mu}v_{i}^{\mu} + i \hat{v}_{j}^{\mu}v_{j}^{\mu}}
		\end{split}
	\end{equation}
	so that, performing the integrals we have
	\begin{equation}
		\begin{split}
			\mathbb{E}_{\boldsymbol{\xi}^\mu}\left[\lambda_{\mu}^{a}\lambda_{\mu}^{b}\right] 
			&= \mu_2 \frac{1}{N} \sum_{i}w_i^{a}w_i^{b} + \frac{1}{N}\sum_{i\ne j}w_i^{a}w_j^{b} \int D v_{i}^{\mu} Dv_{j}^{\mu} \sigma\left(v_{i}^{\mu}\right)\sigma\left(v_{j}^{\mu}\right) \left[ 1 + \left(\frac{1}{D}\sum_{k}F_{ki}F_{kj}\right) v_i^\mu v_j^\mu \right]\\
			&= \left(\mu_2 - \mu_1^2 - \mu_0^2\right) \frac{1}{N} \sum_{i}w_i^{a} w_i^b + \frac{\mu_0^2}{N} \sum_i w_i^a \sum_j w_j^b  + \mu_1^2  \frac{1}{D} \sum_{k=1}^{D} s_k^a s_k^b
		\end{split}
	\end{equation} 
	where we have defined the ``projected'' weights $s_k^a$ as
	\begin{equation}
		\label{eq::projected_weights}
		s_k^a \equiv \frac{1}{\sqrt{N}} \sum_{i=1}^{N} F_{ki} w_i^a \,.
	\end{equation}
	The covariance is therefore
	\begin{equation}
		\mathbb{E}_{\boldsymbol{\xi}^\mu}\left[\lambda_{\mu}^{a}\lambda_{\mu}^{b}\right] - \mathbb{E}_{\boldsymbol{\xi}^\mu}\left[\lambda_{\mu}^{a}\right] \mathbb{E}_{\boldsymbol{\xi}^\mu}\left[\lambda_{\mu}^{b} \right] = \mu_\star^2 \frac{1}{N} \sum_{i}w_i^{a} w_i^b + \mu_1^2  \frac{1}{D} \sum_{k=1}^{D} s_k^a s_k^b \,.
	\end{equation}
	We also define the ``projected'' teacher weights as
	\begin{equation}
		s_i^T \equiv \frac{1}{D} \sum_{k=1}^D F_{ki} w_k^T \,.
	\end{equation}
	Using again assumptions~\eqref{eq::assumption1} and~\eqref{eq::assumption2} we get for the cross term
	\begin{equation}
		\begin{split}
			\mathbb{E}_{\boldsymbol{\xi}^\mu}\left[u_{\mu}\lambda_{\mu}^{a}\right] 
			&=\int\prod_{i}\frac{dv_{i}^{\mu}d\hat{v}_{i}^{\mu}}{2\pi}\frac{du_{\mu}d\hat{u}_{\mu}}{2\pi}e^{i\sum_{i}\hat{v}_{i}^{\mu}v_{i}^{\mu}+i\hat{u}_{\mu}u_{\mu}}\left[\frac{u_{\mu}}{\sqrt{N}}\sum_{i}w_i^{a}\sigma\left(v_{i}^{\mu}\right)\right]e^{-\frac{\hat{u}_{\mu}^{2}}{2}-\frac{1}{2}\sum_{ij}\left(\frac{1}{D}\sum_{k}F_{ki}F_{kj}\right)\hat{v}_{i}^{\mu}\hat{v}_{j}^{\mu}-\hat{u}_{\mu}\sum_{i}\hat{v}_{i}^{\mu} s_i^T} \\
			&= \frac{1}{\sqrt{N}}\sum_{i}w_i^{a} \int \frac{dv_{\mu}d\hat{v}_{\mu}}{2\pi}\frac{du_{\mu}d\hat{u}_{\mu}}{2\pi}e^{i\hat{v}_{\mu}v_{\mu}+i\hat{u}_{\mu}u_{\mu}}\left[u_{\mu} \sigma\left(v_{\mu}\right)\right]e^{-\frac{\hat{u}_{\mu}^{2}}{2} - \frac{\hat v_\mu^2}{2} -\hat{u}_{\mu}\hat{v}_{\mu} s_i^T}\\
			&= 
			\mu_1 \frac{1}{\sqrt{N}} \sum_i w_i^a  s_i^T = \mu_1 \frac{1}{D} \sum_{k=1}^{D} s_k^a w_k^T \,.
		\end{split}
	\end{equation}
	The distribution of random variables $u^\mu$ and $\lambda_a^\mu$ therefore can be written as a multivariate Gaussian.
	The final result reads
	\begin{equation}
		P\left(u^\mu, \left\{ \lambda_a^\mu \right\}  \right)  = \frac{1}{\sqrt{2\pi \det \Sigma}} e^{-\frac{1}{2} \sum_{\gamma, \delta = 0}^n \left(\Upsilon_\gamma^\mu - \rho_\gamma \right) \left( \Sigma^{-1} \right)_{\gamma \delta} \left( \Upsilon_\delta^\mu - \rho_\delta \right)}
	\end{equation}
	where $\Upsilon_0^\mu \equiv u^\mu$ and $\Upsilon_a^\mu \equiv \lambda_a^\mu$, $\forall a = 1, \dots, n$. The mean vector is $\rho_0 = 0$ and $\rho_a = \frac{\mu_0}{\sqrt{N}}\sum_{i}w_i^{a}$ for $a = 1, \dots, n$; the covariance is
	\begin{equation}
		\Sigma \equiv \begin{pmatrix} 
			1 & M_a \\
			M_a & Q_{ab} 
		\end{pmatrix}
	\end{equation}
	where
	\begin{subequations}
		\label{eq::order_params_capital}
		\begin{align}
			\label{eq::order_params_capital_a}
			M_a &=  \mu_1 \frac{1}{D} \sum_{k=1}^{D} s_k^a w_k^T \equiv \mu_1 r_a\\
			\label{eq::order_params_capital_b}
			Q_{ab} &= \mu_\star^2 \frac{1}{N} \sum_{i=1}^Nw_i^{a} w_i^b + \mu_1^2  \frac{1}{D} \sum_{k=1}^{D} s_k^a s_k^b \equiv \mu_\star^2 q_{ab} + \mu_1^2  p_{ab}
		\end{align}
	\end{subequations}
	The average over the replicated partition function therefore takes the form
	\begin{equation}
		\label{eq::part_func_after_average1}
		\begin{split}
			\mathbb{E}_{\left\{\boldsymbol{\xi}^\mu\right\}} [ Z^n] &= \int \prod_{i a}  dw_i^a \, P_{w}(\boldsymbol w^a)  \int \prod_{\mu} d u^\mu \prod_{\mu a} d \lambda^\mu_a\, e^{-\beta \sum_{\mu=1}^{P} \sum_{a=1}^n \Theta\left( - \text{sign}(u^\mu) \lambda^\mu_a + \kappa \right)} P\left(u^\mu, \left\{ \lambda_a^\mu \right\}  \right)\\
		\end{split}
	\end{equation}
	or, equivalently
	\begin{equation}
		\label{eq::part_func_after_average2}
		\begin{split}
			\mathbb{E}_{\left\{\boldsymbol{\xi}^\mu\right\}} [Z^n] &= \int \prod_{i a}  dw_i^a \, P_{w}(\boldsymbol w^a) \int \prod_{\mu} \frac{d u^\mu d \hat u^\mu}{2\pi} \prod_{\mu a} \frac{d \lambda^\mu_a d \hat \lambda^\mu_a}{2\pi}\, e^{-\beta \sum_{\mu=1}^{P} \sum_{a=1}^n \Theta\left( - \text{sign}(u^\mu) \lambda^\mu_a + \kappa \right)} \\
			& \times \prod_{\mu} e^{i u^\mu \hat u^\mu + i \sum_a \left( \lambda_a^\mu - \rho_a\right) \hat \lambda_a^\mu - \frac{(\hat u^\mu)^2}{2} -  \frac{1}{2} \sum_{ab} Q_{ab} \hat \lambda^\mu_a \hat \lambda^\mu_b -  \sum_a M_a \hat u^\mu \hat \lambda^\mu_a} \,.
		\end{split}
	\end{equation}
	Therefore the analytical expression of the average over patterns is similar to the one of the non-overparameterized teacher-student scenario~\cite{Gardner_1989,Gyordyi1990}, except for two important differences. Firstly, $M_a$, i.e. the overlap between the teacher and the student with replica index $a$ has a different definition (see eq.~\eqref{eq::order_params_capital_a}) since the two architectures live in spaces with different dimensions. Secondly, also the definition of the overlap matrix $Q_{ab}$ changes (see eq.~\eqref{eq::order_params_capital_b}). In particular notice that an additional matrix of overlaps $p_{ab}$ appears; this represents the overlap between the projection (in the teacher space) of the weights of two students with replica indexes $a$ and $b$. 
	
	Notice that equation~\eqref{eq::part_func_after_average1} can be obtained starting from~\eqref{eq::replicated_part_func}, also by using the following mapping (Gaussian covariate model)
	\begin{equation}
		\tilde \xi_i^\mu = \sigma\left( \frac{1}{\sqrt{D}} \sum_{k=1}^{D} F_{ki} \xi_k^\mu \right) = \mu_0 + \frac{\mu_1}{\sqrt{D}} \sum_{k=1}^{D} F_{ki} \xi_k^\mu + \mu_\star \eta_i^\mu
	\end{equation}
	where $\eta_i \sim \mathcal{N}(0, 1)$ are i.i.d. standard Gaussian random variables. 
	This means that in the thermodynamic limit~\eqref{eq::thermodynamic_limit}, the statistical properties of the random feature model are equivalent to a Gaussian covariate model, in which each projected pattern $\tilde{\boldsymbol{\xi}}^\mu$ is a \emph{linear} combination of the patterns components $\xi_k^\mu$ plus noise. The strength of the noise depends on the degree of non-linearity of the activation function $\sigma$. This was already noticed in~\cite{Mei2019}.
	
	In the following we will limit ourselves to the case $\sigma(x) = \text{sign}(x)$, but our analytical results are valid to the class of functions $\sigma(\cdot)$ for which $\mu_0 = 0$; this will also impose $\rho_a = 0$, reducing the number of terms in the calculations. 
	
	\subsection{Average over features and introduction of the order parameters}~\label{sec::average}
	Inserting the definition of the projected weights~\eqref{eq::projected_weights}  using delta functions it becomes easy to perform the average over random Gaussian features. We get a terms of the following form
	\begin{equation}
		\int \prod_{k a} \frac{d s_k^a d \hat s_k^a}{2\pi} \, e^{i \sum_{ka} s_k^a \hat s_k^a} \prod_{ki} \mathbb{E}_{F_{ki}} \left[ e^{-i \frac{F_{ki}}{\sqrt{N}} \sum_a \hat s_k^a w_i^a} \right] = \int \prod_{k a} \frac{d s_k^a d \hat s_k^a}{2\pi} \, e^{i \sum_{ka} s_k^a \hat s_k^a - \frac{1}{2} \sum_{ab, k} \hat s_k^a \hat s_k^b \left( \frac{1}{N} \sum_i w_i^a w_i^b \right)} \,.
	\end{equation}
	Next we can safely impose the definitions of the order parameters
	\begin{equation}
		q_{ab} \equiv \frac{1}{N} \sum_i w_i^a w_i^b\,, \qquad p_{ab} \equiv \frac{1}{D} \sum_k s_k^a s_k^b\,, \qquad r_a \equiv \frac{1}{D} \sum_k s_k^a w_k^T\,,
	\end{equation}
	Notice that $q_{aa} = 1$ since we have binary weights. Denoting by $\overline{\cdots}$ the average over both patterns and random features, the final result reads
	\begin{equation}
		\overline{Z^n} = \int \prod_{a<b} \frac{d q_{ab} d \hat q_{ab}}{2 \pi} \prod_{a \le b} \frac{d p_{ab} d \hat p_{ab}}{2 \pi} \prod_a \frac{d r_{a} d \hat r_{a}}{2 \pi} \, e^{N \phi}
	\end{equation}
	where 
	\begin{subequations}
		\label{eq::termsSP}
		\begin{align}
			\phi &= -\sum_{a<b} q_{ab} \hat q_{ab} - \frac{\alpha_D}{2} \sum_{ab} p_{ab} \hat p_{ab} - \alpha_D \sum_{a} r_a \hat r_a + G_{SS} + \alpha_D G_{SE} + \alpha G_{E} \\
			G_{SS} &= \ln \int \prod_a dw_a \, P_{w}(w_a) \, e^{ \, \frac{1}{2}\sum_{a	\ne b} \hat q_{ab} w_a w_b}\\
			\label{eq::termsSP_GSE}
			G_{SE} &= \ln \int \prod_a \frac{d s_a d \hat s_a}{2\pi} \, e^{i \sum_a s_a \hat s_a + \sum_a \hat r_a s_a + \frac{1}{2}\sum_{ab} \hat p_{ab} s_a s_b - \frac{1}{2} \sum_{ab} q_{ab} \hat s_a \hat s_b} \\
			G_{E} &= \ln \int \prod_a \frac{d \lambda_a d \hat \lambda_a}{2\pi} \frac{d u d \hat u}{2\pi} \, e^{i u \hat u + i \sum_a \lambda_a \hat \lambda_a -\beta \sum_a \Theta\left(- \text{sign}(u) \lambda_a + \kappa \right) - \frac{\hat u^2}{2} - \frac{1}{2} \sum_{ab} Q_{ab} \hat \lambda_a \hat \lambda_b - \hat u \sum_a M_a \hat \lambda_a }
		\end{align}
	\end{subequations}
	and $M_a$, $Q_{ab}$ are defined in terms of $q_{ab}$, $p_{ab}$, $r_a$ in~\eqref{eq::order_params_capital}, and as usual $\alpha_D \equiv D/N$. Notice that $G_{SS}$ is the usual ``entropic'' contribution in a perceptron storing random patterns, whereas $G_E$ is the usual ``energetic'' contribution in the teacher student setting. $G_{SE}$ is a new term that we call ``entropic-energetic'' since it depends on both overlaps $q_{ab}$ and conjugated ones $\hat p_{ab}$, $\hat r_a$. Notice that $G_{SE}$ can be computed analytically, since it contains only Gaussian integrals. It reads
	\begin{equation}
		G_{SE} = - \frac{1}{2} \ln \det \left( \mathbb{I} - q \hat p \right) + \frac{1}{2} \sum_{ab} \hat r_a \left[ \left(\mathbb{I} - q \hat p \right)^{-1} q  \right]_{ab} \hat r_b \,.
	\end{equation}

	\subsection{Replica-Symmetric ansatz}~\label{sec::Typical_RS}
	We impose a Replica-Symmetric (RS) ansatz for the order parameters: $q_{ab} = \delta_{ab} + q \left(1 - \delta_{ab} \right)$, $\hat q_{ab} = \hat q \left(1 - \delta_{ab} \right)$; $p_{ab} = p_d \delta_{ab} + p \left(1 - \delta_{ab} \right)$, $\hat p_{ab} = - \hat p_d \delta_{ab} + \hat p \left(1 - \delta_{ab} \right)$ and $r_a = r$, $\hat r_a =  \hat r$. 
	
	We obtain
	\begin{subequations}
		\begin{align}
			\mathcal{G}_{SS} &\equiv \frac{\hat q}{2} + \lim\limits_{n \to 0} \frac{G_{SS}}{n} = \int Dx \, \ln 2\cosh\left(\sqrt{\hat q} x \right) \\
			\mathcal{G}_{SE} &\equiv \lim\limits_{n \to 0} \frac{G_{SE}}{n} = - \frac{1}{2} \frac{q}{1-q} - \frac{1}{2} \ln \left[ 1 + (\hat p + \hat p_d) (1-q) \right] + \frac{1}{2} \frac{(\hat p + \hat r^2)(1-q) + \frac{q}{1-q}}{1+ (\hat p + \hat p_d)(1-q)} \\
			\mathcal{G}_{E} &\equiv \lim\limits_{n \to 0} \frac{G_{E}}{n} = 2 \int Dx \, H \left( - \frac{M x}{\sqrt{Q - M^2}} \right) \ln H_\beta \left( \frac{\kappa - \sqrt{Q} x}{\sqrt{Q_d - Q}} \right)
		\end{align}
	\end{subequations}
	where $M \equiv \mu_1 r$, $Q \equiv \mu_\star^2 q + \mu_1^2 p$, $Q_d \equiv \mu_\star^2 + \mu_1^2 p_d$. We have also defined 
	\begin{subequations}
		\begin{align}
			H\left( x \right) &\equiv \frac{1}{2} \text{Erfc} \left( \frac{x}{\sqrt{2}} \right) \,, \\
			H_\beta\left( x \right) &\equiv e^{-\beta} + \left(1-e^{-\beta}\right) H\left( x \right) \,.
		\end{align}
	\end{subequations}
	The free entropy of the system is
	\begin{equation}
		\phi = -\frac{\hat q}{2} (1-q) + \frac{\alpha_D}{2} \left( p_d \hat p_d + p \hat p \right) - \alpha_D r \hat r + \mathcal{G}_{SS} + \alpha_D \mathcal{G}_{SE} + \alpha \mathcal{G}_{E} \,.
	\end{equation}
	The order parameters $q$, $p$, $p_d$, $r$, $\hat q$, $\hat p$, $\hat p_d$, $\hat r$ are found by saddle point equations
	\begin{equation}
		\label{eq::RS_SP_equations}
		\begin{aligned}
			q &= 1 - 2\frac{\partial \mathcal{G}_{SS}}{\partial \hat q}\,, &  p &= - 2 \frac{\partial \mathcal{G}_{SE}}{\partial \hat p}\,, & p_d &= - 2 \frac{\partial \mathcal{G}_{SE}}{\partial \hat p_d}\,, & r &= \frac{\partial \mathcal{G}_{SE}}{\partial \hat r}\,, \\
			\hat q &= - 2\alpha_D \frac{\partial \mathcal{G}_{SE}}{\partial q} - 2\alpha\frac{\partial \mathcal{G}_{E}}{\partial q}\,, & \hat p &= - 2 \alpha_T \frac{\partial \mathcal{G}_{E}}{\partial p}\,, & \hat p_d &= - 2 \alpha_T \frac{\partial \mathcal{G}_{E}}{\partial p_d}\,, & \hat r &= \alpha_T \frac{\partial \mathcal{G}_{E}}{\partial r}\,,
		\end{aligned}
	\end{equation}
	
\begin{figure}
		\begin{centering}
			\includegraphics[width=0.49\linewidth]{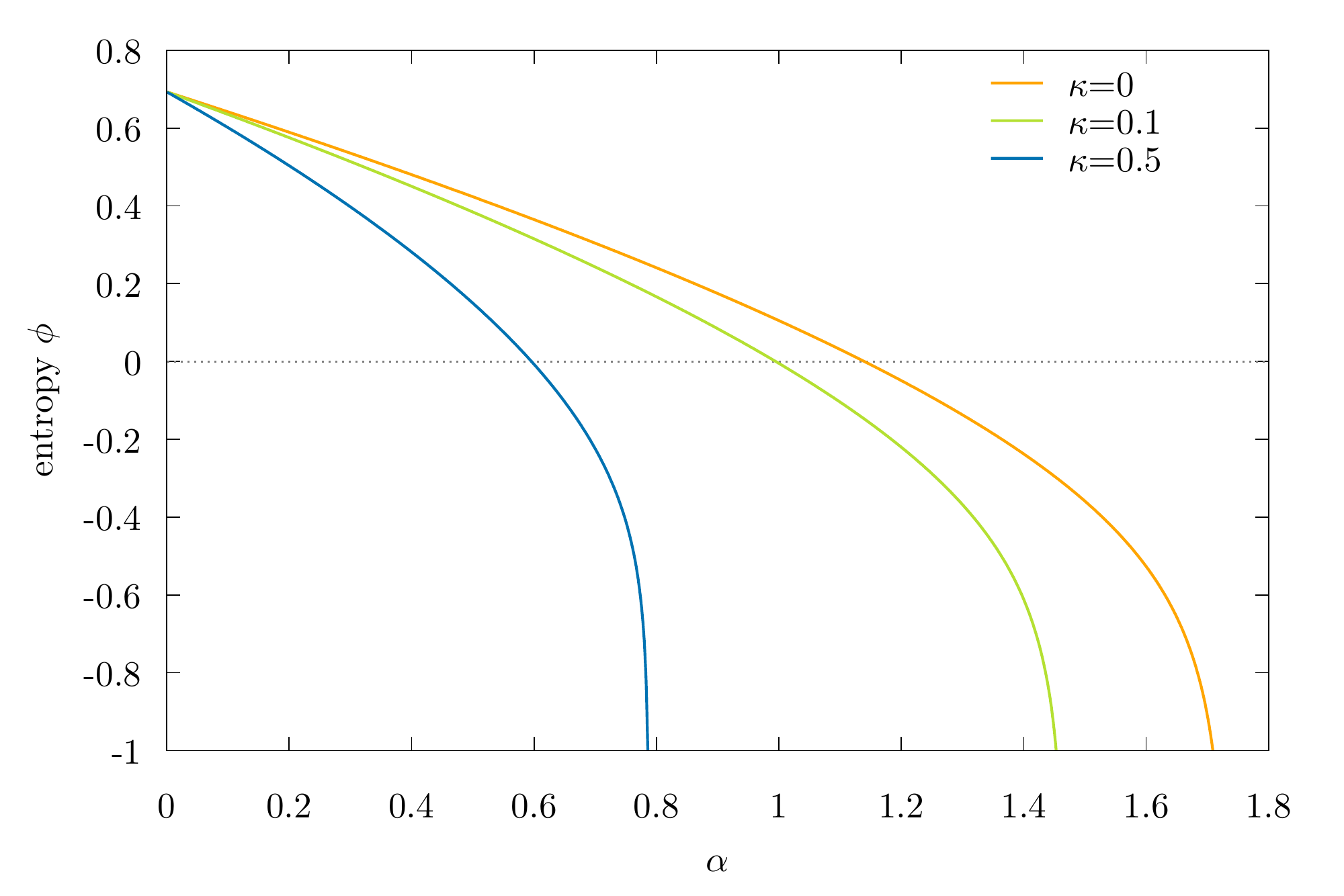}
			\includegraphics[width=0.49\linewidth]{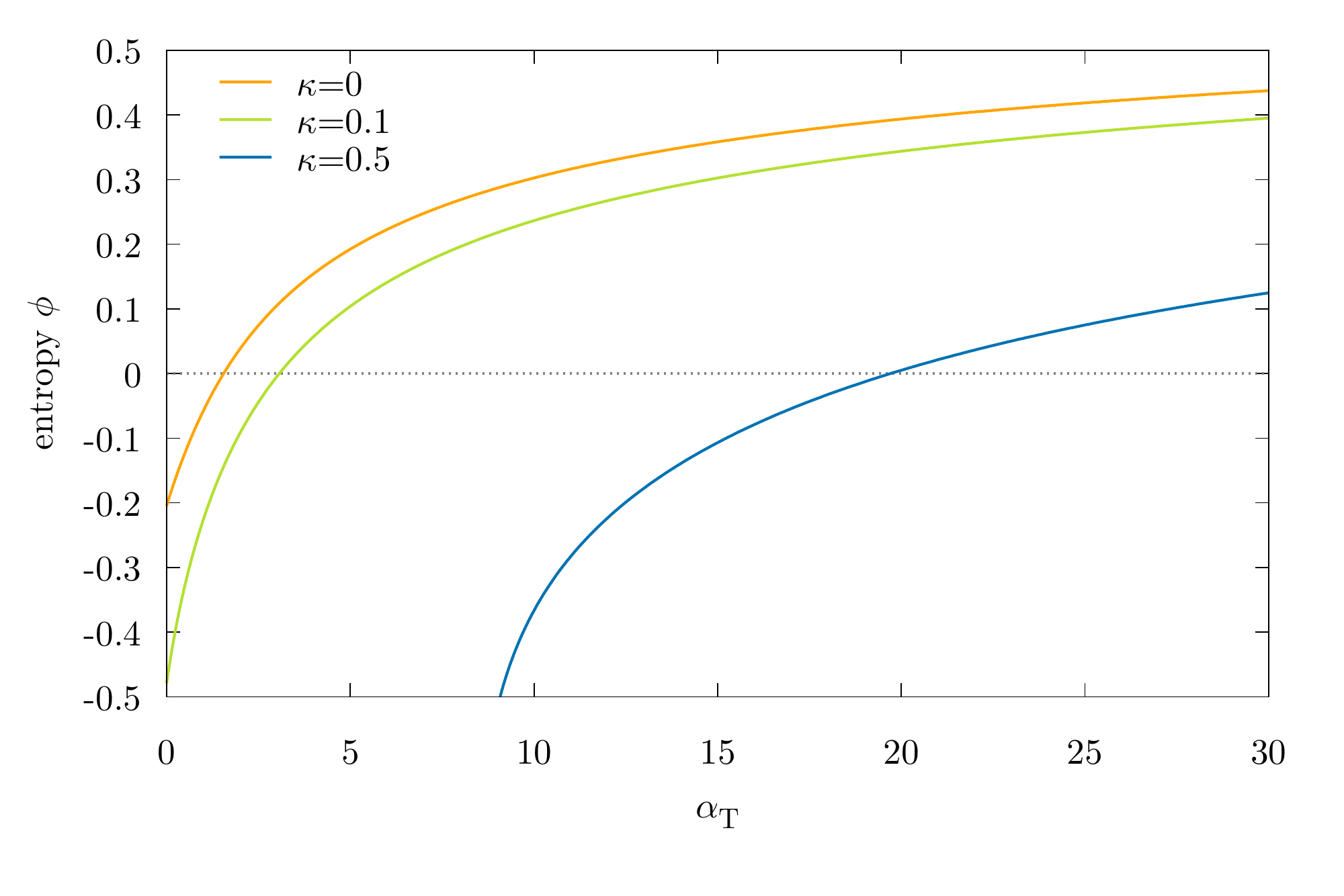}
		\end{centering}
		\caption{Plot of the entropy as a function of $\alpha$ for fixed $\alpha_T=3$ (left panel) and as a function of $\alpha_T$ for fixed $\alpha = 1$ (right panel) for different values of the margin $\kappa$. }
		\label{Fig::entropy} 
	\end{figure}

	As in the simple binary perceptron~\cite{krauth1989storage}, the ``interpolation threshold'' or critical capacity is found by looking to the value of $\alpha$ for which the RS free entropy vanishes. We show in~\ref{Fig::entropy} the behaviour of the entropy (i.e. the free entropy in the $\beta \to \infty$ limit) as a function of $\alpha$ (for a fixed value of $\alpha_T$) and $\alpha_T$ (for a fixed value of $\alpha$) for different margins.

	\subsubsection{Generalization error}
	To compute the generalization error, we extract a new pattern $\boldsymbol \xi^\star$ and label $y^\star$ and we compute the average number of errors. Denoting by $\left< \cdot \right>$ the ensemble average, we have
	\begin{equation}
		\begin{split}
			\epsilon_g &\equiv \langle \mathbb{E}_{\boldsymbol{\xi}^\star} \Theta\left( - y^\star \hat y^\star \right) \rangle \\
			&= \int d u d \lambda\,  \Theta\left( - u \lambda \right)
			\mathbb{E}_{\boldsymbol{\xi}^\star}  \left\langle \delta\left( u - \frac{1}{\sqrt{D}} \sum_{k=1}^{D} w_k^T \xi_k^\star \right) \delta \left( \lambda - \frac{1}{\sqrt{N}}  \sum_{i=1}^{N} w_i \, \sigma\left( \frac{1}{\sqrt{D}} \sum_{k=1}^{D} F_{ki} \xi_k^\star \right)  \right) \right\rangle \\
			&= \int \frac{d u d \hat u}{2\pi} \frac{d \lambda d \hat \lambda}{2\pi}\, \Theta\left( - u \lambda \right) e^{i u \hat u + 
				i \lambda \hat \lambda	- \frac{\hat u^2}{2} -  \frac{1}{2} Q_{d} \hat \lambda^2 -  M \hat u \hat \lambda} 
			= 2 \int_0^{\infty} Du \, H\left( - \frac{M u}{\sqrt{Q_d-M^2}} \right) \,.
		\end{split}
	\end{equation}
	Performing the last integral we finally obtain
	\begin{equation}
		\epsilon_g = \frac{1}{\pi} \arccos\left( \frac{M}{\sqrt{Q_d}} \right)\,,
	\end{equation}
	which is nothing but the standard formula of the generalization error for the classical teacher-student problem, but written in terms of the ``projected'' overlap with the teacher $M$ and the ``projected'' norm of the weights $Q_d$.

	\subsubsection{Storage problem}
	When the dimension of the teacher is much larger than the number of patterns in the training set $\alpha_T = \frac{P}{D} \ll 0$, the problem is as if the student sees random patterns. 
	Indeed the saddle point equations~\eqref{eq::RS_SP_equations} 
	reduce in the limit $\alpha_T \to 0$ to
	\begin{equation}
		\label{eq::RS_SP_equations_storage}
		\begin{aligned}
			q &= 1 - 2\frac{\partial \mathcal{G}_{SS}}{\partial \hat q}\,, &  p &= q\,, & p_d &= 1 \,, & r &= 0 \,, \\
			\hat q &= - 2\alpha\frac{\partial \mathcal{G}_{E}}{\partial q}\,, & \hat p &= 0\,, & \hat p_d &= 0 \,, & \hat r &= 0 \,.
		\end{aligned}
	\end{equation}
	Therefore $Q = \left( \mu_\star^2 + \mu_1^2 \right)q$, $Q_d = \mu_\star^2 + \mu_1^2$ and the free entropy reduces to
	\begin{equation}
		\phi = -\frac{\hat q}{2} (1-q) + \mathcal{G}_{SS}  + \alpha \mathcal{G}_{E} \,.
	\end{equation}
	with
	\begin{subequations}
		\begin{align}
			\mathcal{G}_{SS} &= \int Dx \, \ln \cosh\left(\sqrt{\hat q} x \right) \\
			\mathcal{G}_{E} &= \int Dx \, \ln H_\beta \left( - \sqrt{\frac{q}{1 - q}} x \right) \,.
		\end{align}
	\end{subequations}
	This is exactly the free entropy of the storage problem as derived by Gardner~\cite{gardner1988The,gardner1988optimal}. Notice that this limit is achieved independently of the non-linearity $\sigma(\cdot)$ used.

	\subsubsection{Overparameterization limit}
	\begin{figure}
		\begin{centering}
			\includegraphics[width=0.49\columnwidth]{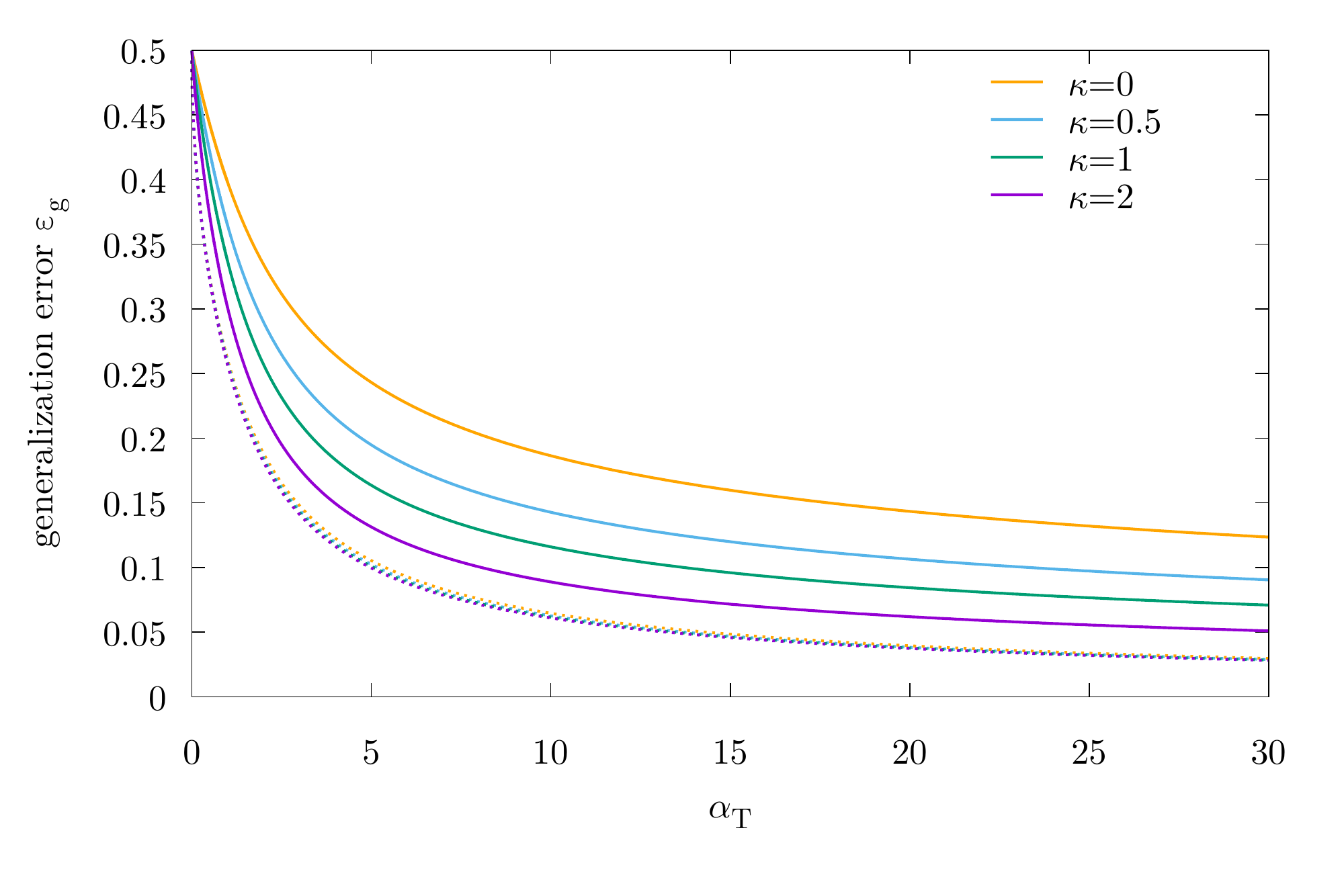}
		\end{centering}
		\caption{
			Generalization error in the large overparameterization limit ($\alpha \to 0$) versus $\alpha_T = P/D$ for different values of the margin $\kappa$. The dotted lines represent the corresponding generalization error of the barycenter of typical solutions (see section \ref{sec::Barycenter}) }
		\label{Fig::gen_err_plateau} 
	\end{figure}
	Here we want to address analytically the infinite overparameterization limit, i.e. $\alpha \to 0$ for a fixed value of $\alpha_T$. In this limit also $\alpha_D = \frac{\alpha}{\alpha_T}$ is vanishing, therefore from saddle point equations~\eqref{eq::RS_SP_equations} we see that $\hat q \to 0$ and consequently $q \to 0$, meaning that typical solutions are uncorrelated in the space of the students. However there is still information about the teacher, so the corresponding overlap in the space of the teacher is not zero. Furthermore we can eliminate all other conjugated parameters $\hat p$, $\hat p_d$ and $\hat r$ by expressing them in terms of the other order parameters. The entropy can be written as
	\begin{equation}
		\phi \simeq \ln 2 + \frac{\alpha}{\alpha_T} \delta \phi
	\end{equation}
	where
	\begin{subequations}
		\begin{align}
			\delta \phi &= \frac{1}{2} \left(1 - p_d - \frac{r^2}{p_d-p} \right) + \mathcal{G}_{SE} + \alpha_T \mathcal{G}_E \\
			\mathcal{G}_{SE} &= \frac{1}{2} \left( \frac{p}{p_d-p} + \ln(p_d - p) \right) \\
			\mathcal{G}_{E} &= 2 \int Dx \, H \left( - \frac{r x}{\sqrt{p-r^2}} \right) \ln H_\beta \left( \frac{\kappa - \sqrt{p} x }{\sqrt{\frac{\mu_\star^2}{\mu_1^2} + p_d - p}} \right) 
		\end{align}
	\end{subequations}
	Notice that $\mathcal{G}_{E}$ apart for the dependence on $\sigma(\cdot)$ is identical to the energetic term of the classical teacher-student problem. Instead $\mathcal{G}_{SE}$ is identical to the entropic term of a spherical perceptron storing random patterns. 
	
	By solving the corresponding saddle point equations, we are able to numerically compute the plateau of the generalization error; this is plotted as a function of $\alpha_T$ in Fig.~\ref{Fig::gen_err_plateau} for different values of the margin. 
	
	\subsubsection{Stability distribution}

	The stability of the weights $\boldsymbol{w}$ given a pattern $\xi^\mu$ and its corresponding label $y^\mu$ is defined as
	\begin{equation}
		\Delta^\mu \equiv y^\mu \lambda^\mu = \frac{y^\mu}{\sqrt{N}} \sum_{i=1}^{N} w_i \sigma\left( \frac{1}{\sqrt{D}} \sum_{k=1}^{D} F_{ki} \xi_k^\mu \right) \,.
	\end{equation}
	We restrict for simplicity to the case of zero margin $\kappa$. Once the saddle point equations~\eqref{eq::RS_SP_equations} are solved, we can compute the stability distribution
	\begin{equation}
		P(\Delta) \equiv \langle \delta\left( \Delta - \Delta^\mu \right) \rangle = \overline{\frac{1}{Z}\int \prod_i dw_i \, P_{w}(\boldsymbol w) \, e^{-\beta \sum_{\mu=1}^{P} \Theta\left( - y^\mu \hat y^\mu \right)} \delta\left( \Delta - \Delta^\mu \right)}
	\end{equation} 
	using the replica method. We obtain
	\begin{equation}
		P(\Delta) = \lim\limits_{n \to 0} \int \frac{d u d \hat u}{2\pi} \prod_a \frac{d \lambda_a d \hat \lambda_a}{2\pi} \, e^{i u \hat u + i \sum_a \lambda_a \hat \lambda_a - \frac{\hat u^2}{2} - \beta \sum_a \Theta\left(-u \lambda_a\right) - \frac{1}{2}\sum_{ab} Q_{ab} \hat \lambda_a \hat \lambda_b - \hat u\sum_a M_a \hat \lambda_a} \delta\left( \Delta - \text{sign}(u) \lambda_1 \right)\,,
	\end{equation}
	that in the RS ansatz reduces to
	\begin{equation}
		\label{eq::P(delta)}
		P(\Delta) = \frac{2 e^{-\beta \Theta(-\Delta)}}{\sqrt{Q_d - Q}} \int Dx \, G\left( \frac{\Delta - \sqrt{Q} x}{\sqrt{Q_d - Q}} \right) \frac{H\left(- \frac{M x}{\sqrt{Q-M^2}}\right)}{H_\beta \left( - \sqrt{\frac{Q}{Q_d-Q}} x \right)} \,,
	\end{equation}
	where $G(x) \equiv \frac{e^{-x^2/2}}{\sqrt{2\pi}}$. 
	
	In Fig.~\ref{Fig::stabilities} we show the distribution of stabilities of typical solutions for different values of $\alpha_T$. The maximum of the distribution appears to be near the origin, especially for low values of $\alpha_T$; this has been already noted to be a characteristic of ``sharp'' solutions in one-layer~\cite{engel-vandenbroek} and two-layer neural networks~\cite{relu_locent} in contrast to ``flat'' or high local entropy ones, for which it is usually noted that a low probability of having a small stability (i.e. targeting high local entropy regions induces a soft margin).

	\section{Agreement with numerical simulations}
	
	\begin{figure}
		\begin{centering}
			\includegraphics[width=0.49\columnwidth]{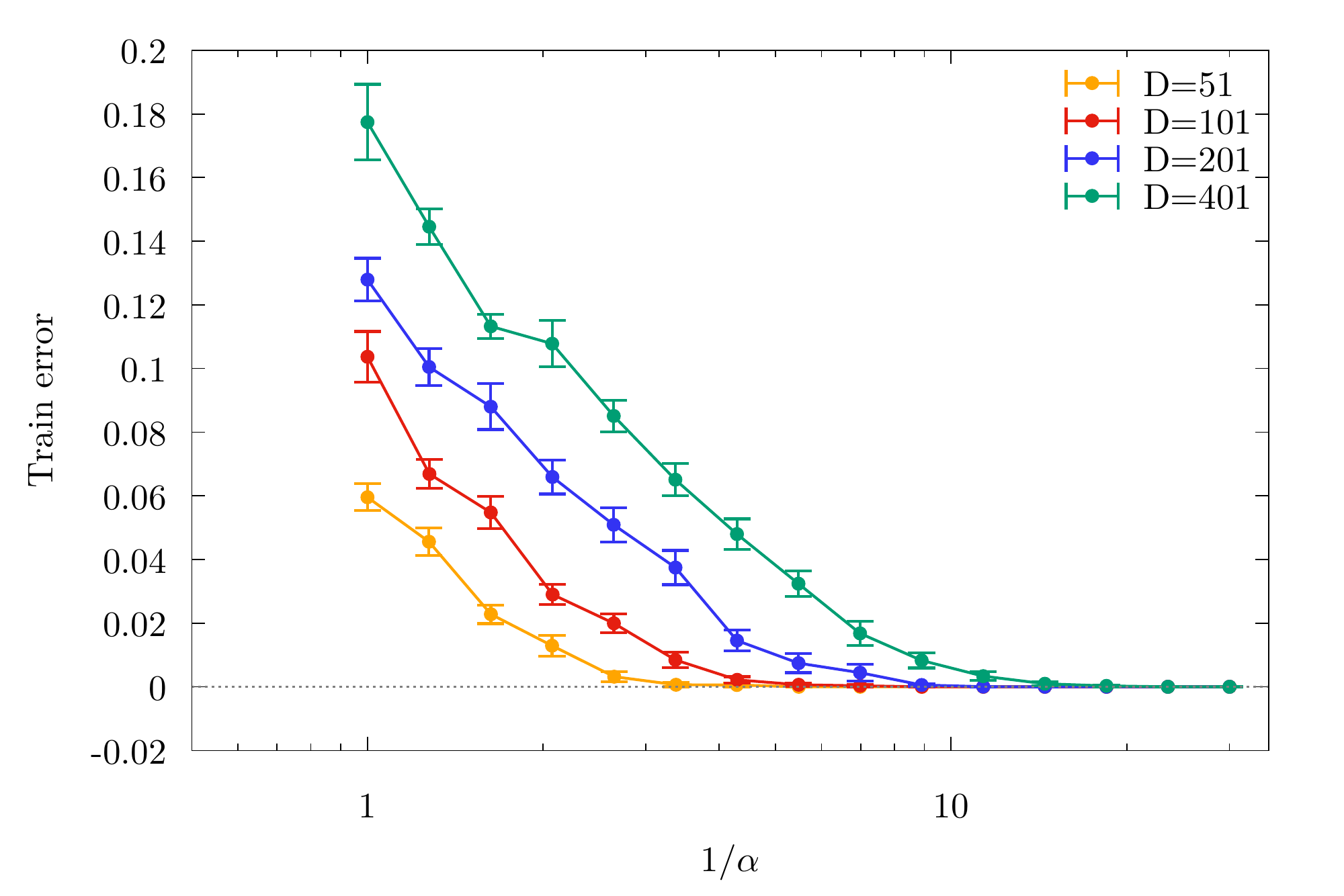}
			\includegraphics[width=0.49\columnwidth]{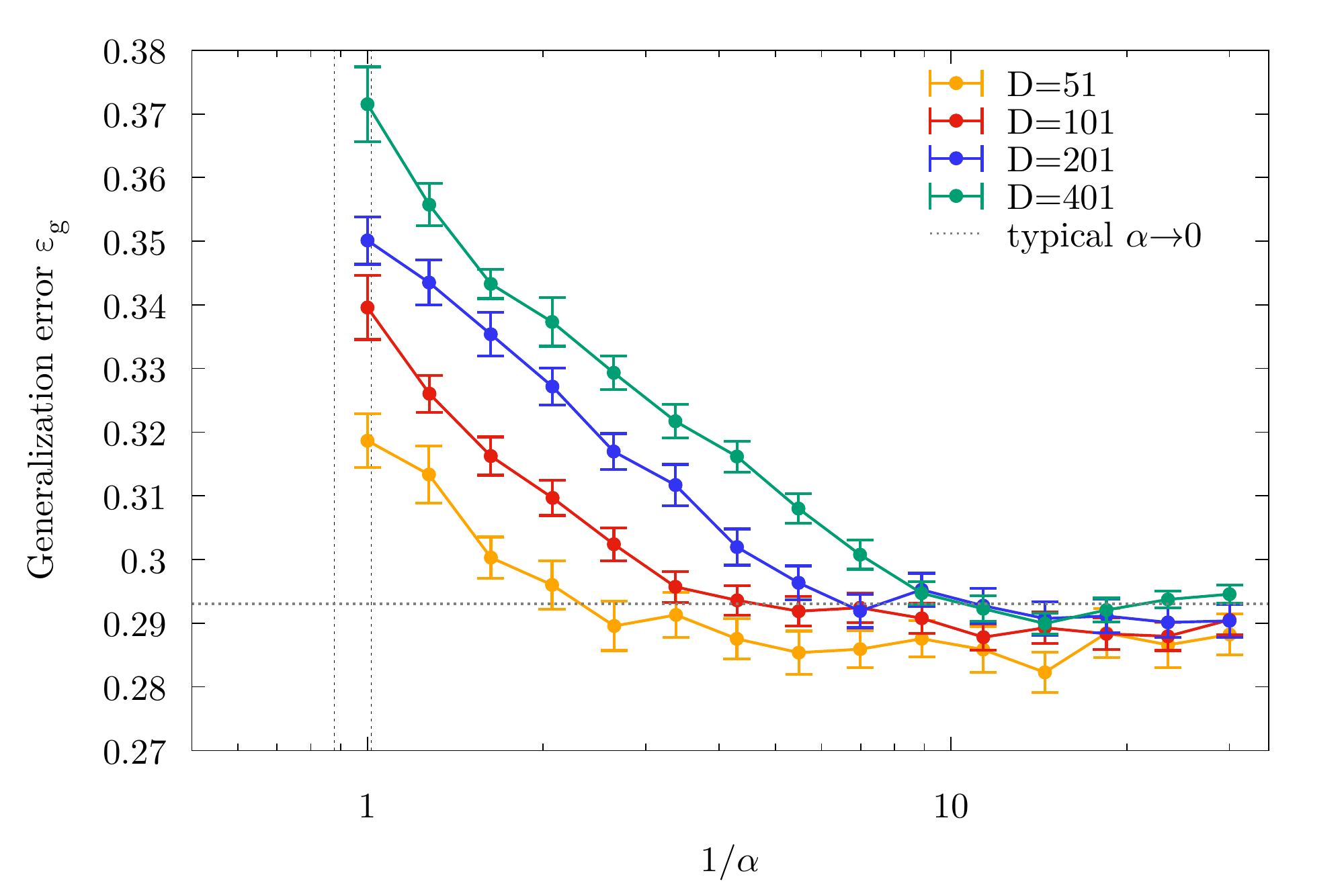}
		\end{centering}
		\caption{Train (left panel) and test error (right panel) of the SA algorithm as a function of $1/\alpha$. Different colors represent the result of the simulations with different values of $D$, while maintaining fixed $\alpha_T = 3$. The points are averages over $40$ samples for $D=51,101$, $20$ samples for $D=201, 401$, and $2$ independent runs per sample. Approaching the thermodynamic limit, it is harder to find a solution. Nonetheless for a fixed system size we can reach zero or sufficiently small the training errors in the overparameterized regime. The corresponding generalization error matches the replica theory result (dotted horizontal line). }
		\label{Fig::SA} 
	\end{figure}
	We have performed some numerical simulations in order to corroborate analytical results of typical solutions. We have used very simple algorithms that have the Gibbs distribution as stationary probability measure such as the zero-temperature Monte Carlo (MCT0) and the Simulated Annealing algorithm (SA)~\cite{KirkpatrickSA}. 
	
\begin{figure}
		\begin{centering}
			\includegraphics[width=0.49\columnwidth]{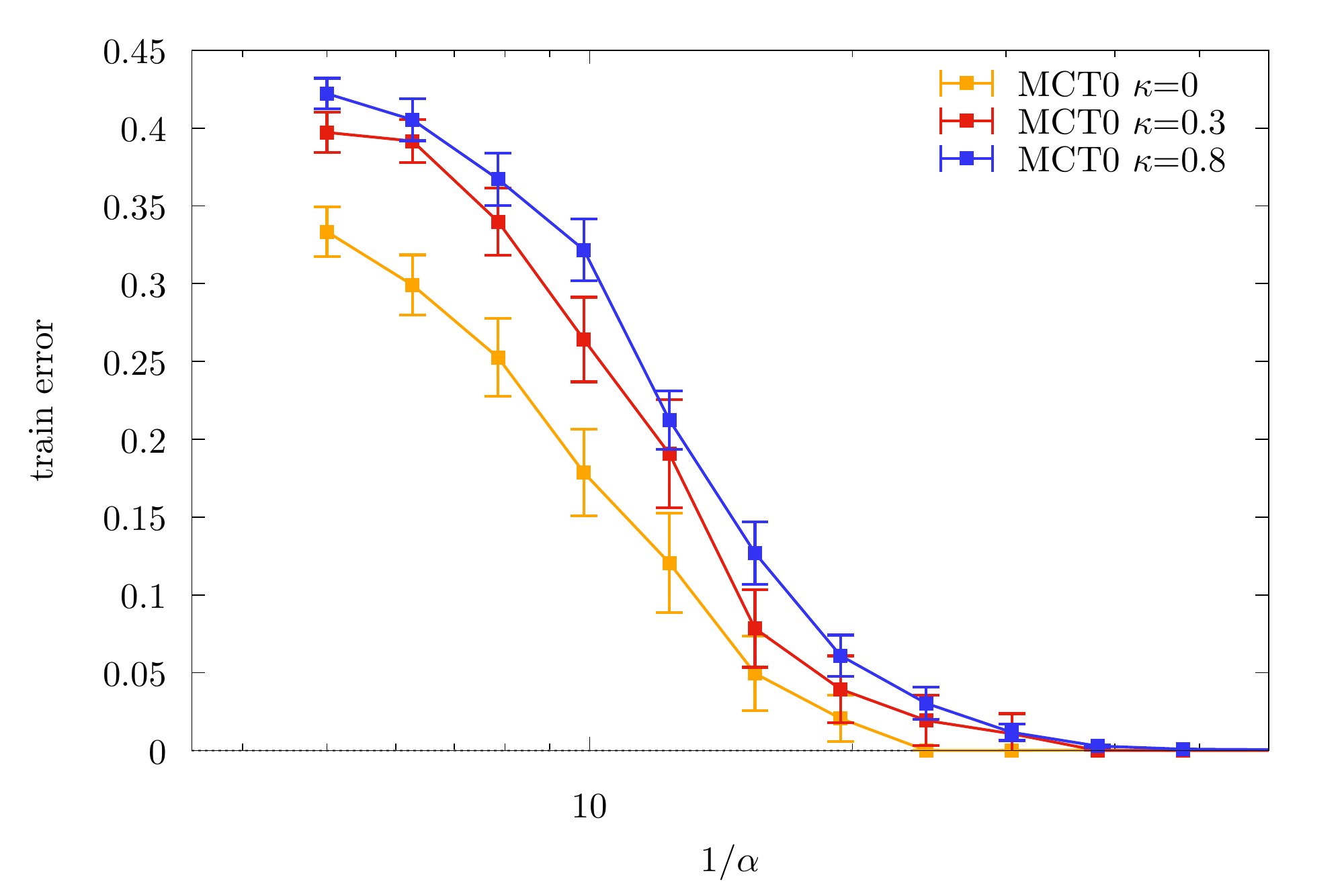}
			\includegraphics[width=0.49\columnwidth]{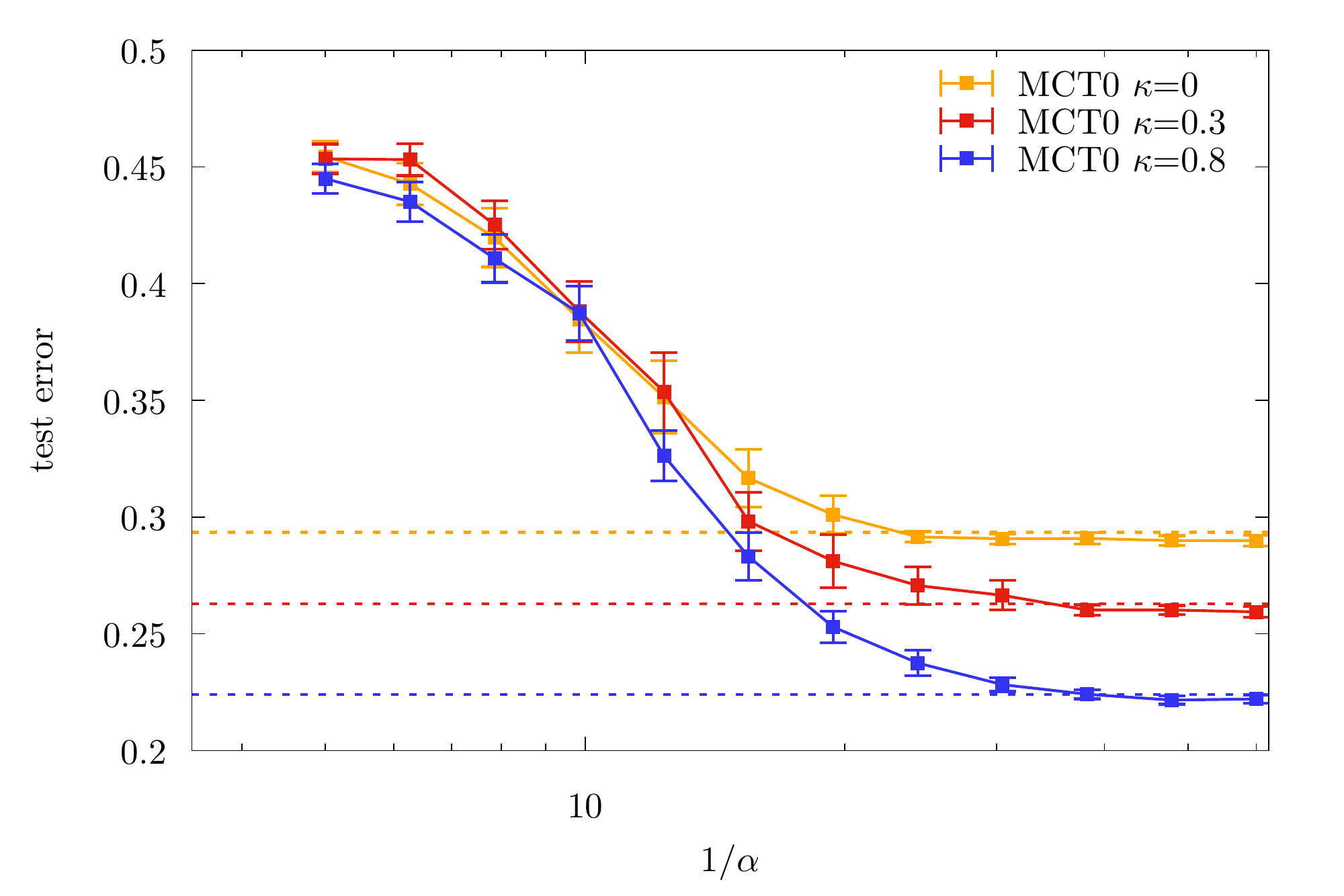}
		\end{centering}
		\caption{Train (left panel) and test error (right panel) of zero-temperature Monte Carlo algorithm as a function of $1/\alpha$ for different values of margins. In the simulations we fixed $D=201$ and $P=603$ i.e. $\alpha_T = 3$, and we ran the algorithm for a fixed number of sweeps ($200$). Points are averages over $10$ samples and $2$ random restarts for each sample. We show also in dashed the analytical predictions coming from replica theory.}
		\label{Fig::MCT0_margin} 
	\end{figure}

\begin{figure}
		\begin{centering}
			\includegraphics[width=0.49\columnwidth]{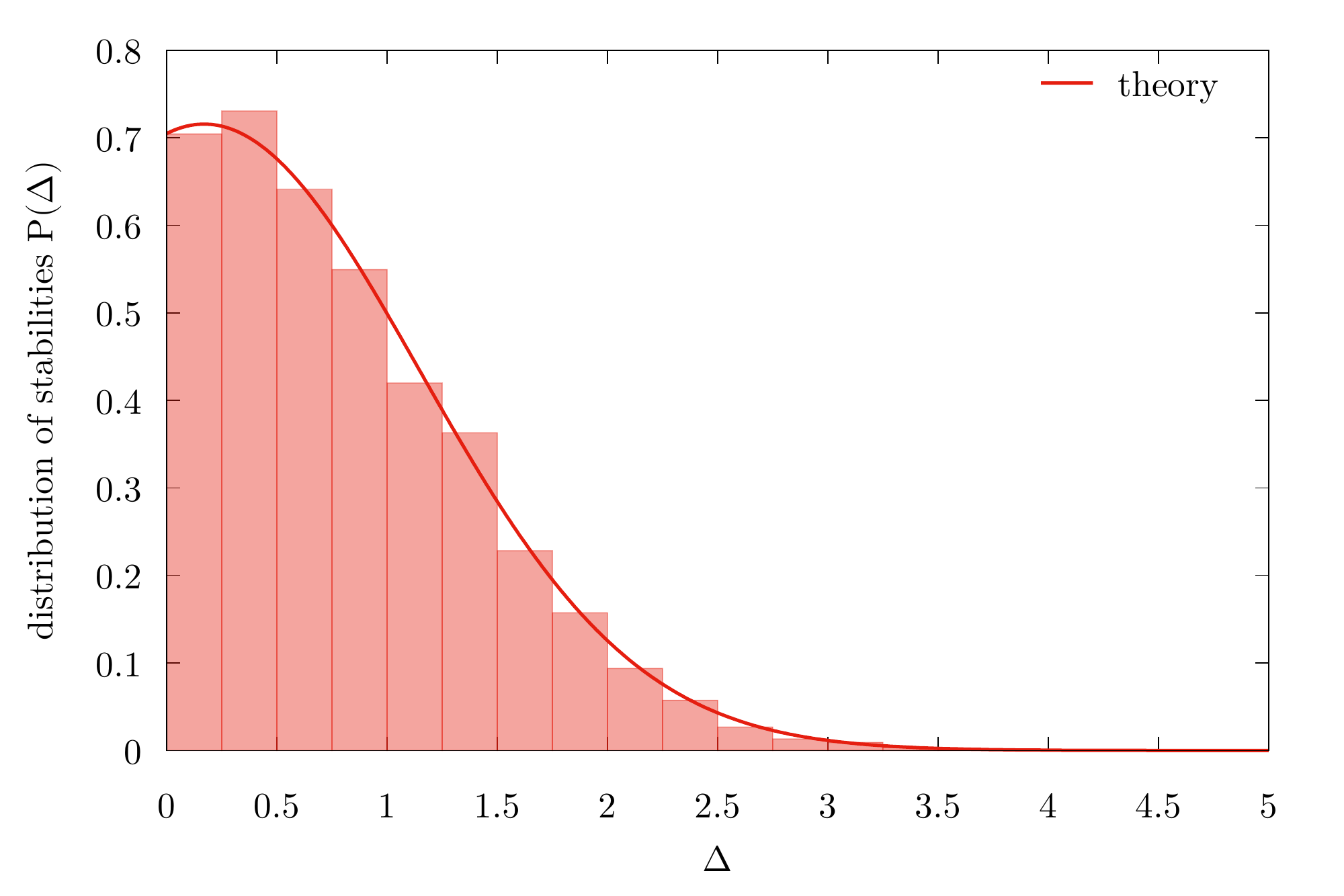}
			\includegraphics[width=0.49\columnwidth]{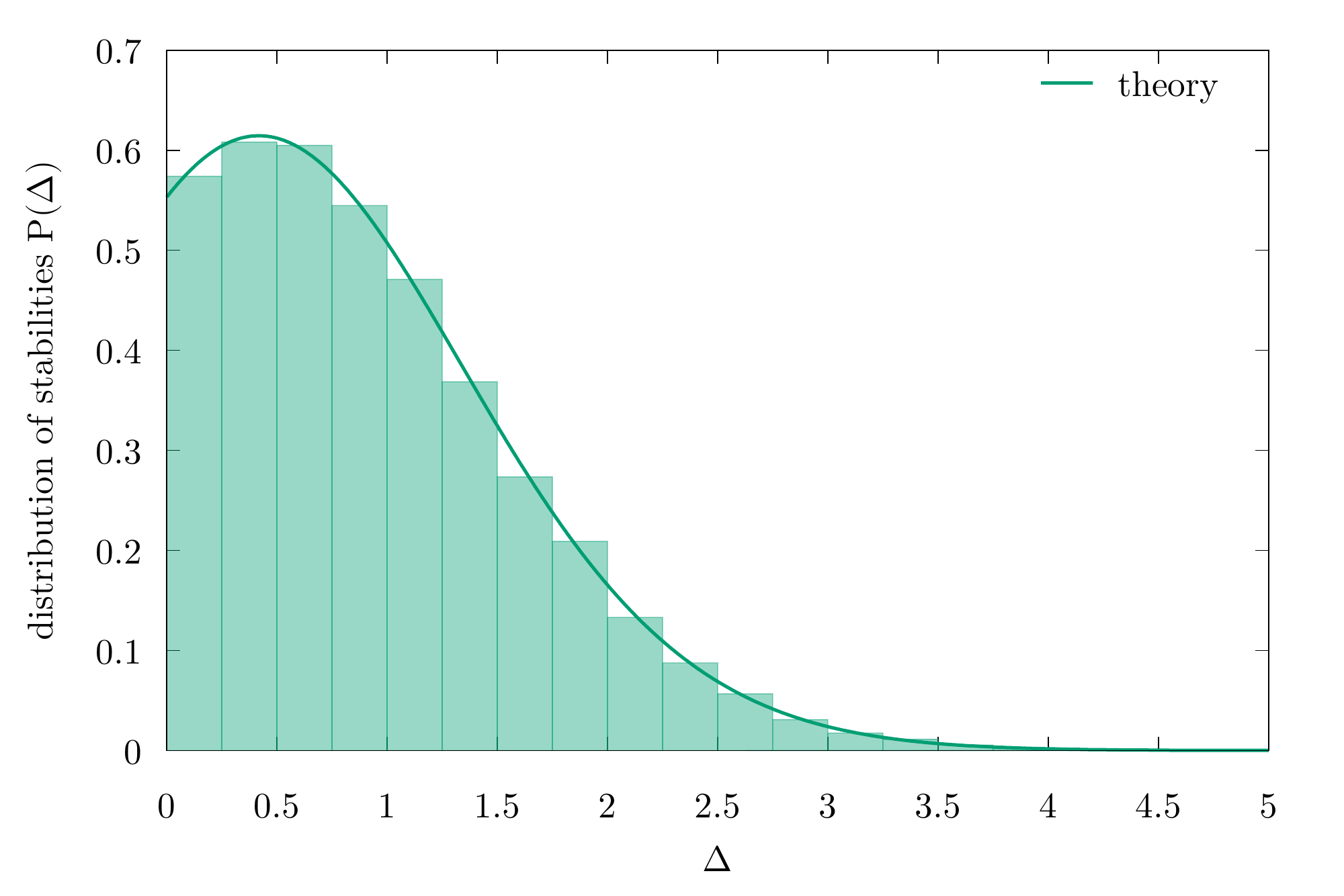}
			\includegraphics[width=0.49\columnwidth]{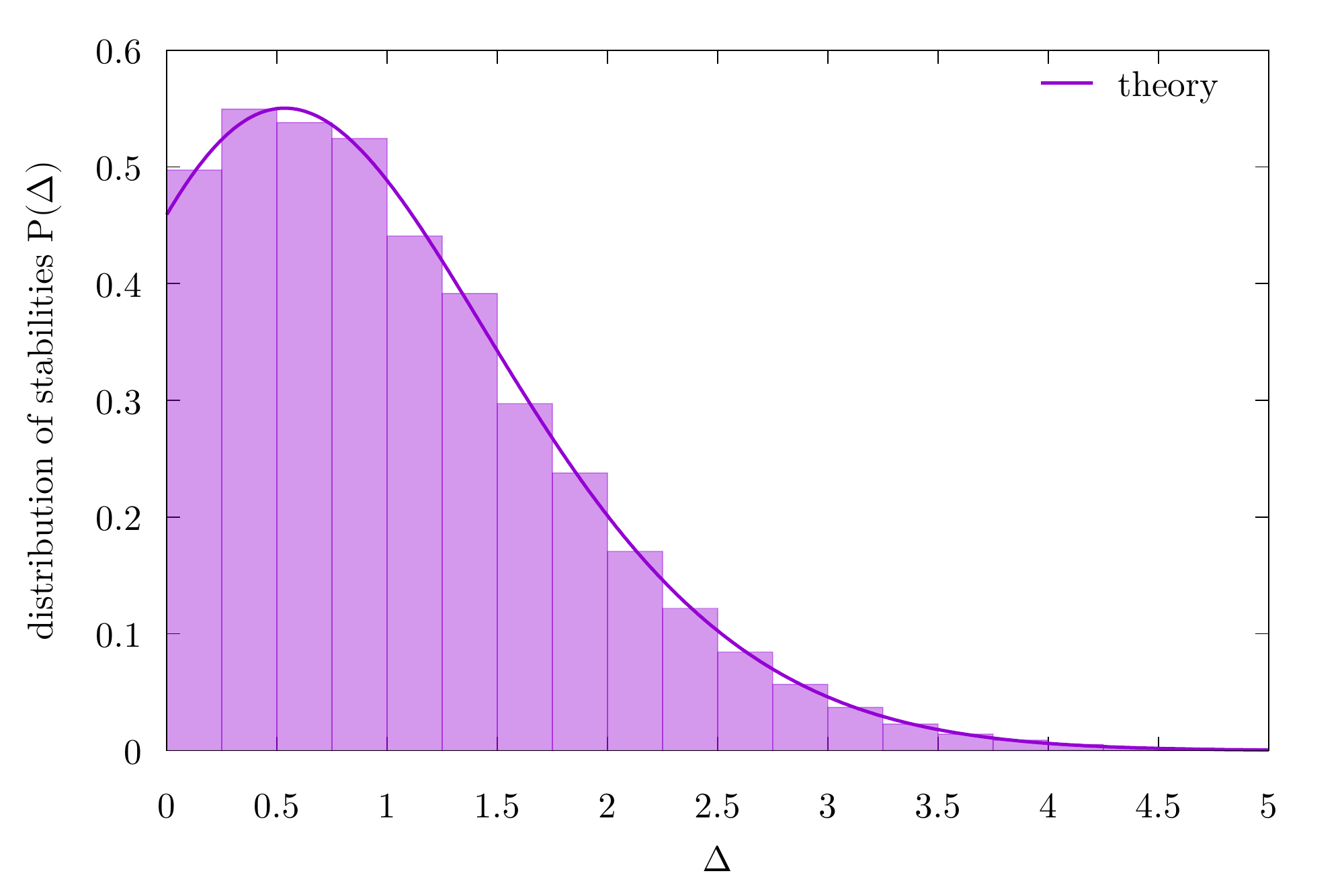}
		\end{centering}
		\caption{Histograms of the distribution of stabilities for zero temperature Monte Carlo solutions for $\alpha_T = 1.4$ (top left), $\alpha_T = 5$ (top right) and $\alpha_T = 10$ (bottom). The algorithm was ran until a solution is found or a maximum number of sweeps ($200$) is reached. The histograms are averaged over $40$ samples and $2$ random restarts for each sample. The full line is the replica prediction of the distribution for typical solutions in the thermodynamic limit given by equation~\eqref{eq::P(delta)}.}
		\label{Fig::stabilities} 
	\end{figure}

	Both algorithms have difficulties in finding solutions since the dominant set of minima consist of isolated point like clusters with vanishing internal entropy. Nevertheless, for finite size systems, in the highly overparameterized regime those algorithms are able to find solutions (see left panel of Fig.~\ref{Fig::SA} for the behaviour of the train error as a function of $1/\alpha$ obtained by using the SA algorithm for different system sizes). The statistical properties of those solutions are in agreement with the predictions of the replica theory (see right panel of Fig.~\ref{Fig::SA}). The same results hold for MCT0: in Fig.~\ref{Fig::MCT0_margin} we show that the generalization error obtained by replica theory for several values of the margin is in perfect agreement with that obtained by numerical simulations. Notice how increasing the margin makes finding the solution more difficult (since they are rarer); however when solutions start to be accessible, increasing the margin increases the accuracy on the test set. This is consistent with  the fact that even if high margin solutions lie in flat regions of the loss landscape (see main text) they are still isolated between each other. 
	Finally in Fig.~\ref{Fig::stabilities} we show the agreement between the analytical (see equation~\eqref{eq::P(delta)}) and numerical distribution of stabilities for different values of $\alpha_T$ obtained by MCT0. 
	
	\section{Bayesian generalization error} \label{sec::Barycenter}
	We want to compute the average probability that the ensemble of students generalizes correctly with respect to the teacher, i.e. the probability that the average of the outputs of the students on a random new pattern has different sign than that given by
	the teacher:
	\begin{equation}
		\epsilon_g^B = \mathbb{E}_{\boldsymbol{\xi}^\star} 
		\, \overline{ \Theta\left( - y^\star \langle \hat y^\star \rangle_{\boldsymbol{w} | \left\{ \boldsymbol{\xi}^\mu \right\}} \right) }
	\end{equation}
	In the previous equation $\boldsymbol{\xi}^\star$, $y^\star$ are respectively a test pattern and its corresponding label (computed using equation~\eqref{eq::label}); $\hat y^\star$ is the output of the student given input $\boldsymbol{\xi}^\star$ as in~\eqref{eq::prediction}; $\overline{\cdots}$ is the average over training patterns $\left\{ \boldsymbol{\xi}^\mu \right\}$ and random features $F$, and finally $\langle \cdot \rangle_{\boldsymbol{w} | \left\{ \boldsymbol{\xi}^\mu \right\}}$ is the average over the posterior distribution, namely the average over the probability distribution of student weights given training data
	\begin{multline}
		\label{eq::ensemble_average_students}
		\langle \hat y^\star \rangle_{\boldsymbol{w} | \left\{ \boldsymbol{\xi}^\mu \right\}} = \frac{1}{Z} \int \prod_i dw_i \, P_{w}(\boldsymbol{w}) \, \text{sign}\left( \frac{1}{\sqrt{N}} \sum_i w_i \sigma\left( \frac{1}{\sqrt{D}} \sum_k F_{ki} \xi_k^\star \right) \right) \\
		\times e^{-\beta \sum_{\mu} \Theta\left[ - \left( \frac{1}{\sqrt{D}} \sum_k \xi_k^\mu \right) \left( \frac{1}{\sqrt{N}} \sum_i w_i \sigma\left( \frac{1}{\sqrt{D}} \sum_k F_{ki} \xi_k^\mu \right) \right) \right] }
	\end{multline}
	We have used again $w_k^T=1$ without loss of generality. 
	We start the computation by extracting the definitions $\hat y^\star$ by using delta functions
	\begin{equation}
		\label{eq::Bayesian_Gen_error1}
		\epsilon_g^B = \int \frac{d x d \hat x}{2\pi} \, e^{i x \hat x} \, \mathbb{E}_{\boldsymbol{\xi}^\star} \Theta\left[ - \left( \frac{1}{\sqrt{D}} \sum_k \xi_k^\star \right)x \right] \overline{ e^{-i \hat x \left\langle y^\star \right\rangle_{\boldsymbol{w} | \left\{ \boldsymbol{\xi}^\mu \right\}}}} \,.
	\end{equation}
	Next, we use the following identity
	\begin{equation}
		\label{eq::trick}
		e^{-i \hat x \left\langle y^\star \right\rangle_{\boldsymbol{w} | \left\{ \boldsymbol{\xi}^\mu \right\}}} = \sum_{s=0}^{\infty} \frac{(-i \hat x)^s}{s!} \left\langle y^\star \right\rangle_{\boldsymbol{w} | \left\{ \boldsymbol{\xi}^\mu \right\}}^s \,,
	\end{equation}
	which enables us to perform the average over training patterns and random features. Two replica indexes are needed: the first one is due to the factor $Z^{-1}$ in equation~\eqref{eq::ensemble_average_students} which can be re-written as
	$\frac{1}{Z} = \lim\limits_{n \to 0} Z^{n-1}$; the second one is due to the power $s$ in~\eqref{eq::trick}. We use as before $a$, $b \in [n]$, whereas indexes $l$, $m \in [s]$ for the new replicas. We have
	\begin{multline}
		\left\langle y^\star \right\rangle_{\boldsymbol{w} | \left\{ \boldsymbol{\xi}^\mu \right\}}^s = \lim\limits_{n \to 0} \int \prod_l \frac{d h_l d \hat h_l}{2 \pi} e^{i \sum_l h_l \hat h_l} \prod_l \text{sign}(h_l) \int \prod_{ial} dw_i^{la} P_w(\boldsymbol{w}^{la}) \\
		\times e^{-\beta \sum_{\mu a} \Theta\left[ - \left(\frac{1}{D} \sum_k \xi_k^\mu\right) \left( \frac{1}{\sqrt{N}} \sum_i w_i^{la} \sigma\left( \frac{1}{\sqrt{D}} \sum_k F_{ki} \xi_k^\mu \right) \right)\right]- \frac{i}{\sqrt{N}} \sum_l \hat h_l \sum_i w_i^{l1} \sigma\left( \frac{1}{\sqrt{D}} \sum_k F_{ki} \xi_k^\star \right)}
	\end{multline}
	The average over patterns and features is now straightforward using the central limit theorem of section~\ref{sec::CLT}. We finally get
	\begin{multline}
		\overline{\left\langle y^\star \right\rangle_{\boldsymbol{w} | \left\{ \boldsymbol{\xi}^\mu \right\}}^s} = \lim\limits_{n \to 0} \int \prod_l \frac{d h_l d \hat h_l}{2 \pi} e^{i \sum_l h_l \hat h_l - \frac{\mu_\star^2}{2} \sum_{lm} q_{11}^{lm} \hat h_l \hat h_m} \prod_l \text{sign}(h_l) \\
		\times \int \prod_{\substack{a<b \\ lm}}\frac{d q_{ab}^{lm} d \hat q_{ab}^{lm}}{2\pi} \prod_{\substack{a \le b \\ lm}}\frac{d p_{ab}^{lm} d \hat p_{ab}^{lm}}{2\pi} \prod_{al} \frac{d r_a^l d \hat r_a^l}{2\pi} e^{N \phi'}
	\end{multline}
	where we have defined
	\begin{subequations}
		\label{eq::Bayes_terms}
		\begin{align}
			\phi' &= -\sum_{a<b} \sum_{lm} q_{ab}^{lm} \hat q_{ab}^{lm} - \frac{\alpha_D}{2} \sum_{ab} p_{ab}^{lm} \hat p_{ab}^{lm} - \alpha_D \sum_{a} r_a^l \hat r_a^l + G_{SS} + \alpha_D G'_{SE} + \alpha G_{E}\\
			G_{SS} &= \ln \int \prod_{al} dw_{la} \, P_{w}(w_{la}) \, e^{ \, \frac{1}{2} \sum_{a\ne b} \hat q_{ab}^{lm} w_{la} w_{mb}}\\
			G'_{SE} &= \frac{1}{D} \sum_k \ln \int \prod_{al} \frac{d s_k^{la} d \hat s_k^{la}}{2\pi} \, e^{i \sum_{la} s_k^{la} \hat s_k^{la} + \sum_{la} \hat r_{a}^l s_k^{la} + \frac{1}{2}\sum_{ab} \sum_{lm} \hat p_{ab}^{lm} s_k^{la} s_k^{mb} - \frac{1}{2} \sum_{ab} \sum_{lm} q_{ab}^{lm} \hat s_k^{la} \hat s_k^{mb} - i \frac{\mu_1 \xi_k^\star}{\sqrt{D}} \sum_l \hat h_l s_k^{l1} } \\
			G_{E} &= \ln \int \prod_{la} \frac{d \lambda_{la} d \hat \lambda_{la}}{2\pi} \frac{d u d \hat u}{2\pi} \, e^{i u \hat u + i \sum_{la} \lambda_{la} \hat \lambda_{la} -\beta \sum_{la} \Theta\left(-u \lambda_{la}\right) - \frac{\hat u^2}{2} - \frac{1}{2} \sum_{ab} \sum_{lm} Q_{ab}^{lm} \hat \lambda_{la} \hat \lambda_{mb} - \hat u \sum_{la} M_{a}^l \hat \lambda_{la} }
		\end{align}
	\end{subequations}
	Apart for the different numbers of replicas $G_{SS}$ and $G_{E}$ have the same expression as before, see equations~\eqref{eq::termsSP}. The entropic-energetic term instead is the same as before apart for an additional term that depends on the test pattern $\boldsymbol{\xi}^\star$; for this reason we denote it  with a prime index, 
	\begin{equation}
		G'_{SE} = G_{SE} + \frac{1}{D} \delta G_{SE}
	\end{equation}
	where $G_{SE}$ is given in~\eqref{eq::termsSP_GSE} and
	\begin{subequations}
		\begin{align}
			\label{eq::deltaGSE}
			\delta G_{SE} &\equiv \sum_k \ln \llangle e^{- \frac{\mu_1 \xi_k^\star}{\sqrt{D}} \sum_l \hat h_l s_k^{l1}} \rrangle_k \\
			\llangle \bullet \rrangle_k &\equiv \frac{ \displaystyle \int \prod_{al} \frac{d s_k^{la} d \hat s_k^{la}}{2\pi} \, e^{i \sum_{la} s_k^{la} \hat s_k^{la} + \sum_{la} \hat r_{a}^l s_k^{la} + \frac{1}{2}\sum_{ab} \sum_{lm} \hat p_{ab}^{lm} s_k^{la} s_k^{mb} - \frac{1}{2} \sum_{ab} \sum_{lm} q_{ab}^{lm} \hat s_k^{la} \hat s_k^{mb}} \bullet}{\displaystyle \int \prod_{al} \frac{d s_k^{la} d \hat s_k^{la}}{2\pi} \, e^{i \sum_{la} s_k^{la} \hat s_k^{la} + \sum_{la} \hat r_{a}^l s_k^{la} + \frac{1}{2}\sum_{ab} \sum_{lm} \hat p_{ab}^{lm} s_k^{la} s_k^{mb} - \frac{1}{2} \sum_{ab} \sum_{lm} q_{ab}^{lm} \hat s_k^{la} \hat s_k^{mb}}}
		\end{align}
	\end{subequations}
	Given that at first order in $D$, $G'_{SE}$ is equal to $G_{SE}$, we therefore have the same saddle point equations for every ansatz over replicas, as expected. Next, we can expand for large $D$ equation~\eqref{eq::deltaGSE}
	\begin{equation}
		\delta G_{SE} = \ln \prod_k \left[ 1 - i \frac{\mu_1}{\sqrt{D}} \xi_k^\star \sum_l \llangle s_k^{l1} \rrangle_k \hat h_l - \frac{\mu_1^2}{2D} (\xi_k^\star)^2 \sum_{lm} \llangle s_k^{l1} s_k^{m1} \rrangle_k \hat h_l \hat h_m\right] \,.
	\end{equation}
	As can be seen from~\eqref{eq::Bayes_terms}, the measure $\llangle \bullet \rrangle_k$ is related to the derivatives of $G_{SE}$; therefore we can use the saddle point equations and substitute the complicated integral expression with corresponding order parameters
	\begin{subequations}
		\begin{align}
			\llangle s_k^{la} \rrangle_k &= \frac{\partial G_{SE}}{\partial \hat r_a^l} = r_a^l \\
			\llangle s_k^{la} s_k^{mb} \rrangle_k &= \frac{\partial G_{SE}}{\partial \hat p_{ab}^{lm}} = p_{ab}^{lm} 		
		\end{align}
	\end{subequations}
	Notice how the right-hand expressions do not depend on $k$ anymore, since $G_{SE}$ is factorized over this index. We obtain
	\begin{equation}
		\begin{split}
			\delta G_{SE} &= \ln \prod_k \left[ 1 - i \frac{\mu_1}{\sqrt{D}} \xi_k^\star \sum_l r_1^l \hat h_l - \frac{\mu_1^2}{2D} (\xi_k^\star)^2 \sum_{lm} p_{11}^{lm} \hat h_l \hat h_m\right] \\
			&\simeq - i \mu_1 \left( \frac{1}{\sqrt{D}} \sum_k \xi_k^\star \right) \sum_l r_1^l \hat h_l - \frac{\mu_1^2}{2} \left( \frac{1}{D} \sum_k (\xi_k^\star)^2\right) \sum_{lm} \left( p_{11}^{lm} - r_1^l r_1^m \right) \hat h_l \hat h_m \,.
		\end{split}
	\end{equation}
	In the RS ansatz we get
	\begin{equation}
		\begin{split}
			\delta G_{SE} &\simeq - i \mu_1 \left( \frac{1}{\sqrt{D}} \sum_k \xi_k^\star \right) r \sum_l \hat h_l - \frac{\mu_1^2}{2} \left( \frac{1}{D} \sum_k (\xi_k^\star)^2\right) \left[  \left( p - r^2 \right) \left(\sum_{l} \hat h_l \right)^2 + \left( p_{d} - p \right) \sum_{l} \hat h_l^2 \right] \,.
		\end{split}
	\end{equation}
	so that 
	\begin{multline}
		\overline{\left\langle y^\star \right\rangle_{\boldsymbol{w} | \left\{ \boldsymbol{\xi}^\mu \right\}}^s} = 
		\int \prod_l \frac{d h_l d \hat h_l}{2 \pi} e^{i \sum_l \left( h_l - \mu_1 m_\xi^\star r \right) \hat h_l - \frac{1}{2} \left(\mu_\star^2 (1-q) + \mu_1^2 \sigma_\xi^\star \left( p_{d} - p \right) \right) \sum_{l} \hat h_l^2 - \frac{1}{2} \left( \mu_1^2 \sigma_\xi^\star \left( p - r^2 \right) + \mu_\star^2 q \right)  \left(\sum_{l} \hat h_l \right)^2 } \prod_l \text{sign}(h_l) 
	\end{multline}
	where
	\begin{subequations}
		\begin{align}
			m_\xi^\star &\equiv \frac{1}{\sqrt{D}} \sum_k \xi_k^\star\,, \\
			\sigma_\xi^\star &\equiv \frac{1}{D} \sum_k \left( \xi_k^\star \right)^2 \,.
		\end{align}
	\end{subequations}
	Using an Hubbard-Stratonovich transformation we finally obtain
	\begin{equation}
		\begin{split}
			\overline{\left\langle y^\star \right\rangle_{\boldsymbol{w} | \left\{ \boldsymbol{\xi}^\mu \right\}}^s} &=
			\int Dz \left[ \int \frac{d h d \hat h}{2 \pi} e^{i \left( h - \mu_1 m_\xi^\star r - \sqrt{ \mu_1^2 \sigma_\xi^\star \left( p - r^2 \right) + \mu_\star^2 q } \, z \right) \hat h - \frac{1}{2} \left(\mu_\star^2 (1-q) + \mu_1^2 \sigma_\xi^\star \left( p_{d} - p \right) \right) \hat h^2} \text{sign}(h) \right]^s \\
			&= \int Dz \left[ \int Dh \, \text{sign}\left(\mu_1 m_\xi^\star r + \sqrt{\mu_\star^2 (1-q) + \mu_1^2 \sigma_\xi^\star \left( p_{d} - p \right)} \, h + \sqrt{ \mu_1^2 \sigma_\xi^\star \left( p - r^2 \right) + \mu_\star^2 q } \, z \right) \right]^s \\
			&= \int Dz \left[ \text{erf} \left( \frac{\mu_1 m_\xi^\star r + \sqrt{ \mu_1^2 \sigma_\xi^\star \left( p - r^2 \right) + \mu_\star^2 q } \, z }{\sqrt{2\mu_\star^2 (1-q) + 2\mu_1^2 \sigma_\xi^\star \left( p_{d} - p \right)}} \right) \right]^s
		\end{split}
	\end{equation}
	Inserting this expression into~\eqref{eq::trick} and~\eqref{eq::Bayesian_Gen_error1} we find
	\begin{equation}
		\begin{split}
			\label{eq::gen_error_barycenter}
			\epsilon_g^B &= 
			\int Dz \, \mathbb{E}_{\boldsymbol{\xi}^\star} \Theta\left[ - m_\xi^\star \, \text{erf} \left( \frac{\mu_1 m_\xi^\star r + \sqrt{ \mu_1^2 \sigma_\xi^\star \left( p - r^2 \right) + \mu_\star^2 q } \, z }{\sqrt{2\mu_\star^2 (1-q) + 2\mu_1^2 \sigma_\xi^\star \left( p_{d} - p \right)}} \right) \right] \\
			&= \int Dz Du \, \Theta\left[ - u \, \left( \mu_1 r u + \sqrt{ \mu_1^2 \left( p - r^2 \right) + \mu_\star^2 q } \, z  \right) \right] = 2 \int_0^\infty Du \, H\left( \frac{\mu_1 r u}{\sqrt{ \mu_1^2 \left( p - r^2 \right) + \mu_\star^2 q } } \right) \\ 
			&= \frac{1}{\pi} \arccos\left( \frac{\mu_1 r}{\sqrt{\mu_1^2 p + \mu_\star^2 q}} \right) = \frac{1}{\pi} \arccos\left( \frac{M}{\sqrt{Q}} \right)
		\end{split}
	\end{equation}
	Notice that if we want to compute the generalization error of the barycenter of typical solutions with a given margin $\kappa$, the formula above remains the same. The only dependence on the margin is implicit in the order parameters $q$, $p$ and $r$. 
	
	The behaviour of the generalization error of the barycenter of typical solutions with vanishing and non-vanishing margins can be found in Fig.~\ref{Fig::gen_err_plateau}. As shown in the main text the barycenter achieving the minimal generalization error has a margin $\kappa_{\text{opt}}$ that undergoes a transition when crossing the value $\alpha = \alpha^{*}$: $\kappa_{\text{opt}}=0$ for $\alpha>\alpha^*$ whereas it becomes larger then zero when $\alpha < \alpha^*$. 
	
	\section{Local entropy lanscape of solutions}
	\subsection{Analytical approach: Franz-Parisi entropy}
	To study the local entropy landscape of solutions around a given typical configuration we use the Franz-Parisi approach~\cite{franz1995recipes,huang2014origin}. The Franz-Parisi free entropy is defined as
	\begin{equation}
		\label{eq::FP_free_entropy}
		\Phi_{FP}(t_1) = \frac{1}{Z} \int \prod_i d \tilde w_i \, P_{w}(\tilde{\boldsymbol{w}}) e^{- \tilde{\beta} \sum_{\mu} \ell_{NE} \left( - y^\mu \tilde{\lambda}^\mu; \tilde \kappa \right)} \ln \mathcal{N}(\tilde{\boldsymbol{w}}, t_1)
	\end{equation}
	where $\mathcal{N}(\tilde{\boldsymbol{w}}, t_1)$ is the number of configurations $\boldsymbol{w}$ extracted from the Gibbs measure that have an overlap $t_1$ with the reference configuration $\tilde{\boldsymbol w}$
	\begin{equation}
		\label{eq::FP_N}
		\mathcal{N}(\tilde{\boldsymbol{w}}, t_1) \equiv \int \prod_i dw_i \, P_{w}(\boldsymbol w) e^{-\beta \sum_{\mu} \ell_{NE} \left( - y^\mu \lambda^\mu; \kappa \right)} \delta \left( \sum_{i} w_i \tilde w_i - N t_1 \right) \,.
	\end{equation}
	In order to compute the Franz-Parisi free entropy, we introduce two sets of replicas, one for the partition function $Z$ in the denominator of~\eqref{eq::FP_free_entropy} (replica index $a=1, \dots, n$), and the other one for the logarithm in the same equation (replica index $c=1,\dots, s$)
	\begin{equation}
		\begin{split}
			\Phi_{FP}(t_1) &= \lim\limits_{n \to 0} \lim\limits_{s \to 0} \partial_s \int \prod_{ia} d \tilde w_i^a \, \prod_{a=1}^n P_{w}(\tilde{\boldsymbol{w}}^a) e^{- \tilde{\beta} \sum_{\mu a} \ell_{NE} \left( - y^\mu \tilde{\lambda}^\mu_a; \tilde \kappa \right)} \mathcal{N}^s(\tilde{\boldsymbol{w}}^{a=1}, t_1) \\
			&= \lim\limits_{\substack{n \to 0 \\ s \to 0}} \partial_s \int \prod_{ia} d \tilde w_i^a \int \prod_{ia} d w_i^c  \, \prod_{a=1}^n P_{w}(\tilde{\boldsymbol{w}}^a) \prod_{c=1}^s P_{w}(\boldsymbol{w}^a) \prod_c \delta \left( \sum_i w_i^c \tilde{w}_i^1 - N t_1 \right)\\
			&\times e^{- \tilde{\beta} \sum_{\mu a} \ell_{NE} \left( - y^\mu \tilde{\lambda}^\mu_a; \tilde \kappa \right) - \beta \sum_{\mu c} \ell_{NE} \left( - y^\mu \lambda^\mu_c; \kappa \right)}
		\end{split}
	\end{equation} 
	The computation is more involved, but proceeds in the same way as before; first of all we extract the teacher and student preactivations (both for the reference and constrained configurations)
	\begin{equation}
		\begin{split}
			\Phi_{FP}(t_1) &= \lim\limits_{\substack{n \to 0 \\ s \to 0}} \partial_s \int \prod_{\mu} \frac{d u^\mu d \hat u^\mu}{2\pi} \prod_{\mu a} \frac{d \tilde{\lambda}^\mu_a d \hat{\tilde{\lambda}}^\mu_a}{2\pi} \prod_{\mu c} \frac{d \lambda_c^\mu d \hat \lambda^\mu_c}{2\pi} \prod_{\mu} e^{i u_\mu \hat u_\mu + i \sum_a \tilde{\lambda}^\mu_a \hat{\tilde{\lambda}}^\mu_a + i \sum_c \lambda^\mu_c \hat{\lambda}^\mu_c}\\ 
			&\times \int \prod_{ia} d \tilde w_i^a \prod_{ia} d w_i^c  \, \prod_{a=1}^n P_{w}(\tilde{\boldsymbol{w}}^a) \prod_{c=1}^s P_{w}(\boldsymbol{w}^a) \prod_c \delta \left( \sum_i w_i^c \tilde{w}_i^1 - N t_1 \right)\\
			&\times \prod_{\mu} e^{- \tilde{\beta} \sum_{a} \ell_{NE} \left( - y^\mu \tilde{\lambda}^\mu_a; \tilde \kappa \right) - \beta \sum_{c} \ell_{NE} \left( - y^\mu \lambda^\mu_c; \kappa \right) - i \hat u_\mu \frac{1}{\sqrt{D}} \sum_k w_k^T \xi_k^\mu - i \sum_a \hat{\tilde{\lambda}}^\mu_a \frac{1}{\sqrt{N}} \sum_i w_i^a \tilde{\xi}_i^\mu - i \sum_c \hat \lambda_c^\mu \frac{1}{\sqrt{N}} \sum_i w_i^c \tilde{\xi}_i^\mu } \,.
		\end{split}
	\end{equation}
	Then we average over the patterns and features, using the central limit theorem of Section~\ref{sec::CLT}. We finally find
	\begin{equation}
		\begin{split}
			\Phi_{FP}(t_1) &= \lim\limits_{\substack{n \to 0 \\ s \to 0}} \partial_s \int \prod_{a<b} \frac{d \tilde{q}_{ab} d \hat{\tilde{q}}_{ab}}{2 \pi} \prod_{c<d} \frac{d q_{cd} d \hat q_{cd}}{2 \pi} \prod_{a \le b} \frac{d \tilde{p}_{ab} d \hat{\tilde{p}}_{ab}}{2 \pi} \prod_{c \le d} \frac{d p_{cd} d \hat p_{cd}}{2 \pi} \prod_a \frac{d \tilde r_{a} d \hat{\tilde{r}}_{a}}{2 \pi} \prod_c \frac{d r_{c} d \hat r_{c}}{2 \pi} \\
			&\times \int \prod_{ac} \frac{d k_{ac} d\hat k_{ac}}{2\pi} \prod_{c, a \ne 1} \frac{d t_{ac} d \hat t_{ac}}{2\pi} e^{N\phi_{FP}}
		\end{split}
	\end{equation}
	where
	\begin{subequations}
		\begin{align}
			\phi_{FP} &= -\sum_{a<b} \tilde q_{ab} \hat{\tilde q}_{ab} - \frac{\alpha_D}{2} \sum_{ab} \tilde p_{ab} \hat{\tilde p}_{ab} - \alpha_D \sum_{a} \tilde r_a \hat{\tilde r}_a -\sum_{c<d} q_{cd} \hat q_{cd} - \frac{\alpha_D}{2} \sum_{cd} p_{cd} \hat p_{cd} - \alpha_D \sum_{c} r_c \hat r_c \\
			&\nonumber \quad - \alpha_D \sum_{ac} k_{ac} \hat k_{ac} - \sum_{ac} t_{ac} \hat t_{ac} + G_{SS} + \alpha_D G_{SE} + \alpha G_{E} \\
			G_{SS} &= \ln \int \prod_a d \tilde w_a \, P_{w}(\tilde w_a) \, \int \prod_c dw_c \, P_{w}(w_c) \, e^{ \, \frac{1}{2}\sum_{a \ne b} \hat{\tilde q}_{ab} \tilde w_a \tilde w_b - \frac{1}{2}\sum_{a	\ne b} \hat q_{cd} w_c w_d +  \sum_{ac} \hat t_{ac} \tilde w_a w_c}\\
			\label{eq::FP_GSE}
			G_{SE} &= \ln \int \prod_a \frac{d \tilde s_a d \hat{\tilde s}_a}{2\pi} \prod_c \frac{d s_c d \hat{s}_c}{2\pi} \, e^{i \sum_a \tilde s_a \hat{\tilde s}_a + i \sum_c s_c \hat s_c + \sum_a \hat{\tilde{r}}_a \tilde s_a + \sum_c \hat r_c s_c + \frac{1}{2}\sum_{ab} \hat{\tilde p}_{ab} \tilde s_a \tilde s_b + \frac{1}{2}\sum_{cd} \hat p_{cd} s_c s_d} \\
			&\nonumber  \times e^{- \frac{1}{2} \sum_{ab} \tilde q_{ab} \hat{\tilde s}_a \hat{\tilde s}_b  - \frac{1}{2} \sum_{cd} q_{cd} \hat s_c \hat s_d + \sum_{ac} \hat k_{ac} \tilde s_a s_c - \sum_{ac} t_{ac} \hat{\tilde{s}}_a \hat s_c} \\
			G_{E} &= \ln \int \prod_a \frac{d \tilde \lambda_a d \hat{\tilde \lambda}_a}{2\pi} \prod_a \frac{d \lambda_c d \hat \lambda_c}{2\pi} \frac{d u d \hat u}{2\pi} \, e^{i u \hat u + i \sum_a \tilde \lambda_a \hat{\tilde \lambda}_a + i \sum_{c} \lambda_c \hat{\lambda}_c - \tilde \beta \sum_c \Theta\left(- \text{sign}(u) \tilde \lambda_c + \tilde\kappa \right) -\beta \sum_a \Theta\left(- \text{sign}(u) \lambda_a + \kappa \right)} \\
			&\nonumber \times e^{- \frac{\hat u^2}{2} - \frac{1}{2} \sum_{ab} \tilde Q_{ab} \hat{\tilde \lambda}_a \hat{\tilde \lambda}_b - \frac{1}{2} \sum_{cd} Q_{cd} \hat \lambda_c \hat \lambda_d - \hat u \sum_a \tilde M_a \hat{\tilde \lambda}_a - \hat u \sum_c M_c \hat \lambda_c - \sum_{ac} T_{ac} \hat{\tilde \lambda}_a \hat \lambda_c}
		\end{align}
	\end{subequations}
	All the order parameters appearing in the previous formulas are
	\begin{align*}
		M_{c} &= \mu_1 r_{c}\,, & r_{c} &= \frac{1}{D} \sum_{k=1}^D s_k^c\,, & s_k^a &= \frac{1}{\sqrt{N}} \sum_{i=1}^{N} F_{ki} w_i \\	
		\tilde M_{a} &= \mu_1 \tilde r_{a}\,, & \tilde r_{a} &= \frac{1}{D} \sum_{k=1}^D \tilde s_k^a\,, &  \tilde s_k^a &= \frac{1}{\sqrt{N}} \sum_{i=1}^{N} F_{ki} \tilde w_i \\
		Q_{cd} &= \mu_1^2 p_{cd} + \mu_\star^2 q_{cd}\,, & p_{cd} &= \frac{1}{D} \sum_{k=1}^{D} s_k^c s_k^d\,, & q_{cd} &= \frac{1}{N} \sum_{i=1}^{N} w_i^c w_i^d\\
		\tilde Q_{ab} &= \mu_1^2 \tilde p_{ab} + \mu_\star^2 \tilde q_{ab}\,, & \tilde p_{ab} &= \frac{1}{D} \sum_{k=1}^{D} \tilde s_k^a \tilde s_k^b\,, & \tilde q_{ab} &= \frac{1}{N} \sum_{i=1}^{N} \tilde w_i^a \tilde w_i^b\\
		T_{ac} &= \mu_1^2  k_{ac} + \mu_\star^2 t_{ac}\,, & k_{ac} &= \frac{1}{D} \sum_{k=1}^{D} \tilde s_k^a s_k^c\,, & t_{ac} &= \frac{1}{N} \sum_{i=1}^{N} \tilde w_i^a w_k^c \,.
	\end{align*}
	We have understood that $t_{1c} \equiv t_1$ as this condition is imposed by the delta function in equation~\eqref{eq::FP_N}.
	
	Notice that, as in Section~\ref{sec::average}, the entropic-energetic term $G_{SE}$ is Gaussian, so it can be readily solved. Defining the quantities
	\begin{subequations}
		\begin{align}
			\hat{\overline{r}}_{\alpha} &\equiv \left( \hat{\tilde{r}}_a, \hat r_c \right) \in \mathbb{R}^{n+s}\,,\\	
			\overline{q}_{\alpha \beta} &\equiv \begin{pmatrix}
				\tilde{q}_{ab} & t_{ac} \\
				t_{ac} & q_{cd}
			\end{pmatrix} \in \mathbb{R}^{(n+s) \times (n+s)} \\
			\hat{\overline{p}}_{\alpha \beta} &\equiv \begin{pmatrix}
				\hat{\tilde{p}}_{ab} & k_{ac} \\
				k_{ac} & \hat p_{cd}
			\end{pmatrix} \in \mathbb{R}^{(n+s) \times (n+s)}
		\end{align}
	\end{subequations}
	it can be seen that~\eqref{eq::FP_GSE} can be written in the same way as~\eqref{eq::termsSP_GSE} in terms of $\hat{\overline{r}}_{\alpha}$, $\overline{q}_{\alpha \beta}$ and $\hat{\overline{p}}_{\alpha \beta}$, so that 
	\begin{equation}
		G_{SE} = - \frac{1}{2} \ln \det \left( \mathbb{I} - \overline{q} \, \hat{\overline{p}} \right) + \frac{1}{2} \sum_{\alpha \beta} \hat{\overline{r}}_\alpha \left[ \left(\mathbb{I} - \overline{q} \, \hat{\overline{p}} \right)^{-1} \overline{q}  \right]_{\alpha \beta} \hat{\overline{r}}_\beta \,.
	\end{equation}
	
	\subsubsection{RS ansatz}
	We impose an RS ansatz over the order parameters: 
	\begin{subequations}
		\begin{align}
			q_{ab} &= \delta_{ab} + q \left(1 - \delta_{ab} \right) & \hat q_{ab} &= \hat q \left(1 - \delta_{ab} \right)\\
			p_{ab} &= p_d \delta_{ab} + p \left(1 - \delta_{ab} \right) & \hat p_{ab} &= - \hat p_d \delta_{ab} + \hat p \left(1 - \delta_{ab} \right)\\
			r_a &= r & \hat r_a &= r\\
			k_{ac} &= k_1 \delta_{a1} + k_0 (1-\delta_{a1}) & \hat k_{ac} &= \hat k_1 \delta_{a1} + \hat k_0 (1-\delta_{a1}) \\
			t_{ac} &= t_1 \delta_{a1} + t_0 (1-\delta_{a1}) & \hat t_{ac} &= \hat t_1 \delta_{a1} + \hat t_0 (1-\delta_{a1})
		\end{align}
	\end{subequations}
	A similar ansatz is imposed for the tilde order parameters. 
	
	\subsubsection{Entropic-Entropic and Energetic terms}
	Let us start from the entropic-entropic contribution. This term is exactly equal to the entropic contribution of a storage problem~\cite{huang2014origin}. It is equal to
	\begin{equation}
		\mathcal{G}_{SS} \equiv \frac{\hat q}{2} + \lim\limits_{\substack{n \to 0 \\ s \to 0}} \partial_s G_{SS} = \int Dx \frac{\sum_{\tilde w = \pm 1} e^{\sqrt{\hat{\tilde q}} \tilde w x } \int Dy \ln 2 \cosh\left( \sqrt{\hat{q}-\frac{\hat t_0^2}{\hat{\tilde q}}} y + \frac{\hat t_0}{\sqrt{\hat{\tilde q}}} x + (\hat t_1 - \hat t_0) \tilde w \right)}{2 \cosh \left( \sqrt{\hat{\tilde q}} x \right)}
	\end{equation}
	The energetic term is a bit more involved. It is however equal (apart for a redefinition of order parameters) to the energetic term that is obtained in a classic teacher-student problem. It is equal to
	\begin{equation}
		\mathcal{G}_{E} \equiv \lim\limits_{\substack{n \to 0 \\ s \to 0}} \partial_s G_{E} = 2\int Dx Dy \, \frac{H_{\tilde \beta}\left(u(x,y) \right)}{H_{\tilde \beta} \left( h(x) \right)} \int_{h(x)}^{\infty} D z \, \ln H_{\beta} \left( v(x,y,z) \right)
	\end{equation}
	with
	\begin{subequations}
		\begin{align}
			u(x,y) &\equiv \displaystyle \frac{\tilde M \sqrt{\Gamma} \left( b y - a x \right) - M \sqrt{\tilde Q} y}{\sqrt{(\tilde Q - \tilde M^2)(\Gamma - M^2) - (T_0 - M \tilde M)^2}} \\
			v(x,y,z) &\equiv \frac{\kappa - \sqrt{\Gamma} (a y + b x) - \frac{T_1-T_0}{\sqrt{\tilde Q_d - \tilde Q}} z}{\sqrt{Q_d - Q}}\\
			h(x) &\equiv \frac{\tilde \kappa - \sqrt{\tilde Q} x}{\sqrt{\tilde Q_d - \tilde Q}}\\
			\Gamma &\equiv Q - \frac{(T_1-T_0)^2}{\tilde{Q}_d- \tilde{Q}} \\
			b &\equiv \frac{T_0}{\sqrt{\tilde Q \Gamma} }\,, \qquad a \equiv \sqrt{1-b^2}
		\end{align}
	\end{subequations}
	
	\subsubsection{Entropic-Energetic term}
	Following a series of algebraic manipulations, the entropic-energetic term reads
	\begin{equation}
		\begin{split}
			\mathcal{G}_{SE} &\equiv \lim\limits_{\substack{n \to 0 \\ s \to 0}} \partial_s G_{SE} = - \frac{1}{2} \ln \eta + \frac{1}{2\eta} \left[ \left( \hat p + \hat r^2 \right)(1-q) - \left( \hat p_d + \hat p \right) q \right] + \\
			&+\frac{1}{2 \eta \tilde \eta} \left\{ (1-q)(\hat k_1 - \hat k_0)^2 \left[ 1-q +
			\frac{1}{\tilde \eta}\left( (1-\tilde q)^2 (\hat{\tilde p} + \hat{\tilde r}^2) +q \right) \right] + 2(\hat k_1 - \hat k_0) \left[ (1-q)(1-\tilde q)(\hat k_0 + \hat r \hat{\tilde r}) + t_0 \right]\right\}	\\
			&+\frac{1}{2 \eta \tilde \eta} \left\{ (\hat{p}_d + \hat p )(t_1-t_0)^2 \left[ \hat{\tilde p}_d + \hat{\tilde p} - \frac{1}{\tilde \eta} \left( \hat{\tilde p} + \hat{\tilde r}^2 + (\hat{\tilde p}_d + \hat{\tilde p})^2 \tilde q\right) \right] + 2(t_1-t_0) \left[ \left( \hat p_d + \hat p \right)\left( \hat{\tilde p}_d + \hat{\tilde p} \right) t_0 + \hat k_0 + \hat r \hat{\tilde r}  \right]\right\} \\
			&+\frac{(\hat k_1 - \hat k_0) (t_1 - t_0)}{\eta \tilde \eta} \left[ 1 + \frac{1}{\tilde \eta} \left[ (1-\tilde q)\left(\hat{\tilde p} + \hat{\tilde r}^2\right) - (\hat{\tilde p}_d + \hat{\tilde p}) \tilde q \right] \right]
		\end{split}
	\end{equation}
	where we have defined the quantities
	\begin{subequations}
		\begin{align}
			\eta &\equiv 1 + (\hat p_d + \hat p)(1-q) \\
			\tilde \eta &\equiv 1 + (\hat{\tilde p}_d + \hat{\tilde p})(1-\tilde q) 
		\end{align}
	\end{subequations}	
	
	\subsubsection{Final expression of the free entropy}
	The RS Franz-Parisi free entropy is finally
	\begin{equation}
		\begin{split}
			\Phi_{FP}(t_1) = - \frac{\hat q}{2}(1-q) + \frac{\alpha_D}{2} \left( p_d \hat p_d + p \hat p \right) - \alpha_D r \hat r -\alpha_D (k_1 \hat k_1 - k_0 \hat k_0) - t_1 \hat t_1 + t_0 \hat t_0 + \mathcal{G}_{SS} + \alpha_D \mathcal{G}_{SE} + \alpha \mathcal{G}_{E}
		\end{split}
	\end{equation}
	The tilde order parameters being those one characterizing the reference configuration will satisfy the RS saddle point equation analyzed in Section~\ref{sec::Typical_RS}. The order parameters $q$, $\hat q$, $p_d$, $\hat p_d$, $p$, $\hat p$, $r$, $\hat r$, $k_1$, $\hat k_1$, $k_0$, $\hat k_0$, $\hat t_1$, $t_0$ and $\hat t_0$ are found by solving the saddle point equations obtained by taking the corresponding derivatives of the Franz-Parisi entropy and imposing them to be equal to zero. 
	
	\subsection{Numerical experiments}
	\begin{figure}
		\begin{centering}
			\includegraphics[width=0.49\columnwidth]{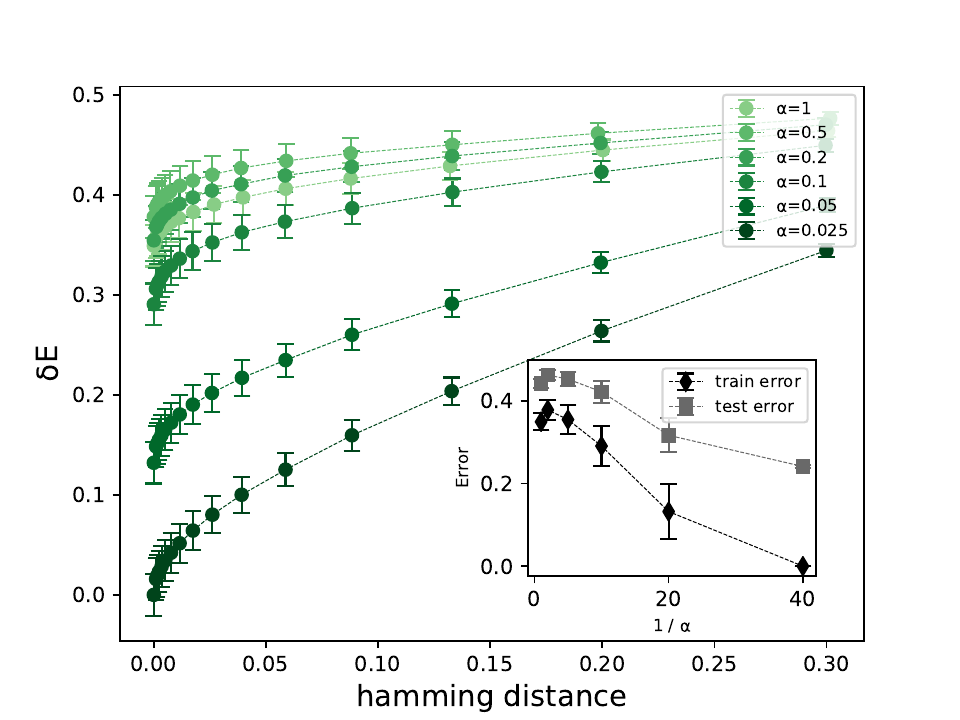}
			\includegraphics[width=0.49\columnwidth]{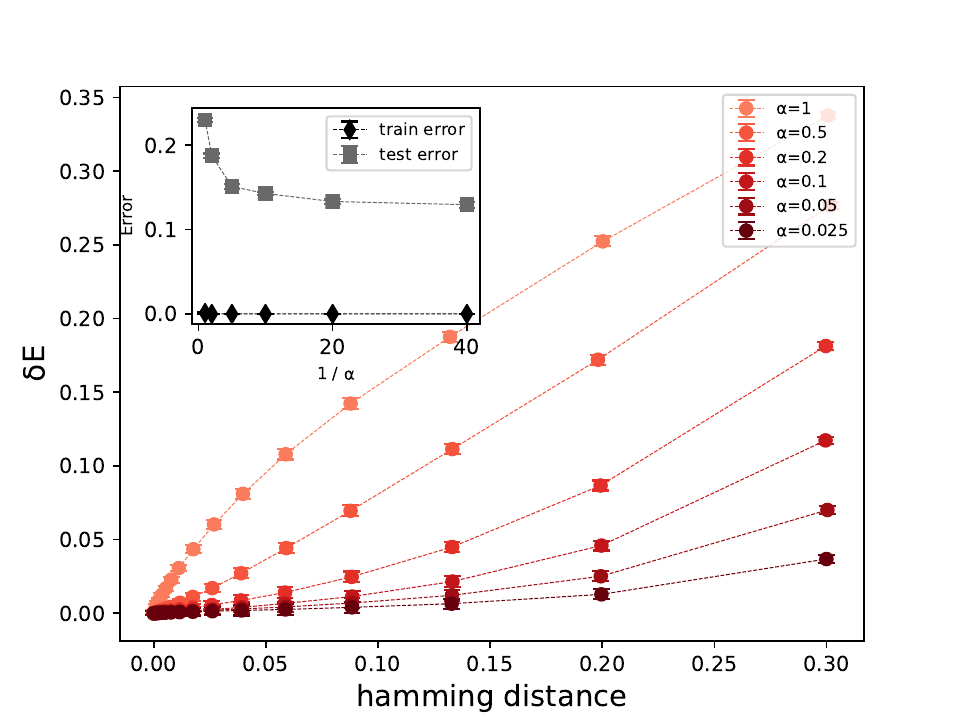}
			\includegraphics[width=0.49\columnwidth]{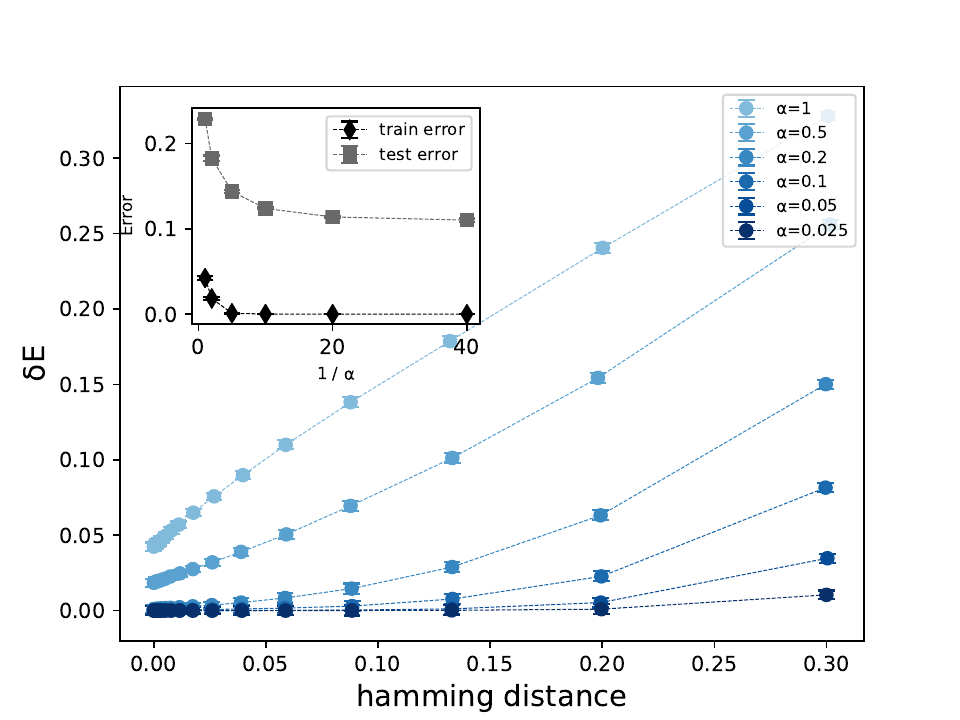}
		\end{centering}
		\caption{Local energy $\delta E$ as a function of distance from the reference solution in the overparameterized perceptron model fixing $\alpha_T=5$, for different values of $\alpha$ and for MCT0, SBPI and BP solutions. Train and test error are depicted in the insets. While MCT0 solutions are sharp, even in the $\alpha \to 0$ limit, other algorithms find solutions whose flatness increases as the student is more overparameterized.}
		\label{Fig::local_energy} 
	\end{figure}
	{\em Local energy curves.} To compare the geometrical structure of solutions found by different algorithms, we computed the local energy profiles~\cite{jiang2019fantastic,pittorino2021entropic}. 
	For all the architectures we have analyzed in the main text (continuous tree committee machine, multi-layer perceptrons, CNN), we have computed the local energy as follows. Given a solution to the learning problem, we perturbed it using a multiplicative Gaussian noise that acts on the network weights as follows
	\begin{equation*}
		W \xrightarrow[]{} W \left( 1+ \eta \right)
	\end{equation*}
	where $W$ is a weight of the network while the variance of the noise $\eta$ is tuned to obtain perturbed vectors that have increasing distance from $W$. 
	After the perturbation, the networks are normalized as explained in the main text, and we measure their Euclidean distance from the original solution. We repeatedly perturb every solution for each level of noise, and collect the distances and training errors. The ``local energy'' curve is the average training error rate displayed as a function of the average distance.
	
	In the main text we show curves for the multi-layer perceptron architecture and for the CNN.
	
	Here we show in Fig.~\ref{Fig::local_energy} the local energy curves for three different algorithms trained on the (binary) overparameterized perceptron: MCT0, SBPI and BP. In general one can argue that overparameterizing the network leads to higher flatness; however for the MCT0 algorithm, the solution is always sharp, as predicted by the replica theory.

	\begin{figure}
		\begin{centering}
			\includegraphics[width=0.49\columnwidth]{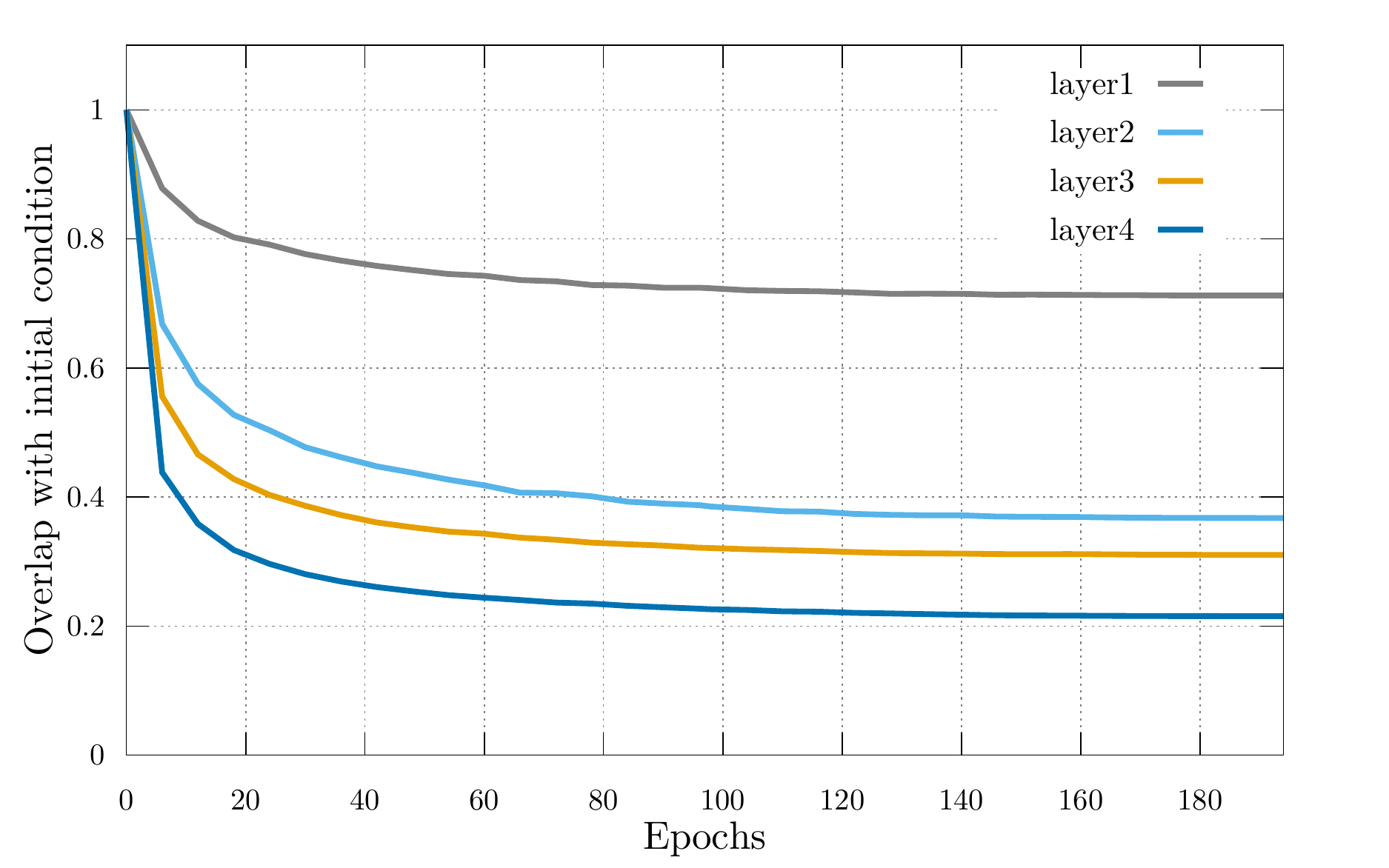}
		\end{centering}
		\caption{Overlap of a solution with its initial condition (epoch $0$) as a function of the number of epochs. The overlap is represented layer by layer and it refers to ADAM optimizer, a number of filters proportional to $C=20$, and a $lr = 0.01$.}
		\label{Fig::lazy regime} 
	\end{figure}

	{\em Lazy regime description for CNNs trained on CIFAR10 dataset.}
	These big architectures have been shown to work in the so-called lazy training regime: the initial layers changes less than the following ones. This can be quantified numerically studying the overlap between the single layer weight configuration at epoch $0$ and at epoch $t$, as shown in Fig.~\ref{Fig::lazy regime}. The overlap decreases for every layer, with a drop rate that depends on the layer position.

	\section{Material and Methods}
	{\bf Gaussian Equivalence theorem.} It has been shown by~\cite{Mei2019}, that in the thermodynamic limit~\eqref{eq::thermodynamic_limit}, the statistical properties of the random feature model are equivalent to a Gaussian covariate model, in which each projected pattern $\boldsymbol{\tilde{\xi}}^\mu$ is a \emph{linear} combination of the patterns components $\xi_k^\mu$ plus noise. The strength of the noise depends on the degree of non-linearity of the activation function $\sigma$. In mathematical terms the following mapping between different models holds
	\begin{equation}
		\tilde \xi_i^\mu = \sigma\left( \frac{1}{\sqrt{D}} \sum_{k=1}^{D} F_{ki} \xi_k^\mu \right) = \mu_0 + \frac{\mu_1}{\sqrt{D}} \sum_{k=1}^{D} F_{ki} \xi_k^\mu + \mu_\star \eta_i^\mu
	\end{equation}
	where $\eta_i \sim \mathcal{N}(0, 1)$ are i.i.d. standard Gaussian random variables and $\mu_0 = \int Dz \, \sigma(z)$, $\mu_1 = \int Dz \, z \, \sigma(z)$, $\mu_2 = \int Dz \, \sigma^2(z)$, $\mu_\star^2 = \mu_2 - \mu_1^2 - \mu_0^2$ with $Dz \equiv \frac{e^{-z^2/2}}{\sqrt{2\pi}}$. We provide a sketch of the proof, based on the explicit computations of the moments, in the SI and refer to~\cite{Mei2019,Goldt2020} for more details.
	
	{\bf Numerical Experiments on the Binary Perceptron.} Here we report the details for the numerical experiments performed on the overparameterized binary perceptron. 
	(SA) Simulated Annealing, based on a standard Metropolis algorithm that attempts one weight flip at a time, is run until either a solution is found or a maximum number of sweeps ($4000$) is reached, where a sweep consists of $N$ attempted moves. We used an initial inverse temperature $\beta=1.0$ that is increased at every sweep with a linear increment $\Delta\beta=5\cdot10^{-3}$. (SBPI) For a complete description of the SBPI algorithm see ref.~\cite{baldassi2009generalization}. In the numerical experiments we set the maximum number of allowed iterations to $500$ and used a threshold $\theta_{\text{m}}=2$ and a probability $p_{\text{s}}=0.3$ of updating the synapses with a stability $0\leq \Delta^{\mu} \leq \theta_{\text{m}}$. (BP) We used a standard BP implementation with damping $\delta=0.5$ and a maximum number of updates fixed to $200$. The magnetization are randomly initialized with a uniform distribution in the interval $[-\epsilon, \epsilon]$ with $\epsilon=10^{-2}$. (fBP) For the focusing BP algorithm we used the same initialization of BP, and a damping factor $\delta=0.9$. We set the number of virtual replicas to $y=10$ and update the messages until convergence for $30$ steps, each time increasing the coupling strength $\gamma$ according to $\gamma = \text{atanh}\left( i / 29 \right)$ where $i = 0,\dots , 29$ (using by convention $\gamma=10$ for the last step). (BNet) We used the standard implementation of BinaryNet (see. ref~\cite{hubara2016binarized}) using $\text{sign}$ activation function and cross-entropy loss, without using batch normalization. We fixed the learning rate $\eta=5\cdot10^{-3}$ and ran a full batch gradient update for $2000$ epochs.
	
	{\bf Numerical Experiments on the committee machine.} Here we report the details for the numerical experiments performed on the overparameterized continuous tree-like committee machine, for the two algorithms used. (fBP) In all experiments, we set $y = 10$ and ranged $\gamma$ between 0.5 and 30 with an exponential schedule divided into $30$ steps; at each step, the algorithm was run (with damping $\delta=0.1$) until convergence (with a convergence criterion set to $\epsilon=10^{-2}$) or at most $200$ iterations. (SGD) The expression for the cross-entropy loss in the binary classification case, with a scale parameter $\gamma$, is: $f_\gamma\left(x\right) = -\frac{x}{2}+\frac{1}{2\gamma}\log\left(2\cosh\left(\gamma x\right)\right)$. This is just the standard expression but with the input $x$ scaled by $\gamma$ (which is equivalent to setting the norm of the input weights to $\gamma$) and the output scaled by $1/\gamma$ (equivalent to scaling the gradients by $1/\gamma$). In all the experiments, we set the batch size to $100$, the maximum number of epochs to $700$, and the learning rate to $10^{-2}$. The weights were initialized from a uniform distribution and
	then normalized for each unit. The norm parameters $\gamma$ and $\beta$ were initialized at the values $\gamma=10$, $\beta=1$ and multiplied by $1 + 10^{-4}$, $1 + 10^{-2}$, respectively, after each epoch. The algorithm stopped as soon as it found a solution (this was determined using the desired architecture with $\mathrm{sign}$ activation and output functions, which is equivalent to letting $\beta,\gamma\to\infty$ and checking for a zero-error).
	
	{\bf Numerical Experiments on the Multi-layer neural network} Our implementation follows closely the one of ref.~\cite{geiger2020scaling}. We set the learning rate to $10^{-4}$ and train the model with full batch gradient descent with ADAM optimization for a fixed number of epochs ($5000$).
	In all the simulations we used ReLU non-linearities and the square-hinge loss with a margin fixed to $1$.
	In order to train the model with the adversarial initialization, we first trained the network using SGD with a fixed number of epochs ($5000$) (with minibatches of size $128$ and learinig rate set to $5\cdot10^{-3}$) on a modified train set in which the labels have been randomized, then we used the resulting weights as the initial condition for ADAM.
	
	{\bf Numerical Experiments on CNNs.} We used standard PyTorch initialization (HE for ReLU activation function) and the cross-entropy loss function for all the experiments. We took 5 independent samples for both optimizers: SGD with momentum and ADAM. In all experiments, the learning rate was $10^{-2}$, the batch size $50$ and the number of epochs was $200$. For SGD the momentum was set to $0.3$. All the additional parameters were taken as in the default Pytorch settings. The number of network parameters was controlled by the value to $C$ as defined in the main text.
	
\end{document}